\documentclass[lettersize,journal]{IEEEtran}
\usepackage{amsmath,amsfonts}
\usepackage{algorithmic}
\usepackage{amssymb}
\usepackage{array}
\usepackage[caption=false,font=normalsize,labelfont=sf,textfont=sf]{subfig}
\usepackage{threeparttable}
\usepackage{textcomp}
\usepackage[dvipsnames]{xcolor}
\usepackage{stfloats}
\usepackage{url}
\usepackage{verbatim}
\usepackage{graphicx}
\usepackage{cite}
\usepackage{multirow}
\usepackage[inkscapeformat=png]{svg}
\usepackage[colorlinks=true,linkcolor=NavyBlue,citecolor=BrickRed]{hyperref}
\usepackage[capitalise]{cleveref}
\usepackage[normalem]{ulem}
\usepackage{enumitem}
\usepackage{arydshln}

\newcommand{\cy}[1]{\textcolor{black}{#1}}

\usepackage[switch]{lineno}

\begin{document}

\title{CromSS: Cross-modal pretraining with noisy labels for remote sensing image segmentation}

\author{Chenying Liu,~\IEEEmembership{Student Member,~IEEE}, 
Conrad M Albrecht,~\IEEEmembership{Member,~IEEE}, 
Yi Wang,~\IEEEmembership{Student Member,~IEEE}, 
Xiao Xiang Zhu,~\IEEEmembership{Fellow,~IEEE}
\thanks{*Accepted by IEEE TGRS.

C. Liu is with the Chair of Data Science in Earth Observation, Technical University of Munich (TUM) and the Remote Sensing Technology Institute, German Aerospace Center (DLR) and Munich Center for Machine Learning (MCML), Germany. 

C. M. Albrecht is with the Remote Sensing Technology Institute, German Aerospace Center (DLR). 

Y. Wang is with the Chair of Data Science in Earth Observation, Technical University of Munich (TUM). 

X. X. Zhu is with the Chair of Data Science in Earth Observation, Technical University of Munich (TUM) and Munich Center for Machine Learning (MCML), Germany.

The work of C.\ Liu, Y.\ Wang and C.\ Albrecht was funded by the Helmholtz Association through the Framework of \textit{HelmholtzAI}, grant ID: \texttt{ZT-I-PF-5-01} -- \textit{Local Unit Munich Unit @Aeronautics, Space and Transport (MASTr)}. The compute related to this work is supported by the Helmholtz Association's Initiative and Networking Fund on the HAICORE@FZJ partition. C. Albrecht receives additional funding from the European Union's Horizon Europe research and innovation programme under grant agreement No.\ \texttt{101082130} (\textit{EvoLand}). The work of X.\ X.\ Zhu is supported by the German Federal Ministry of Education and Research (BMBF) in the framework of the international future AI lab ``AI4EO -- Artificial Intelligence for Earth Observation: Reasoning, Uncertainties, Ethics and Beyond'' (grant number: 01DD20001). The work of C.\ Liu and X.\ X.\ Zhu is also supported by Munich Center for Machine Learning (MCML).}

}

\markboth{Journal of \LaTeX\ Class Files,~Vol.~14, No.~8, August~2021}%
{Shell \MakeLowercase{\textit{et al.}}: A Sample Article Using IEEEtran.cls for IEEE Journals}

\IEEEpubid{0000--0000/00\$00.00~\copyright~2021 IEEE}

\maketitle

\begin{abstract}
\cy{We explore the potential of large-scale noisily labeled data to enhance feature learning by pretraining semantic segmentation models within a multi-modal framework for geospatial applications. We propose a novel \underline{Cro}ss-modal \underline{S}ample \underline{S}election (\textit{CromSS}) method, a weakly supervised pretraining strategy designed to improve feature representations through cross-modal consistency and noise mitigation techniques. Unlike conventional pretraining approaches, CromSS exploits massive amounts of noisy and easy-to-come-by labels for improved feature learning beneficial to semantic segmentation tasks. We investigate middle and late fusion strategies to optimize the multi-modal pretraining architecture design. We also introduce a cross-modal sample selection module to mitigate the adverse effects of label noise, which employs a cross-modal entangling strategy to refine the estimated confidence masks within each modality to guide the sampling process. Additionally, we introduce a spatial-temporal label smoothing technique to counteract overconfidence for enhanced robustness against noisy labels.
To validate our approach, we assembled the multi-modal dataset, \textit{NoLDO-S12}, which consists of a large-scale noisy label subset from Google's Dynamic World (DW) dataset for pretraining and two downstream subsets with high-quality labels from Google DW and OpenStreetMap (OSM) for transfer learning. Experimental results on two downstream tasks and the publicly available \textit{DFC2020} dataset demonstrate that when effectively utilized, the low-cost noisy labels can significantly enhance feature learning for segmentation tasks. The data, codes, and pretrained weights are freely available at \url{https://github.com/zhu-xlab/CromSS}.}
\end{abstract}

\begin{IEEEkeywords}
pretraining, noisy labels, semantic segmentation, multi-modal deep learning, sample selection, SSL4EO-S12 dataset, geospatial artificial intelligence 
\end{IEEEkeywords}


\section{Introduction} \label{sec:intro}

\IEEEPARstart{I}n spite of the effectiveness of deep learning in big data processing \cite{zhao_artificial_2024}, obtaining enough accurate labeled data for deep learning model training is challenging due to the high cost of the annotation process \cite{li_learning_2021}. Self-supervised learning (SSL) is promising to address this issue by means of learning hidden representations from unlabeled data \cite{wang_self-supervised_2022, zhu_foundations_2024}.  
Popular SSL algorithms include generative Masked Autoencoders (MAE) \cite{he_masked_2022} and contrastive learning methods such as DINO \cite{caron_emerging_2021} and MoCo \cite{chen_improved_2020}. MAE is inspired by image reconstruction, as most works utilizing vision transformers (ViTs) \cite{dosovitskiy_image_2021}. When applying MAE to remote sensing (RS), RingMo introduced a patch-incomplete strategy to reduce the loss of dense and small objects for Masked Image Modeling (MIM) pretraining \cite{sun_ringmo_2023}. SatMAE employed extra temporal embeddings and spectral positional encodings to explore temporal information in multi-spectral images \cite{cong_satmae_2022}, which is improved with multi-scale pretraining by satMAE++ \cite{noman_rethinking_2024}. Scale-MAE improved MIM towards better scale awareness \cite{reed_scale-mae_2023}. On the other hand, contrastive learning methods, which push the semantically similar input patches to be close in representation space, can make a difference for both convolutional backbones like ResNets \cite{he_deep_2016} and ViTs. To this end, the data augmentation strategies used to generate positive and negative samples are key. In the RS domain, such strategies are designed with geo-locations \cite{ayush_geography-aware_2021}, temporal or seasonal information \cite{manas_seasonal_2021}, multi-size cropping \cite{wanyan_dino-mc_2023}, etc. \cite{mall_change-aware_2023}. However, these methods primarily target image-level information, e.g., for classification tasks, leading to suboptimal performance when applied to tasks requiring pixel-level information such as semantic segmentation \cite{wang_ssl4eo-s12_2023}. This discrepancy motivates us to explore alternative strategies to enhance the efficacy of pretrained models for dense mapping tasks like segmentation. 

\IEEEpubidadjcol

Before the adoption of SSL, pretrained models were dominantly trained on large-scale annotated datasets in computer vision, such as ImageNet \cite{deng_imagenet_2009}.
\cy{Recently, the SAM (Segment Anything Model) series \cite{kirillov_segment_2023, ravi_sam_2024} has achieved remarkable success in image segmentation by training on millions of labeled images and videos, demonstrating robust generalization capabilities for diverse downstream tasks like underwater image segmentation \cite{wang_large_2025}.}
Similarly, tailored to remote sensing (RS) image understanding, the SatlasPretrain dataset was developed to support large-scale pretraining on RS data.
However, such vast amounts of accurately annotated labels are rare, driving researchers to explore sources for ``weak'' labels, which are noisy but much lower in cost and easy to obtain on a large scale. 
\cy{Recent studies have demonstrated the resilience of deep learning models to a certain degree of label noise \cite{zhang_understanding_2021, liu_ai02_2024}, showing that the deep learning models trained on large-scale noisy labels can exhibit strong feature learning capabilities and transfer learning performance across various applications. For instance, noisy social-media labels have been successfully used for image classification \cite{mahajan_exploring_2018} and video analysis \cite{ghadiyaram_large-scale_2019}. In remote sensing, crowd-sourced maps such as OpenStreetMap (OSM) have been explored to pretrain models for tasks like building and road extraction \cite{kaiser_learning_2017}. Notably, Maggiori et al. \cite{maggiori_convolutional_2017} observed that the shallow layers of a network pretrained with noisy labels remain largely unchanged after fine-tuning. This implies that the resilience to label noise varies across different parts of a deep learning model, with shallow layers particularly adept at learning robust features even when trained on noisy labels.} In our previous work \cite{liu_task_2024}, we analyzed the segmentation model’s (U-Net) robustness in a per-module fashion to further qualify this phenomenon. We compared two U-Nets trained with exact (ground-truth) and noisy labels. We observe that: 1) the encoder features visually share similar spatial characteristics, in which case the closer the convolutional layer gets to the U-Net’s output, the more the features become contaminated by label noise; 2) similar weight statistics govern the encoder layers, while those in the decoder follow diverging weight statistics towards the semantic segmentation outputs. These observations highlight that encoder features are less biased by label noise, yet they benefit from the semantics provided by pixel-level noisy label masks. We attribute this phenomenon to encoders’ preference to learn basic spatial features more from the input data than from the labels. From this perspective, transferring such weakly supervised pretrained encoder weights to semantic segmentation downstream applications could be highly beneficial.

\cy{On the other hand, we have observed a surging tendency in multi-modal learning, exemplified by vision-language learning such as image captioning \cite{li_underwater_2025} and visual question answering (VQA) \cite{yuan_easy_2022}.}
In the remote sensing domain, a plethora of satellite data modalities is becoming accessible: from multi-spectral, hyperspectral, to Synthetic Aperture Radar (SAR), to Light Detection And Ranging (LiDAR) data, and many more \cite{schmitt2015, hong_multimodal_2021, sainte_fare_garnot_multi-modal_2022, cai_improving_2023, sun_similarity_2024}. Different modalities can provide complementary information to each other, such as the radiation features of optical images and the structural features of SAR images. This cross-modal enhancement can potentially boost the mutual learning between modalities, even in self-supervised learning for RS data. For example, CROMA extends MAE from a single modality to multiple modalities alongside cross-modal contrastive learning \cite{fuller_croma_2023}. SkySense, based on MAE, too, features a modular design accommodating diverse tasks and modalities \cite{guo_skysense_2023}. DeCUR defines a contrastive loss to explicitly model features as shared or orthogonal among modalities \cite{wang_decur_2023}. DOFA, leveraging the concept of neural plasticity in brain science, integrates various data modalities into a single framework by adjusting to different wavelengths \cite{xiong_neural_2024}. These researches demonstrate that multi-modal learning is beneficial in feature learning and, thus, is promising to make a difference in the noisy label pretraining framework.

{Multi-modal learning mainly includes two kinds of strategies: data fusion and co-learning. Data fusion is to add or concatenate data/features from different modalities in the input, feature, or decision levels \cite{chen_self-supervised_2022}. This strategy usually requires all the modalities in both the training and inference stages. As a more flexible solution, co-learning leverages losses to integrate knowledge from different modalities by enforcing consistency between their outputs to enhance model training \cite{xie_co-learning_2023}. In this case, each modality has specific models and can be employed separately after training. Therefore, we adopt the co-learning framework for multi-modal noisy label pretraining to maintain this flexibility in transfer learning settings. This structure is analogous to multi-model sample selection methods of learning from noisy labels (LNL), such as Co-teaching \cite{huang_co-seg_2021} and Decoupling \cite{malach_decoupling_2017}, where two identical models are trained separately and select samples for each other to avoid overfitting to label noise. The basic idea behind this is that the samples with noisy labels have higher inconsistency than correctly annotated pixels during the training \cite{han_co-teaching_2018}. We expect some similar behaviors in the co-learning noisy label pretraining framework. The diverse features from different modalities can lead to different responses to label noise, making it possible to integrate some sample selection strategies to remove part of the reverse effects of label noise.
}

\cy{In this work, we introduce a novel \underline{Cro}ss-modal \underline{S}ample \underline{S}election method, referred to as \textit{CromSS}, to explore the potential of using large-scale noisy labels and multi-modal learning to enhance the feature learning capabilities of RS image semantic segmentation models.} 
We nest CromSS into a co-learning framework to maintain the flexibility of pretrained models when transferred to downstream tasks. To this end, we employ two U-Nets \cite{ronneberger_u-net_2015} (using ResNet-50 as backbones \cite{he_deep_2016}) to separately extract features and make predictions for each modality, namely optical and SAR. Except for applying segmentation losses to each model, we push the predictions to be close with additional consistency losses between their softmax outputs. We also test an implicit consistency constraint of utilizing shared weights for the two decoders. Following the naming of data fusion, we name the pure co-learning architecture \textit{late fusion} since the two models are only forced to mimic each other at the very end of the outputs. We call the shared decoder architecture \textit{middle fusion} as the implicit constraint is additionally put on the feature level. Notice that fusion happens in a hidden way in our work instead of data addition or concatenation. We include the two architectures without explicit preferences, given that no evidence shows which one is superior. Our experiments provide some references to readers regarding the performance of the two architectures.
Furthermore, we introduce a sample selection strategy within the multi-modal framework encouraged by LNL multi-model-based sample selection methods to alleviate the adverse effects of label noise. We weight the loss functions with the sample selection masks obtained in three steps: confidence mask generation, cross-modal confidence enhancement, and thresholding. We first estimate the two kinds of confidence, label-based and entity-based, for segmentation and consistency losses. Then, the confidence masks are enhanced by fortifying the shared information across modalities. Two thresholding strategies are applied to each enhanced confidence mask to derive final selection masks. Here, we introduce a spatial-temporal label smoothing technique tailored for the used SSL4EO-S12 dataset with 4 timestamps of each given geospatial area (patch) to replace the uniform distribution-based label smoothing to ensure robustness in pretraining.

In our experiments, we utilize Sentinel-1 (S1) of SAR and Sentinel-2 (S2) of multi-spectral data from the SSL4EO-S12 dataset \cite{wang_ssl4eo-s12_2023} as two modalities. We propose a new dataset extended from SSL4EO-S12, called \textit{NoLDO-S12}, which pairs \underline{S}entinel-\underline{1} and \underline{2} data with \underline{No}isy \underline{L}abels sourced from the Google Dynamic World (DW) project \cite{brown_dynamic_2022} for pretraining, alongside two land cover\slash land use segmentation downstream tasks with ground truth (GT) labels sourced from \underline{D}W and \underline{O}penStreetMap (OSM), to evaluate the proposed method. We also involve another publicly available dataset DFC2020 \cite{yokoya_2020_2019} to test the transferability of the model. Experimental results indicate the effectiveness of CromSS with its task-specific pretraining for RS image segmentation.

{In summary, our main contributions are:
\begin{itemize}[leftmargin=5ex,topsep=0.5pt,itemsep=0.5ex,partopsep=0ex,parsep=0ex]
    \item The development of a novel, task-specific noisy label pretraining method called CromSS, which is proposed to explore the potential of noisy labels in feature learning for remote sensing image segmentation, cf. \cref{sec:meth}. CromSS is spotlighted with two characteristics. One is the multi-modal framework nesting CromSS to boost feature learning with noisy labels inspired by recent advances in multi-modal learning. The other is the sample selection strategy based on the enhanced confidence masks by fortifying the shared information across modalities to enhance the performance by excluding partial label noise. 
    \item The curation of a multi-modal dataset named NoLDO-S12 (\cref{sec:data}), which is composed of a pretraining subset with pixel-wise noisy labels and two segmentation downstream tasks for land cover\slash land use classification with Sentinel-1\slash 2 images as inputs, cf. \cref{sec:exp}. The noisy label pretraining subset is the first global dataset for this purpose. The two downstream tasks complement existing land cover\slash land use classification datasets based on Sentinel data.
    \item The experimental evaluation of the proposed CromSS compared with other state-of-the-art pretraining strategies on three different RS image segmentation downstream tasks, cf. \cref{sec:exp}. Furthermore, we analyzed the learned features by different pretraining methods to shed light on the future design of pretraining and feature learning methodologies.
\end{itemize}
}

\section{NoLDO-S12 dataset} \label{sec:data}

NoLDO-S12 contains two splits: SSL4EO-S12@NoL with noisy labels for pretraining based on the SSL4EO-S12 dataset \cite{wang_ssl4eo-s12_2023}, and two downstream datasets, SSL4EO-S12@DW and SSL4EO-S12@OSM, for transfer learning with exact labels. We detail them in \cref{sec:data:pre-train,sec:data:downstream}, respectively. 

\subsection{pretraining task} \label{sec:data:pre-train}

\begin{figure*}
    \centering
    \includegraphics[width=0.95\linewidth]{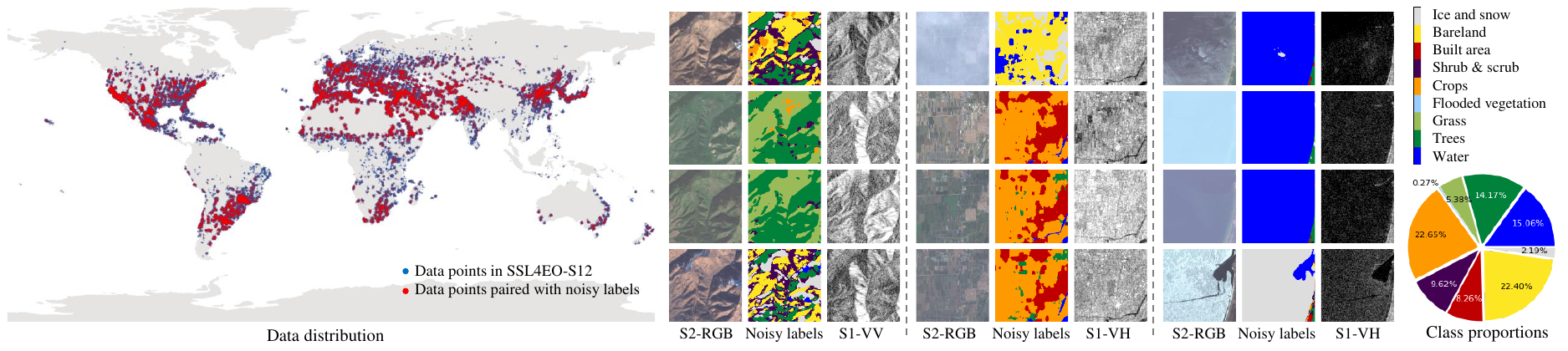}
    \caption{Illustration of the pretraining set SSL4EO-S12@NoL in NoLDO-S12. From left to right: global distribution of samples (left), 4-season samples (top-down) at 3 geolocations (middle), and statistics of the classes of the noisy labels (right).}
    \label{fig:data:pre-train}
\end{figure*}

The pretraining dataset SSL4EO-S12@NoL extends the large-scale, multi-modal, and multi-temporal self-supervision dataset SSL4EO-S12 \cite{wang_ssl4eo-s12_2023}. SSL4EO-S12 contains 251k$\times$ globally sampled Sentinel-1 (S1, radar) and Sentinel-2 (S2, optical) image pairs of 264x264 pixels and 4 seasons. In our case, we employ S1 ground range detected (GRD) SAR images of 2 bands, and S2 level-1C (L1C) multi-spectral data of 13 bands. All bands are resampled to a spatial resolution of 10 meters. We paired these data with the 9-class noisy labels sourced from the DW project on Google Earth Engine. DW generates noisy land cover labels with semi-supervised deep learning based on the Sentinel-2 L1C product. As per \cite{brown_dynamic_2022}, the labels bear an overall accuracy of approx.\ 70\%. We picked the DW project for the following reasons: the 9-class schema aligns with other global land cover land use maps such as ESA's \textit{Land Use and Coverage Area frame Survey} (LUCAS) \cite{joint_research_centre_european_commission_lucas_2020}. DW offers near real-time mapping providing multi-temporal label masks. 
As a result, SSL4EO-S12@NoL noisily labels (NoL) about 41\% (103,793 out of the 251,079 locations) of the SSL4EO-S12 dataset. To keep dataset's multi-temporal characteristics, we only retain the S1-S2-noisy label triples from the locations where all 4 timestamps of S1-S2 pairs in SSL4EO-S12 have corresponding DW labels, as illustrated in \cref{fig:data:pre-train}. This dataset well reflects real-world use cases where noisy labels remain more difficult to obtain compared to bare S1-S2 image pairs.  

\subsection{Downstream tasks} \label{sec:data:downstream}

\begin{figure*}
    \centering
    \includegraphics[width=0.95\linewidth]{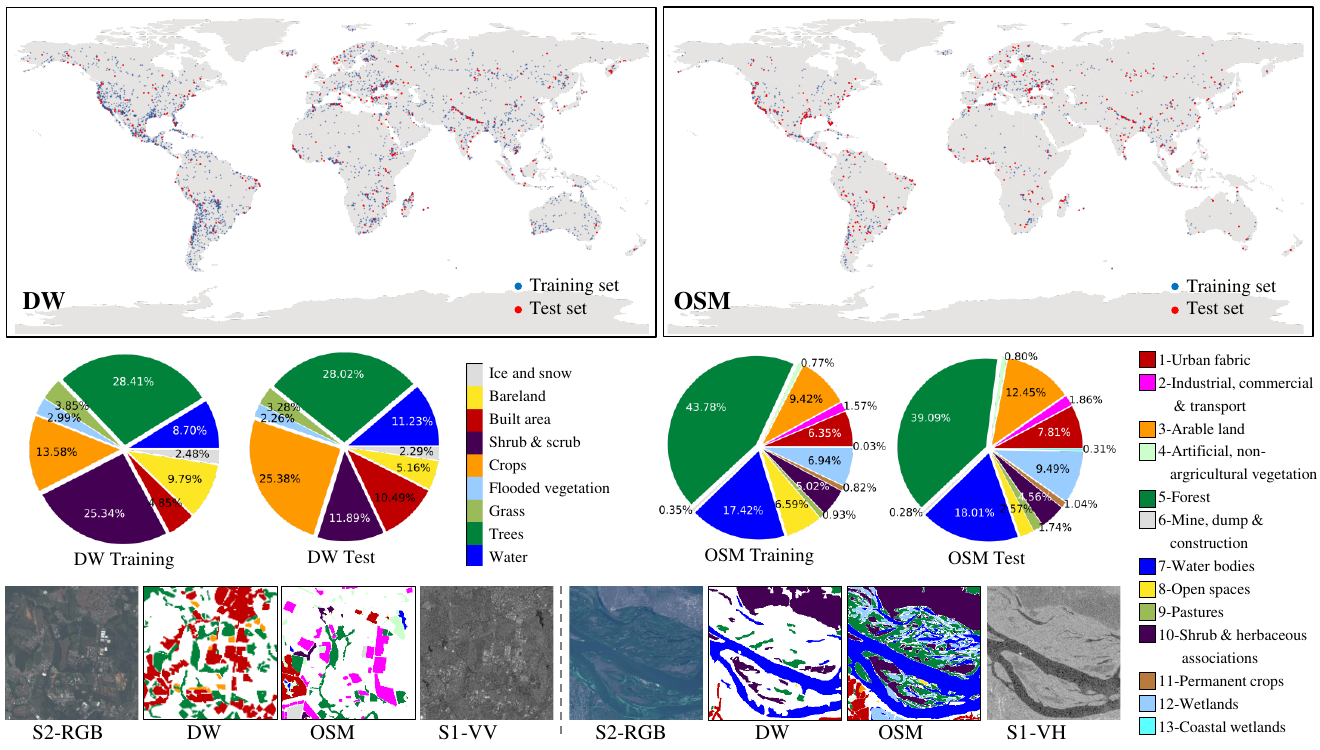}
    \caption{Illustration of the two downstream tasks in NoLDO-S12 with different label sources (SSL4EO-S12@DW and SSL4EO-S12@OSM). Top (left and right): global data distributions(DW and OSM). Middle (left and right): class distributions of training and test sets along with corresponding legends (DW and OSM). Bottom: examples from 2 locations. The legend for DW labels is the same as that in Figure~\ref{fig:data:pre-train}.}
    \label{fig:data:downstream}
\end{figure*}

We construct two downstream datasets, SSL4EO-S12@DW and SSL4EO-S12@OSM. Both are selected on the DW project's manually annotated training and validation datasets. However, they are paired with different label sources from DW and OSM. For DW, the human labeling process allows some ambiguous areas left unmarked (white spots in DW masks in \cref{fig:data:downstream}). We exclusively utilize the expert labeled training subset of 4,194 tiles with given dimensions of 510$\times$510 pixels (training data)\footnote{The DW training labels were downloaded from \url{https://doi.pangaea.de/10.1594/PANGAEA.933475} (\texttt{Expert} directory).} to construct our downstream tasks. The DW hold-out validation set of 409 tiles with given dimensions of 512$\times$512 is used to construct our test data\footnote{The DW hold-out validation set is built from the annotations where at least two out of three domain experts agree on the labels, \textit{Expert Consensus}, downloaded from \url{https://zenodo.org/records/4766508}.}. 
{The validation set was geographically sampled to ensure that its geographical distribution aligns with that of the training set.}
We spatial-temporally aligned the S1 and S2 data for the training and test tiles with Google Earth Engine.
Consequently, 3,574 training tiles and 340 test tiles were successfully paired. These serve as the first downstream task in SSL4EO-S12@DW with a total of 656,758,064 training pixels and 60,398,506 test pixels. {The class distributions can be found in \cref{fig:data:downstream}. SSL4EO-S12@DW test set has a slightly different distribution from the training set. We consider it an acceptable difference inherently caused by geospatial sampling of the splits.}

To test the generability of pretrained models, we construct another downstream dataset with fine-grained labels derived from OpenStreetMap\footnote{OSM labels were downloaded from the OSMLanduse layer at \url{https://osmlanduse.org/\#12/8.7/49.4/0/} via the Web Map Service (WMS) at \url{https://maps.heigit.org/osmlanduse/service}.}. We follow the work of Schultz et al.\ \cite{schultz_open_2017}, adopting a 13-class schema similar to the classification level 2 of the CORINE Landcover classes \cite{noauthor_corine_nodate}, as listed in \cref{fig:data:downstream}. We retrieved 2,996 OSM label masks among the 3,914=3,574+340 DW tiles. The remaining subset of 918 tiles is left without OSM labels. To keep the quality of OSM labels, we conduct an automatic check with DW labels as reference. By definition, we construct a many-to-many mapping of class schema from OSM (integers) to DW in \cref{tab:OSM2DWRel}.
Treating DW labels as ground truth and leveraging the OSM-DW relation in \cref{tab:OSM2DWRel}, for each of the $N$ labeled pixels of a tile: we check whether an OSM label agrees with any of the corresponding DW labels to compute the overall accuracy $acc=TP/N$. $TP$ denotes all true positive matches according to \cref{tab:OSM2DWRel}. Filtering for $acc>.75$, a total of 1,772 tiles remain in the SSL4EO-S12@OSM downstream subset, with 1,372 for training and 400 for test. Due to filtering, we note that only 106 out of the 340 SSL4EO-S12@DW test tiles remain ($512\times512$ pixels per tile). To substitute for the reduction in tiles, we utilize additional 294 tiles from the SSL4EO-S12@DW training dataset ($510\times510$ pixels per tile). For reasons of consistency, the 294 test tiles of size $510\times510$ were padded by a one-pixel zero values boundary to increase them to a size of 512$\times$512 pixels. 
All our (convolutional) semantic segmentation models tested in \cref{sec:exp} (and many more from the literature) are able to consume $512\times512$ pixel images as input.
We implemented additional manual checks and corrections by human inspection on the test set to further improve the reliability of test results. The total numbers of training and test pixels are 165,993,707 and 44,535,192. Please refer to the upper right charts in \cref{fig:data:downstream} for additional data statistics of the SSL4EO-S12@OSM downstream dataset.

\begin{table*}[htp]
\centering
\caption{Many-to-many relation of land cover labels from the OpenStreetMap (OSM) to the Dynamic World (DW) datasets.}
\label{tab:OSM2DWRel}
\scriptsize
\begin{tabular}{c|l||c|l||c|l}
\hline\hline
\bf OSM label ID(s) &\bf DW class names &\bf OSM label ID &\bf DW class names &\bf OSM label ID &\bf DW class names\\
\hline
1, 2 & Built Area               & 3 & Crops              & 4 & Trees, Grass\\
5    & Trees                    & 6 & Bare Ground        & 7 & Water\\
8    & Bare Ground, Snow \& Ice & 9 & Grass              & 10& Grass, Shrub \& Scrub\\
11   & Trees, Crops             & 12& Flooded vegetation & 13& Water, Flooded Vegetation \\
\hline\hline
\end{tabular}
\end{table*}

\section{Methodology} \label{sec:meth}

\begin{figure*}[htp]
    \centering
    \includegraphics[width=0.93\linewidth]{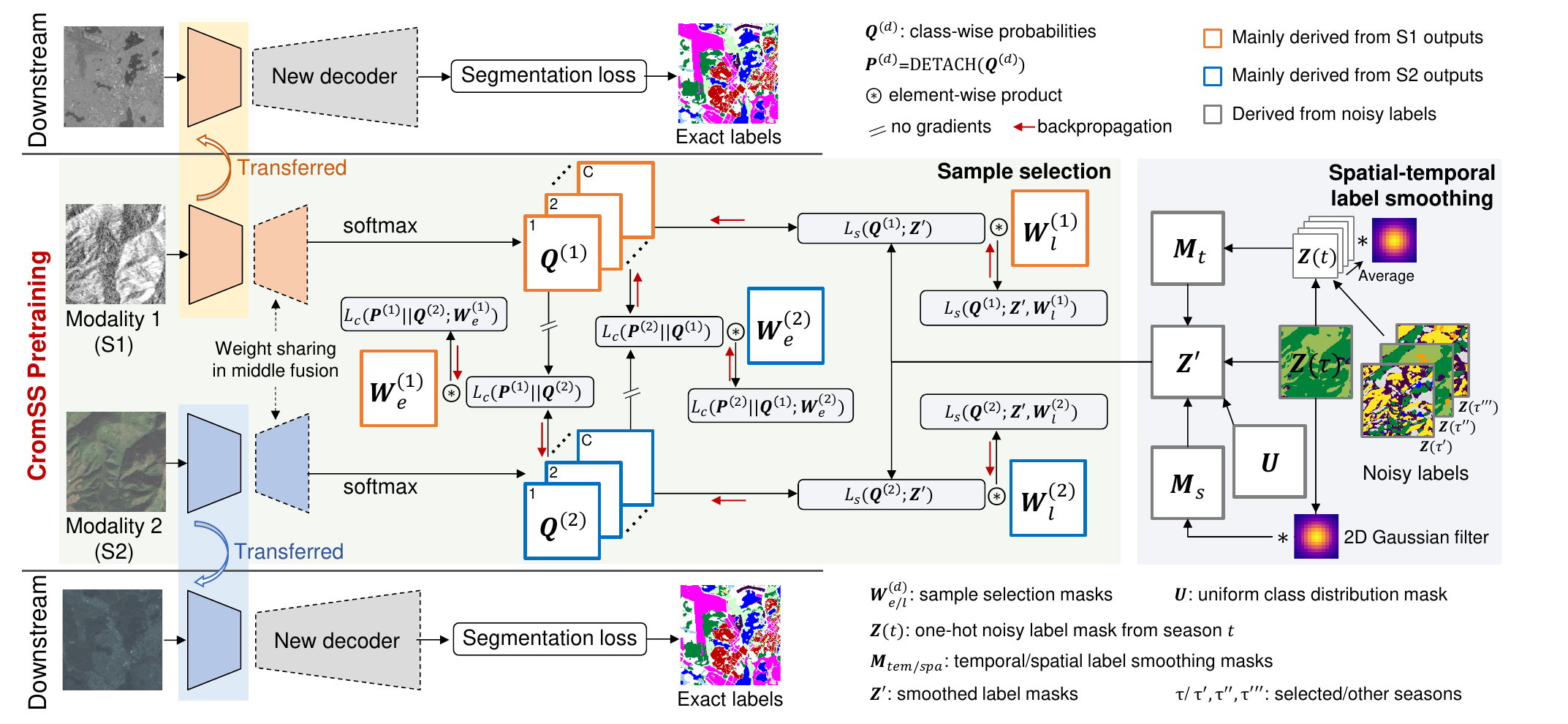}
    \caption{{Overview of the proposed cross-modal sample selection (CromSS) method.}}
    \label{fig:meth:cromss}
\end{figure*}

{As illustrated in \cref{fig:meth:cromss}, the proposed CromSS is a multi-modal weakly supervised pretraining method with noisy labels for remote sensing image segmentation tasks. The pretrained encoders by CromSS can then be transferred to different segmentation downstream tasks in combination with various decoders. 
Overall, CromSS is nested into a co-learning framework using consistency losses to boost the feature learning across modalities in addition to segmentation losses for pretraining with noisy labels. As demonstrated in the middle-left part of \cref{fig:meth:cromss}, two models specific to each modality are employed to separately extract features and make predictions before connecting via consistency losses. We can optionally add an explicit cross-modal constraint by using shared weights for decoders. We name the two architectures \textit{late fusion} and \textit{middle fusion} regarding whether to include the explicit constraint. We present the details of architecture design in \cref{sec:meth:fuse}. 
As highlighted in the green shadow, another spotlight of CromSS is the cross-modal sample selection strategy. We implement it by introducing weighting masks, i.e., selection masks ($W_{e/l}$), into segmentation/consistency loss function calculation to attenuate the impact of label noise. We employ a three-step strategy to generate these selection masks, which will be elaborated in \cref{sec:meth:ss}. Moreover, we include a spatial-temporal label smoothing technique to replace uniform distribution-based label smoothing to ensure robustness in pretraining. We briefly describe this small trick in \cref{sec:meth:smooth}.
}

\subsection{Multi-modal pretraining architectures} \label{sec:meth:fuse}

\begin{figure*}[htp]
    \centering
    \includegraphics[width=.95\linewidth]{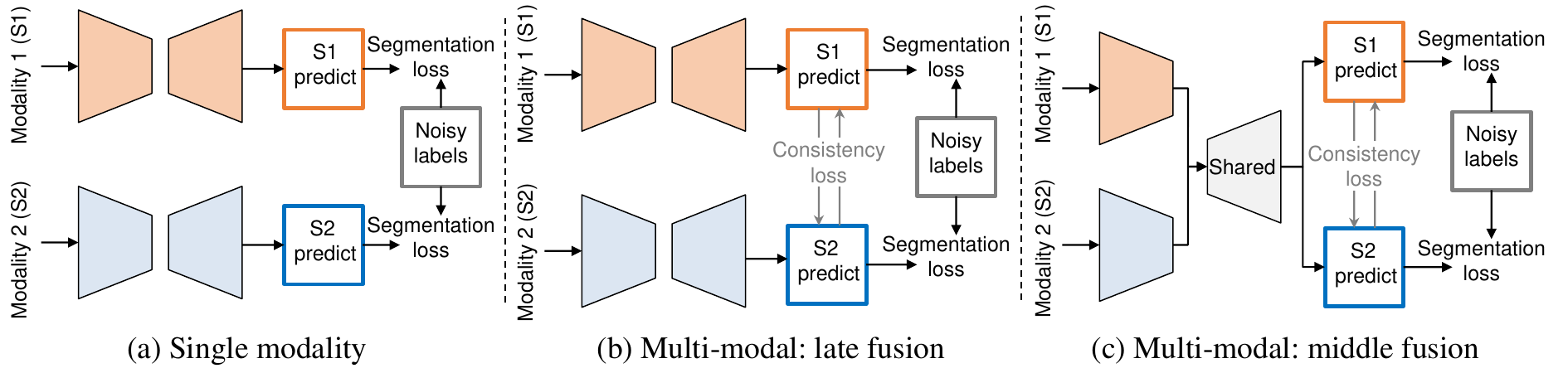}
    \caption{Comparison of single-modal training and multi-modal (middle\slash late) fusion strategies.}
    \label{fig:meth:mm_struct}
\end{figure*}

{We employ middle and late fusion architectures to explore complementary information across modalities to assist noisy label pretraining. Notice that our fusion strategies are implemented in an implicit way \underline{without} adding or concatenating feature vectors of different modalities. As illustrated in \cref{fig:meth:mm_struct}, the major difference between multi-modal and single-modal pretraining with noisy labels is the incorporation of additional consistency losses between the outputs from the two models. While late fusion retains individual decoders, middle fusion shares a common decoder for the two modalities.} The loss function for multi-modal pretraining with noisy labels can be written as
\begin{equation} \label{eq:loss:origin} 
\footnotesize
    L = L_{s}\left(\cy{\boldsymbol{Q}^{(1)}}\middle\vert\cy{\boldsymbol{Y}}\right) + L_{s}\left(\cy{\boldsymbol{Q}^{(2)}}\middle\vert\cy{\boldsymbol{Y}}\right) + L_{c}\left(\cy{\boldsymbol{P}^{(1)}}\middle\|\cy{\boldsymbol{Q}^{(2)}}\right) + L_{c}\left(\cy{\boldsymbol{P}^{(2)}}\middle\|\cy{\boldsymbol{Q}^{(1)}}\right).
\end{equation}
Specifically, we utilize a combined segmentation loss summing {the distribution-based \texttt{CrossEntropy} loss $L_{ce}$ and the region-based \texttt{Dice} loss $L_{dice}$ \cite{jadon_survey_2020} to represent
\begin{equation} \label{eq:loss:seg} 
\footnotesize
\begin{aligned}
    L_{s}\left(\cy{\boldsymbol{Q}^{(d)}}\middle|\cy{\boldsymbol{Y}}\right) & = L_{ce}\left(\cy{\boldsymbol{Q}^{(d)}}\middle|\cy{\boldsymbol{Y}}\right)+L_{dice}\left(\cy{\boldsymbol{Q}^{(d)}}\middle|\cy{\boldsymbol{Y}}\right)
\end{aligned}
\end{equation}
with
\begin{equation} \label{eq:loss:seg-ce} 
\footnotesize
\begin{aligned}
L_{ce}\left(\cy{\boldsymbol{Q}^{(d)}}\middle|\cy{\boldsymbol{Y}}\right) = \left(-\sum_{i,c}z_{ic}\log\left(q_{ic}^{(d)}\right)\right), 
\end{aligned}
\end{equation}
and
\begin{equation} \label{eq:loss:seg-dice} 
\footnotesize
\begin{aligned}
    L_{dice}\left(\cy{\boldsymbol{Q}^{(d)}}\middle|\cy{\boldsymbol{Y}}\right) = \left(1-2\frac{\sum_{i,c}z_{ic}\;q_{ic}^{(d)}}{\sum_{i,c}z_{ic}+q_{ic}^{(d)}}\right), 
\end{aligned}
\end{equation}
}where $z_{ic}\in\{0,1\}$ is the $c$th entry of the one-hot encoded noisy label vector $\mathbf{z}_i$ for the $i$th noisy sample annotation $y_i\in\cy{\boldsymbol{Y}}$. More specifically $z_{ic}=1$ when $c=y_i$, and $z_{ic}=0$ otherwise. The class index $c\in\mathbb{N}$ ranges from $1$ to \cy{C}, the total number of classes. \cy{N} is the number of image pixels with the index \cy{$i=1,\dots, \text{N}$} where \cy{$\text{N}=\text{W}\times \text{H}$}, \cy{W} and \cy{H} are the image width and height, respectively. The symbol $q_{ic}^{(d)}\in\cy{\boldsymbol{Q}^{(d)}}$ denotes the corresponding model prediction of modality $d\in\{1,2\}$ from the \textit{softmax} layer. For the consistency loss $L_{c}$, we pick the Kullback–Leibler (KL) divergence
\begin{equation} \label{eq:loss:con}
\small
    L_c\left(\cy{\boldsymbol{P}^{(d)}}\middle\|\cy{\boldsymbol{Q}^{(d')}}\right) =-\sum_{i,c} p_{ic}^{(d)}\log\left(\frac{q_{ic}^{(d')}}{p_{ic}^{(d)}}\right), 
\end{equation}
where $d\neq d'\in\{1,2\}$ are two modalities, and $p^{(d)}_{ic}\in \cy{\boldsymbol{P}^{(d)}}$ indicates the detached version of $q^{(d)}_{ic}$ without gradients in the backpropagation, cf.\ \cref{fig:meth:cromss_details}. 

In CromSS, we utilize sample selection masks $\cy{\boldsymbol{W}^{(d)}_{l/e}}$ (details in \cref{sec:meth:ss}) to assign smaller weights to the samples possibly associated with noisy labels. In practice, we employ the weighted average strategy to aggregate loss values from each sample (see the right part in \cref{fig:meth:cromss_details}). As a consequence, \cref{eq:loss:seg} and \cref{eq:loss:con} turn to,
\begin{equation}\label{eq:loss:seg-new} \small
\begin{aligned}    
    L_s\left(\cy{\boldsymbol{Q}^{(d)}}\middle\vert \cy{\boldsymbol{Z'}}, \cy{\boldsymbol{W}^{(d)}_l}\right)=
    -\sum_{i,c}w_{l,i}^{(d)}z'_{ic}\log q_{ic}^{(d)} \\
     + \left(1-2\frac{\sum_{i,c}w_{l,i}^{(d)}z'_{ic}q_{ic}^{(d)}}{\sum_{i,c}w_{l,i}^{(d)}\left(z'_{ic}+q_{ic}^{(d)}\right)}\right),
\end{aligned}
\end{equation}
and
\begin{equation} \label{eq:loss:con-new} \small
    L_c\left(\cy{\boldsymbol{P}^{(d)}}\middle\|\cy{\boldsymbol{Q}^{(d')}}\middle\vert \cy{\boldsymbol{W}_e^{(d)}}\right)=
    -\sum_{i,c}w^{(d)}_{e,i}p_{ic}^{(d)}\log\left[{q_{ic}^{(d')}}/{p_{ic}^{(d)}}\right],
\end{equation}
where $w_{l/e,i}^{(d)}\in \cy{\boldsymbol{W}^{(d)}_{l/e}}$ is the weight for sample $i$, $z'_{ic}\in \cy{\boldsymbol{Z'}}$ is the smoothed label from the one-hot counterpart $z_{ic}$ for sample $i$ at class $c$.

\subsection{Sample selection mask generation} \label{sec:meth:ss}

\begin{figure*}[htp]
    \centering
    \includegraphics[width=.95\linewidth]{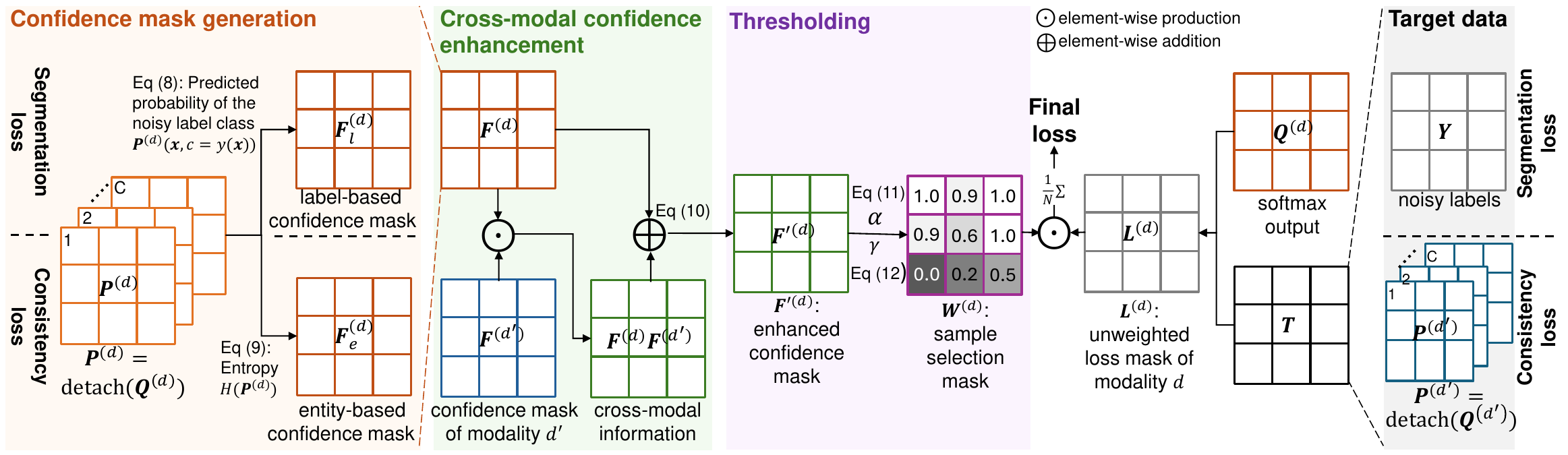}
    \caption{Sample selection mask generation in three steps: confidence mask generation, cross-modal confidence enhancement, and thresholding, where $d$ and $d'$ represent two different modalities, $\alpha$ and $\gamma$ are the thresholds used to derive the sample selection masks for the segmentation and consistency losses, respectively.}
    \label{fig:meth:cromss_details}
\end{figure*}

{In this section, we present the details of how to generate sample selection masks $\cy{\boldsymbol{W}_{l/e}^{(d)}}$ to weigh losses. As depicted by \cref{fig:meth:cromss_details}, the selection mask generation process is mainly composed of three steps: confidence mask generation, cross-modal confidence enhancement, and thresholding. The second step, cross-modal confidence enhancement, is the same for both losses. In the first and third steps, we estimate the confidence values and apply different thresholding strategies for the two kinds of losses separately. We are giving the details of each step in the following.
}

{In the first step, we generate the label-based and the entity-based confidence masks for the segmentation and the consistency losses, respectively. The segmentation loss is used to supervise the model to predict the pixel-wise classification results of an image into given classes. It quantifies the difference between the predicted segmentation map and the ground truth. In our case, we don't have the real ground truth at hand. The labels we used for optimization contain noise or wrong annotations. We expect only the correctly assigned labels to contribute to the loss. That is, the confidence values should reflect the reliability of the labels. To this end, we utilize the softmax outputs at the given annotation class for each sample to generate the label-based confidence mask $\cy{\boldsymbol{F}_{l}}$ as follows,
\begin{equation} \label{eq:conf:lbl}
    f_{l,i}^{(d)}=p_{ic=y_i}^{(d)}
\end{equation}
where $f_{l,i}\in \cy{\boldsymbol{F}_{l}}$ is the estimated label-based confidence score at the $i$th pixel, $y_i$ is the given noisy label class.
}

On the other hand, the consistency loss, in the absence of labels, encourages the models to extract similar features from both modalities for the same scene. We hope the models can learn more from each other's reliable features, namely, the less ambiguous predictions. To achieve this, we estimate entity-based confidence mask $\cy{\boldsymbol{F}_{e}^{(d)}}$ using the entropy $H^{(d)}_i\in\mathbb{R}^+$ of the softmax output vectors as follows,
\begin{equation} \label{eq:conf:entity}
\small
    {
    f_{e,i}^{(d)} = 1-H_i^{(d)}/\text{K} = 1+\frac{1}{\text{K}}\sum_cp_{ic}^{(d)}\log p_{ic}^{(d)}.
    }
\end{equation}
where $f_{e,i}\in \cy{\boldsymbol{F}_{e}}$ is the estimated entity-based confidence score at the $i$th pixel, \cy{$\text{K}=\log \text{C}$} is the upper bound of the entropy $H_i\in[0,\cy{\text{K}}]$ when $p_{ic}=1/$\cy{C} for $c=1,\dots,\cy{\text{C}}$, i.e., equal distribution yields maximum entropy.

Then, we leverage cross-modal information to enhance the confidence masks of individual modalities before generating the final selection mask. Observations reveal that while a model from modality $d=1$ may exhibit high confidence in a label $y$ or prediction \cy{$\boldsymbol{p}$}, another model from modality $d=2$ might assign a low probability to $y$ or express less confidence in \cy{$\boldsymbol{p}$}. This discrepancy arises due to differences in modality characteristics and the inherent randomness in identical training processes. To address this, we reinforce consistent preferences between the two modalities and penalize divergent outcomes in the enhanced confidence masks by
\begin{equation}
\small
    \cy{\boldsymbol{F'}_{l/e}^{(d)}}={\tfrac{1}{2}\left(\cy{\boldsymbol{F}_{l/e}^{(d)}}+\cy{\boldsymbol{F}_{l/e}^{(1)}}\cy{\boldsymbol{F}_{l/e}^{(2)}}\right)}
    =\tfrac{1}{2}\cy{\boldsymbol{F}_{l/e}^{(d)}}\left(1+\cy{\boldsymbol{F}_{l/e}^{(d')}}\right),
\end{equation}
where the factor $\cy{\boldsymbol{F}_{l/e}^{(1)}\boldsymbol{F}_{l/e}^{(2)}}$ serves to magnify the selection probabilities of the samples exhibiting high confidences for both modalities, while diminishing cases where both modalities agree on low confidence scores, and $d'=\{1,2\}/d$. When $\cy{\boldsymbol{F}^{(1)}}$ and $\cy{\boldsymbol{F}^{(2)}}$ values vastly differ, or both exhibit values close to 1/2, the selection probabilities tend to stay closer to $3/8\approx1/2$.\footnote{
Define $f^\pm=1/2\pm\epsilon$ with $\epsilon\in[0,1/2]$, it follows for the quantity $f'^\pm=\tfrac{1}{2}\left(f^\pm+f^+f^-\right)=3/8\pm\epsilon/2+\mathcal{O}\left(\epsilon^2\right)$ such that $f'^\pm\approx3/8\pm \epsilon/2$ with $f'^\pm\approx5/8$ or $1/8$ when $\epsilon=1/2$ (the values vastly differ), and $f'^\pm\approx3/8$ when $\epsilon=0$ (both values are equal to $1/2$), implying $f^\pm\in[0,1]\to f'^\pm\in[1/8,5/8]$.
}

Finally, we apply thresholding to the enhanced confidence masks to derive sample selection masks for loss weighing, with $\alpha$ and $\gamma$ denoting the two hyperparameters used for the segmentation and the consistency losses, respectively. 
Specifically, we define $\alpha \in [0,1]$ as the selection ratio to pick out the most reliable labels. We sort the label-based confidence values in the descending order, and assign the weights to the samples with the $\alpha$ largest confidence values to 1. Notice that our goal is to pretrain the encoders for segmentation tasks. The encoders themselves can combat label noise to some extent \cite{liu_task_2024}. As a result, we use soft weights ranging from 0 to 1 for those unselected ones to avoid the removal of valuable samples with hard weights of 0. The thresholding for $\cy{\boldsymbol{F'}_{l}^{(d)}}$ is formulated as
\begin{equation}\label{eq:conf:select_label}  
    w_{l,i}^{(d)}(\alpha) = \min\left[1,{{f'}^{(d)}_{l,i}}/{f'}_{l}^{(d)}(\alpha)\right],
\end{equation}
where $\cy{\boldsymbol{W}_{l}}$ is the label-based sample selection mask with $w_{l,i}\in \cy{\boldsymbol{W}_{l}}$, the quantity ${f'}_l^{(d)}(\alpha)$ is the $\lfloor\alpha \cy{\text{N}}\rfloor$-th largest value\footnote{%
For practical purposes, it turned out that implementing the $\alpha$-dependent thresholding better performs when applied per class, that is: we utilize $\lfloor\alpha N_c\rfloor$ with $\cy{\text{N}}=\sum_c N_c$ where $N_c$ represents the number of samples per class $c$. Correspondingly, each ${f'}^{(d)}_{l,i}$ gets divided by a class-specific $w_l(\alpha)$ whenever the index $i=1,\dots, \cy{\text{N}}$ denotes a pixel of class $c=y_i$.
} in $\cy{\boldsymbol{F'}_l^{(d)}}$. Models' initial predictions are generally less accurate in the first couple of training epochs. Therefore, we set $\alpha$ gradually decreasing from 1 to the final target selection ratio $\alpha_0>0$ in the first $n_{s}$ epochs.

The consistency loss uses the predicted class distributions from another modality as the target data, which can be regarded as a kind of soft label. In this sense, we keep the soft weight format for the entity-based selection masks. Like \cref{eq:conf:select_label}, we utilize a moderator $\gamma$ to mitigate the negative impact of poor estimations in the early training stage. The entity-based selection mask $\cy{\boldsymbol{W}_e}$ is actually a weighted sum between 1 and the entity-based confidence mask:
\begin{equation}
    w_{e,i}^{(d)}(\gamma) = (1-\gamma)+\gamma{f'}^{(d)}_{e,i},
\end{equation}
where $w_{e,i}\in \cy{\boldsymbol{W}_e}$, and $\gamma$ ramps up from 0 to 1 in the first $n_s$ epochs.

\subsection{Spatial-temporal label smoothing} \label{sec:meth:smooth}

As shown in \cref{fig:meth:cromss} (right), we integrate a spatial-temporal label smoothing technique into CromSS to mitigate overconfidence in model predictions. The classic label smoothing technique replaces the one-hot label vector with the weighted averaging of it and the class uniform distribution \cite{szegedy_rethinking_2016}. In image segmentation, spatial correlations may get exploited for smoothing: Weighted averaging in space is another option to improve label smoothing. Additionally, our pretraining dataset SSL4EO-S12@NoL contains 4 timestamps of a given geospatial area (patch) with S1 and S2 data from different seasons. Therefore, we suggest enhancing label smoothing with spatial-temporal information for image segmentation as follows:
\begin{equation} \label{eq:meth:ls}
\small
    \cy{\boldsymbol{Z'}(\tau)} =\left(1-\beta-\mu\right)\cy{\boldsymbol{Z(\tau)}}+\beta \cy{\mathbf{U}}+\mu\left(\cy{\boldsymbol{M}_s(\tau)}+\cy{\boldsymbol{M}_t}\right)/2,
\end{equation}
where $\cy{\boldsymbol{Z}(\tau)}$ is the one-hot-encoded noisy label mask from a randomly selected season at timestamp $\tau$ in the data augmentation stage. The hyperparameters $\beta$ and $\mu$ control the proportions of the uniform class distribution $\cy{\mathbf{U}=\{1/\text{C}\}^{\text{H}\times \text{W}\times \text{C}}}$ and the spatial-temporal smoothing masks. 
The spatial smoothing mask $\cy{\boldsymbol{M}_{s}}$ is generated with Gaussian kernel $\cy{\mathbf{G}}$ convolved with $\cy{\boldsymbol{Z}(\tau)}$,
\begin{equation}
\small
    \cy{\boldsymbol{M}_s(\tau)} = \cy{\boldsymbol{Z}(\tau)}\ast \cy{\mathbf{G}}.
\end{equation}
The temporal smoothing mask $\cy{\boldsymbol{M}_t}$ is obtained with $\cy{\mathbf{G}}$ applied to the temporally averaged label mask from 4 seasons $\left\langle \cy{\boldsymbol{Z}}\right\rangle_t=\frac{1}{4}\sum_{t'\in\{\tau,\tau',\tau'',\tau'''\}}{\cy{\boldsymbol{Z}(t')}}$, i.e.,
\begin{equation}
\small
    \cy{\boldsymbol{M}}_t = \left\langle \cy{\boldsymbol{Z}}\right\rangle_t\ast \cy{\mathbf{G}}.
\end{equation}

\section{Experiments} \label{sec:exp}

We tested the proposed CromSS methodology by pretraining ResNet-50 encoders nested in U-Nets and transferring them to 3 segmentation downstream tasks. In the following,  we present the settings of our experiments in \cref{sec:exp:setup}, ahead of the detailed results obtained for the downstream tasks compared to state-of-the-art pretrained models in \cref{sec:exp:downstream}. {We explore the roles of each part in CromSS in \cref{sec:exp:ablation}.} Some insights for the pretraining stage are given in \cref{sec:exp:pre-training}. Finally, we discuss lessons learned when generating features by different pretraining methods in \cref{sec:exp:feat}. 

\subsection{Setups} \label{sec:exp:setup}
In the pretraining stage, we train U-Nets with an Adam optimizer \cite{kingma_adam_2017} and an initial learning rate of $5\cdot10^{-3}$. We employed the \texttt{ReduceLROnPlateau} scheduler to dynamically adjust the learning rate. It cuts in half the learning rate when the validation loss is not decreasing over 30 consecutive epochs. We randomly split off 1\% of the entire training set as the validation set. The selection proportion $\alpha$ for generating $\cy{\boldsymbol{W}_l^{(d)}}$ exponentially ramps down from 1 to $\alpha_0=0.5$ for the initial 80 epochs. The weighting factor $\gamma$ for $\cy{\boldsymbol{W}_e^{(d)}}$ is ramping up from 0 to 1 in parallel. The model input data (Sentinel-2 L1C) are normalized to [0,1] after discarding the values outside $\langle x\rangle\pm2\sigma(x)$ using the mean $\langle x\rangle$ and variance values $\sigma^2(x)$ of the SSL4EO-S12 dataset.
The data augmentation strategies we use before feeding the data into the U-Net, are:
\vspace{-.5em}
\begin{itemize}
\item random cropping to 256$\times$256, 
\item random flipping with 0.5 probability,
\item random rotating at probability 0.2,
\item random season selection which feeds the tiles of a randomly selected season.
\end{itemize}
\vspace{-.5em}
\cy{We implemented pretraining on SSL4EO-S12@NoL with 4 NVIDIA A100 GPUs running approx.\ 22 hours for 200 epochs. We configured a batch size of 128 for each GPU working in data-parallel mode, i.e.,\ backpropagated gradients are averaged over a total batch size of 512 data samples.}

When transferred to the downstream segmentation tasks, we embed the pretrained ResNet-50 encoders into different segmentation frameworks. We fine-tuned with the Adam optimizer given a combined loss of \texttt{CrossEntropy} and \texttt{Dice} at a learning rate of $5\cdot10^{-4}$ over 50 epochs. 
Aside from the two downstream tasks in NoLDO-S12, we also incorporate the DFC2020 dataset \cite{yokoya_2020_2019} as an additional downstream task, where we utilize the provided validation set comprising of 986 patches as the fine-tuning training set, and the original test set consisting of 5,128 patches for test. 
The three datasets (SSL4EO-S12@\{DW,OSM\} and DFC2020) represent various testing scenarios for pretrained models. SSL4EO-S12@DW involves a transfer learning task that mirrors the noisy label pretraining data SSL4EO-S12@NoL. However, SSL4EO-S12@OSM and DFC2020 differ in their class definitions from SSL4EO-S12@NoL. Moreover, DFC2020 is characterized by dense annotations, whereas the other two datasets contain unlabeled areas within each patch. The latter requires the use of binary masks to calculate the loss omitting unlabelled pixels.

For DFC2020, SSL4EO-S12@DW, and SSL4EO-S12@OSM, we utilize PSPNets \cite{zhao_pyramid_2017}, DeepLabV3+ \cite{chen_encoder-decoder_2018}, and FPN \cite{lin_feature_2017} as frameworks to nest ResNet-50 backbones, respectively\footnote{Models are constructed with the \texttt{segmentation-models-pytorch} package available at \url{https://github.com/qubvel/segmentation_models.pytorch}.}. The batch size for the three artificial neural networks we set to 32, 56, 32. For data augmentation we employed the random flip with a rate of 0.5 and the random rotation with a rate of 0.2. Besides, random cropping was also included for the SSL4EO-S12@DW and SSL4EO-S12@OSM datasets to resize the training inputs from $510\times$510 as 496$\times$496 and 480$\times$480. 
{We compare CromSS with 7 state-of-the-art self-supervised learning (SSL) pretrained models in Earth observation including DINO \cite{caron_emerging_2021}, MoCo \cite{chen_improved_2020}, DeCUR \cite{wang_decur_2023}, scaleMAE \cite{reed_scale-mae_2023}, satMAE \cite{cong_satmae_2022}, satMAE++ \cite{noman_rethinking_2024}, and DOFA \cite{xiong_neural_2024}\footnote{The pretrained weights with DINO and MoCo were provided by \cite{wang_ssl4eo-s12_2023} on the SSL4EO-S12 dataset.}. DINO, MoCo, and DeCUR provide pretrained ResNet-50 backbones, while the rest are ViT-large models. We fine-tuned UperNets \cite{Xiao_2018_ECCV} for ViT backbones with a learning rate of $10^{-4}$. DeCUR is a multi-modal extension of Barlow Twins that explicitly distinguishes inter- and intra-modal embeddings through multimodal redundancy reduction. ScaleMAE implements a scale-aware masking strategy for geospatial representation learning, yet only takes RGB bands as inputs. SatMAE and satMAE++ encode multi-spectral data (taking 10 bands of S2 excluding B1, B9, B10) as groups of bands with distinct spectral positional encodings. SatMAE++ additionally performs multi-scale pretraining. DOFA adaptively integrates various data modalities into a single framework by encoding wavelength information at the input end. In addition, we also compare CromSS with a supervised pretrained ResNet-50 backbone by Satlas \cite{bastani_satlaspretrain_2023}, which was derived on the SatlasPretrain dataset in a multi-task fashion with separate heads ---however, we took 9 bands of S2 (excluding B1, B8a, B9, B10) as inputs following \cite{bastani_satlaspretrain_2023}. We also took 9 bands for DOFA since it was pretrained on the SatlasPretrain dataset. For experiments with S1 data, we compare CromSS pretrained weights with the MoCo, DeCUR, and DOFA weights. 
}

{We implemented noisy label pretraining with CromSS for two cases of S2 data: taking either all the 13 bands or the 9 high-resolution bands following Satlas as inputs. We report the results of both cases for S2 while only reporting those pretrained along with all S2 bands for S1 in the downstream tasks. To differentiate between the two multi-modal architectures, we append \texttt{midF} and \texttt{lateF} to CromSS to represent the middle and late fusion.
As metrics, we picked overall accuracy (OA), average accuracy (AA), mean intersection over union (mIoU), and mean F1-scores for each class (mF1). The results listed below stem from 3 repeated runs of each downstream setup.
}

\subsection{Transfer learning results} \label{sec:exp:downstream}

\begin{table*}[htp]
\caption{{Transfer learning results (\%) on the DFC2020 dataset from PSPNet with ResNet-50 backbones and UperNet with ViT-large backbones. 3B\slash9B\slash10B\slash13B in brackets indicate the number of S2 bands as inputs for each model. \textit{Frozen} and \textit{Fine-tuned} represent no updates of encoders and optimizing encoders along with decoders in the transfer learning setting. The best results are highlighted in bold. The annotations are the same in the following tables.}}
\centering
\scriptsize
\label{tab:exp:ftdfc}
\begin{tabular}{c|c|c|cccc|cccc}
\hline\hline
\multirow{2}{*}{Inputs} & \multicolumn{2}{c|}{Encoder}                        & \multicolumn{4}{c|}{Frozen}                                                                            & \multicolumn{4}{c}{Fine-tuned}                                                                        \\
\cline{2-11}
& Backbone &   Weights (\#S2 bands)        & OA                      & AA                      & mIoU                    & mF1                     & OA                      & AA                      & mIoU                    & mF1                     \\
\hline\hline
\multirow{6}{*}{S1} & ViT-Large                  & DOFA                       & 49.15$\pm$1.89          & 39.59$\pm$1.40          & 27.51$\pm$0.72          & 36.74$\pm$1.13          & 57.41$\pm$1.44	& 43.74$\pm$1.37	& 31.12$\pm$1.43	& 41.35$\pm$1.80         \\
                     \cline{2-11}
  & \multirow{5}{*}{ResNet50}  & random                     & 53.89$\pm$0.56          & 40.65$\pm$0.38          & 28.96$\pm$0.16          & 38.74$\pm$0.22          & 52.03$\pm$0.56          & 42.02$\pm$0.34          & 28.02$\pm$0.23          & 37.45$\pm$0.29          \\
                     &                            & MoCo                       & 60.74$\pm$0.36          & \textbf{47.48$\pm$0.49} & 34.17$\pm$0.27          & 44.40$\pm$0.36          & 60.30$\pm$0.45          & 45.00$\pm$0.65          & 31.78$\pm$0.42          & 41.78$\pm$0.42          \\
                     &                            & DeCUR                      & 60.22$\pm$0.18          & 46.75$\pm$0.58          & 34.02$\pm$0.18          & 44.24$\pm$0.35          & 57.79$\pm$0.48          & 44.53$\pm$0.55          & 31.23$\pm$0.40          & 41.20$\pm$0.39          \\
                     \cline{3-11}
                     &                            & CromSS-midF          & 60.75$\pm$0.30          & 45.66$\pm$0.66          & 34.12$\pm$0.28          & \textbf{44.47$\pm$0.31}          & 60.07$\pm$0.50          & \textbf{46.44$\pm$0.63}          & 33.48$\pm$0.43          & \textbf{44.05$\pm$0.48}          \\
                     &                            & \textbf{CromSS-lateF}         & \textbf{61.65$\pm$0.34} & 44.80$\pm$0.09          & \textbf{34.20$\pm$0.10}         & 44.14$\pm$0.15          & \textbf{60.69$\pm$0.19}          & 45.75$\pm$0.36          & \textbf{33.63$\pm$0.12}          & 43.86$\pm$0.10          \\
\hline\hline
\multirow{11}{*}{S2} & \multirow{4}{*}{ViT-Large} & scaleMAE (3B)              & 63.49$\pm$0.26        & 53.31$\pm$0.29        &	38.51$\pm$0.20        &	49.90$\pm$0.12        &	68.42$\pm$0.46	        & 55.13$\pm$0.25        &	40.87$\pm$0.31        &	52.18$\pm$0.27         \\
                     &                            & DOFA (9B)          & 63.49$\pm$0.58          & 	48.79$\pm$0.20          & 	37.19$\pm$0.14          & 	48.87$\pm$0.44          & 	64.44$\pm$0.50          & 	49.50$\pm$0.11          & 	37.65$\pm$0.31          & 	48.91$\pm$0.59         \\
                     &                            & satMAE (10B)               & 67.02$\pm$0.41          & 51.91$\pm$0.04          & 40.13$\pm$0.16          & 51.52$\pm$0.31          & 66.47$\pm$0.21	& 50.51$\pm$0.18	& 39.02$\pm$0.16	& 50.48$\pm$0.66          \\
                     &                            & satMAE++ (10B)             & 68.41$\pm$0.06          & 53.15$\pm$0.77          & 40.96$\pm$0.27          & 52.93$\pm$0.32          & 67.21$\pm$0.17	& 51.44$\pm$0.04	& 39.90$\pm$0.26	& 51.32$\pm$0.66          \\
                     \cline{2-11}
 & \multirow{8}{*}{ResNet50}  & random (9B)                & 61.16$\pm$1.32          & 46.76$\pm$0.34          & 34.00$\pm$0.53          & 44.76$\pm$0.61          & 63.63$\pm$0.53          & 50.28$\pm$0.33          & 36.57$\pm$0.31          & 47.58$\pm$0.21          \\
                     &                            & random (13B)               & 56.11$\pm$0.55          & 44.63$\pm$0.22          & 31.26$\pm$0.14          & 41.89$\pm$0.19          & 58.83$\pm$0.87          & 45.89$\pm$0.36          & 33.63$\pm$0.24          & 44.14$\pm$0.29          \\
                     &                            & Satlas (9B)                & 65.91$\pm$0.35          & 51.03$\pm$0.41          & 38.74$\pm$0.33          & 50.27$\pm$0.46          & 64.76$\pm$0.35          & 52.02$\pm$0.61          & 37.85$\pm$0.13          & 49.23$\pm$0.21          \\
                     &                            & DINO (13B)                 & 64.65$\pm$0.12          & 48.53$\pm$0.30          & 37.65$\pm$0.13          & 48.77$\pm$0.15          & 63.75$\pm$0.70          & 49.80$\pm$1.06          & 37.04$\pm$0.55          & 48.95$\pm$0.73          \\
                     &                            & MoCo (13B)                 & 63.19$\pm$0.70          & 50.76$\pm$0.34          & 37.57$\pm$0.72          & 49.83$\pm$0.60          & 60.98$\pm$0.64          & 47.07$\pm$0.36          & 34.75$\pm$0.74          & 46.64$\pm$0.40          \\
                     &                            & DeCUR (13B)                & 63.02$\pm$0.70          & 47.99$\pm$0.57          & 36.82$\pm$0.37          & 47.90$\pm$0.44          & 64.69$\pm$1.03          & 50.90$\pm$0.97          & 38.02$\pm$0.94          & 50.10$\pm$1.10          \\
                     \cline{3-11}
                     & & CromSS-midF (9B)           & 66.00$\pm$0.15          & 55.81$\pm$0.07          & 40.31$\pm$0.40          & 52.59$\pm$0.17          & 69.00$\pm$0.26          & \textbf{57.07$\pm$0.07} & 42.17$\pm$0.13          & 54.02$\pm$0.24          \\
                     &                            & CromSS-lateF (9B)          & 68.95$\pm$0.67          & \textbf{56.01$\pm$0.41} & 41.69$\pm$0.78          & 53.61$\pm$0.94          & 68.82$\pm$0.17          & 55.75$\pm$0.10          & 41.51$\pm$0.28          & 53.34$\pm$0.20          \\
                     &                            & \textbf{CromSS-midF (13B)} & \textbf{69.56$\pm$0.32} & 54.61$\pm$0.25          & \textbf{42.52$\pm$0.11} & \textbf{54.12$\pm$0.16} & \textbf{70.41$\pm$0.35} & 55.11$\pm$0.30          & \textbf{43.21$\pm$0.16} & \textbf{54.91$\pm$0.23} \\
                     &                            & CromSS-lateF (13B)         & 67.96$\pm$0.49          & 52.62$\pm$0.64          & 41.38$\pm$0.34          & 53.25$\pm$0.49          & 68.49$\pm$0.49         & 53.19$\pm$0.59          & 41.91$\pm$0.47          & 53.59$\pm$0.39                  \\       
\hline\hline
\end{tabular}
\end{table*}

\begin{table*}[htp]
\caption{{Transfer learning results (\%) on the SSL4EO-S12@DW dataset from DeepLabv3+ with ResNet-50 and UperNet with ViT-large.}}
\label{tab:exp:ftdw}
\centering
\scriptsize
\begin{tabular}{c|c|c|cccc|cccc}
\hline\hline
\multirow{2}{*}{Inputs} & \multicolumn{2}{c|}{Encoder}                        & \multicolumn{4}{c|}{Frozen}                                                                            & \multicolumn{4}{c}{Fine-tuned}                                                                        \\
\cline{2-11}
& Backbone &   Weights (\#S2 bands)        & OA                      & AA                      & mIoU                    & mF1                     & OA                      & AA                      & mIoU                    & mF1                     \\                                                               
\hline\hline
\multirow{6}{*}{S1}     & ViT-Large                  & DOFA                       & 62.65$\pm$0.20          & 51.53$\pm$0.26          & 36.92$\pm$0.34          & 49.97$\pm$0.45          & 66.91$\pm$0.24          & 57.75$\pm$0.83          & 41.31$\pm$0.19          & 54.66$\pm$0.10          \\
                        \cline{2-11}
                        & \multirow{5}{*}{ResNet50}  & random                     & 56.84$\pm$0.53          & 44.67$\pm$0.55          & 29.85$\pm$0.45          & 41.10$\pm$0.58          & 65.77$\pm$0.28          & 54.98$\pm$0.41          & 39.80$\pm$0.36          & 53.04$\pm$0.50          \\
                        &                            & MoCo                       & 68.47$\pm$0.20          & 62.24$\pm$0.22          & 45.19$\pm$0.47          & 58.95$\pm$0.48          & 68.69$\pm$0.22          & 60.71$\pm$0.69          & \textbf{44.76$\pm$0.37} & \textbf{58.41$\pm$0.42} \\
                        &                            & \textbf{DeCUR}             & \textbf{69.38$\pm$0.16} & \textbf{62.49$\pm$0.38} & \textbf{45.53$\pm$0.08} & \textbf{59.36$\pm$0.21} & \textbf{69.10$\pm$0.18} & \textbf{61.67$\pm$1.62} & 44.72$\pm$0.46          & 58.37$\pm$0.45          \\
                        \cline{3-11}
                        &                            & CromSS-midF (13B)          & 66.25$\pm$0.16          & 54.29$\pm$0.23          & 39.63$\pm$0.12          & 52.17$\pm$0.26          & 68.04$\pm$0.54          & 59.75$\pm$1.19          & 43.71$\pm$0.42          & 57.21$\pm$0.58          \\
                        &                            & CromSS-lateF (13B)         & 66.03$\pm$0.14          & 53.05$\pm$0.12          & 39.63$\pm$0.14          & 52.22$\pm$0.18          & 69.00$\pm$0.44          & 60.47$\pm$1.01          & 44.00$\pm$0.82          & 57.29$\pm$0.94          \\
\hline\hline
\multirow{14}{*}{S2}    & \multirow{4}{*}{ViT-Large} & scaleMAE (3B)              & 75.89$\pm$1.23          & 72.60$\pm$1.67          & 57.37$\pm$1.33          & 68.95$\pm$1.40          & 78.80$\pm$1.50          & 73.57$\pm$2.12          & 59.89$\pm$2.07          & 72.67$\pm$1.78          \\
                        &                            & DOFA (9B)                 & 82.18$\pm$0.27          & 72.07$\pm$0.09          & 58.45$\pm$0.40          & 71.21$\pm$0.39          & 82.25$\pm$0.16         & 73.24$\pm$0.89          & 60.11$\pm$1.13          & 72.53$\pm$0.93          \\
                        &                            & satMAE (10B)               & 77.54$\pm$0.10          & 71.34$\pm$0.10          & 58.37$\pm$0.11          & 71.52$\pm$0.05          & 77.65$\pm$0.11          & 71.84$\pm$0.36          & 57.84$\pm$0.26          & 70.58$\pm$0.19          \\
                        &                            & satMAE++ (10B)             & 76.64$\pm$0.18          & 70.44$\pm$0.18          & 57.04$\pm$0.17          & 69.71$\pm$0.05          & 78.18$\pm$0.10          & 72.83$\pm$0.24          & 58.87$\pm$0.23          & 71.54$\pm$0.22          \\
                        \cline{2-11}
                        & \multirow{10}{*}{ResNet50}  & random (9B)                & 66.79$\pm$0.32          & 62.68$\pm$0.47          & 48.22$\pm$0.32          & 61.22$\pm$0.37          & 74.64$\pm$0.43          & 70.91$\pm$0.75          & 56.51$\pm$0.68          & 69.56$\pm$0.70          \\
                        &                            & random (13B)               & 64.80$\pm$0.47          & 60.30$\pm$0.20          & 46.21$\pm$0.33          & 59.11$\pm$0.30          & 73.36$\pm$0.21          & 68.58$\pm$0.49          & 54.72$\pm$0.25          & 67.76$\pm$0.26          \\
                        &                            & Satlas (9B)                & 75.15$\pm$0.10          & 70.26$\pm$0.15          & 56.39$\pm$0.12          & 69.29$\pm$0.24          & 77.92$\pm$0.67          & 72.59$\pm$0.86          & 59.17$\pm$1.02          & 71.87$\pm$1.00          \\
                        &                            & DINO (13B)                 & 77.64$\pm$0.16          & 72.55$\pm$0.53          & 59.36$\pm$0.46          & 71.98$\pm$0.50          & 77.81$\pm$0.23          & 71.36$\pm$0.58          & 58.05$\pm$0.26          & 70.38$\pm$0.39          \\
                        &                            & MoCo (13B)                 & 77.44$\pm$0.62          & 73.05$\pm$0.40          & 59.52$\pm$0.90          & 72.14$\pm$0.83          & 77.53$\pm$0.08          & 72.76$\pm$0.61          & 58.82$\pm$0.56          & 71.57$\pm$0.48          \\
                        &                            & DeCUR (13B)                & 76.37$\pm$0.18          & 71.28$\pm$0.51          & 57.45$\pm$0.15          & 70.15$\pm$0.28          & 77.20$\pm$0.22           & 72.05$\pm$0.30          & 58.54$\pm$0.39          & 71.32$\pm$0.33          \\
                        \cline{3-11}
                        &             & CromSS-midF (9B)           & 78.85$\pm$0.18          & \textbf{73.77$\pm$0.28} & \textbf{61.39$\pm$0.09} & \textbf{73.79$\pm$0.10} & 79.19$\pm$0.34          & 73.89$\pm$0.56          & 61.12$\pm$0.43          & 73.64$\pm$0.37          \\
                        &                            & \textbf{CromSS-lateF (9B)} & \textbf{78.98$\pm$0.21} & 73.39$\pm$0.60          & 60.93$\pm$0.60          & 73.37$\pm$0.52          & \textbf{79.93$\pm$0.14} & \textbf{74.03$\pm$0.45} & \textbf{62.05$\pm$0.24} & \textbf{74.18$\pm$0.20} \\
                        &                            & CromSS-midF (13B)          & 78.60$\pm$0.35          & 73.04$\pm$0.23          & 60.56$\pm$0.28          & 73.01$\pm$0.24          & 78.64$\pm$0.19          & 73.16$\pm$0.14          & 60.79$\pm$0.12          & 73.20$\pm$0.13          \\
                        &                            & CromSS-lateF (13B)         & 78.48$\pm$0.16          & 72.24$\pm$0.18          & 60.50$\pm$0.13          & 72.86$\pm$0.12          & 79.33$\pm$0.24          & 73.46$\pm$0.32          & 61.10$\pm$0.66          & 73.33$\pm$0.61    \\
\hline\hline
\end{tabular}
\end{table*}

We first present the transfer learning results on the DFC2020 dataset in \cref{tab:exp:ftdfc}. The proposed CromSS demonstrates a notable enhancement in the efficacy of pretrained encoders for remote sensing image segmentation, particularly for S2 multi-spectral data. On this dataset, middle fusion slightly outperforms late fusion. CromSS exhibits advantages in most metrics compared to other state-of-the-art pretraining methods, indicating the potential of using noisy labels for semantic segmentation in the pretraining setups. \cy{However, the improvement of CromSS for S1 radar data turned out to be less significant. We attribute this discrepancy to the different characteristics of two modalities involved in the pretraining task, that is, land cover classification in our case. S1, which we consider a \textit{weak} modality for land cover semantic segmentation, is more susceptible to label noise, thereby requiring stronger cross-modal strategies to combat it with the assistance of S2---research for future investigation.}

\begin{figure*}[htp]
    \centering
    \scriptsize
    \begin{tabular}{p{1.95cm}<{\centering}p{1.95cm}<{\centering}p{1.95cm}<{\centering}p{1.95cm}<{\centering}p{1.95cm}<{\centering}p{1.95cm}<{\centering}p{1.95cm}<{\centering}}
    \includegraphics[width=1.05\linewidth]{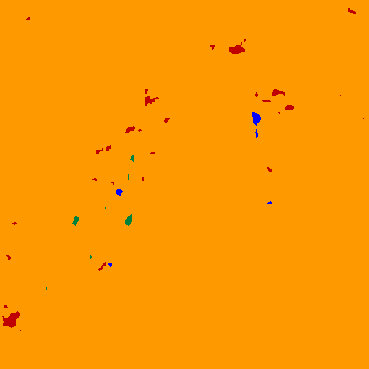} &
    \includegraphics[width=1.05\linewidth]{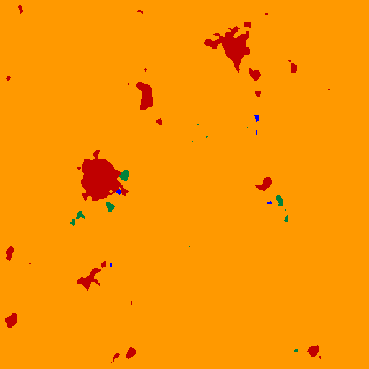} &
    \includegraphics[width=1.05\linewidth]{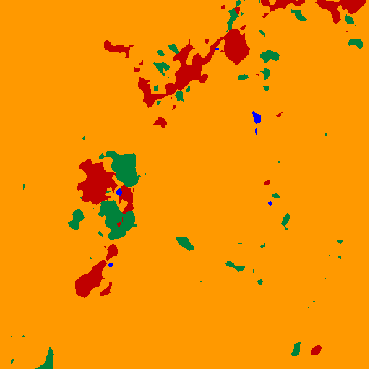} &
    \includegraphics[width=1.05\linewidth]{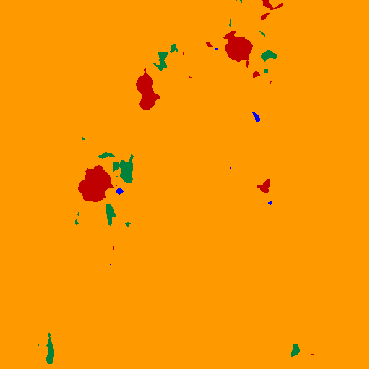} &
    \includegraphics[width=1.05\linewidth]{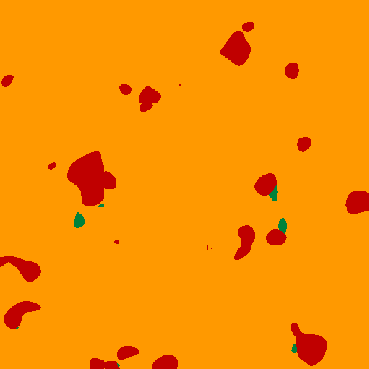} &
    \includegraphics[width=1.05\linewidth]{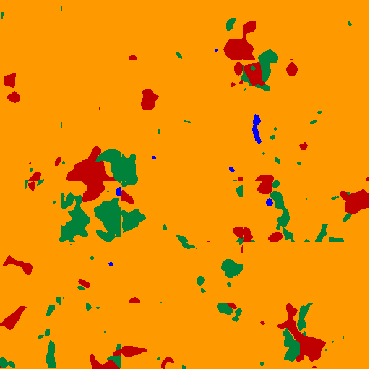} &
    \includegraphics[width=1.05\linewidth]{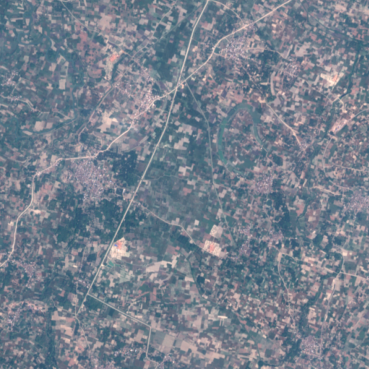} \\
    Satlas & DINO & MoCo & {DeCUR} & {scaleMAE} & {satMAE} & Optical \\
    \includegraphics[width=1.05\linewidth]{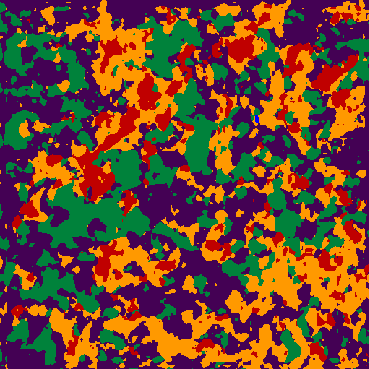} &
    \includegraphics[width=1.05\linewidth]{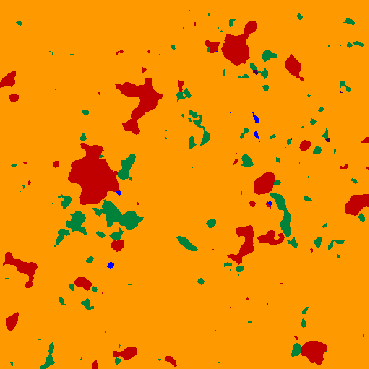} &
    \includegraphics[width=1.05\linewidth]{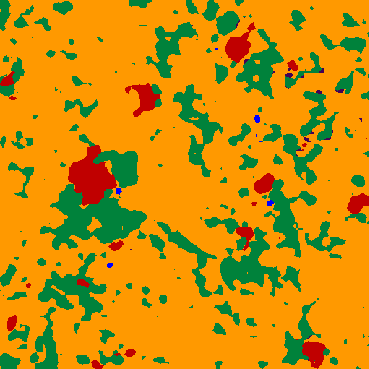} &
    \includegraphics[width=1.05\linewidth]{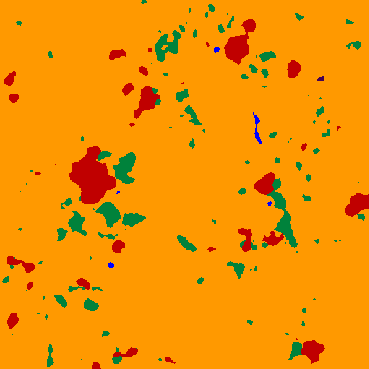} &
    \includegraphics[width=1.05\linewidth]{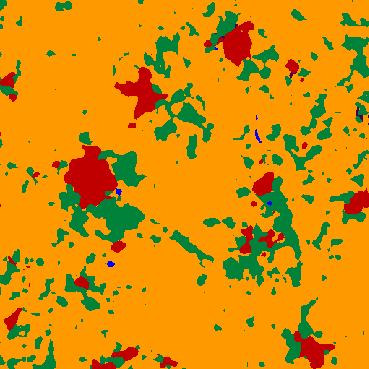} &
    \includegraphics[width=1.05\linewidth]{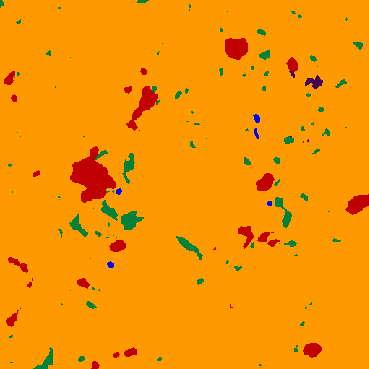} & 
    \includegraphics[width=1.05\linewidth]{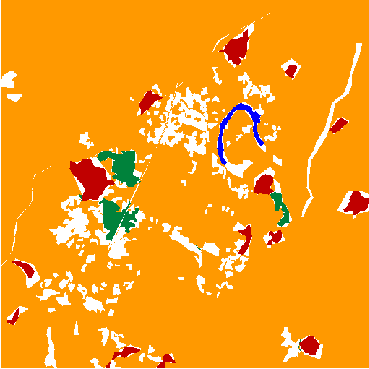} \\
    random & single-modal & midF & lateF & CromSS-midF & CromSS-lateF & GT (DW) \\
    \cline{1-6}
    \\
    \includegraphics[width=1.05\linewidth]{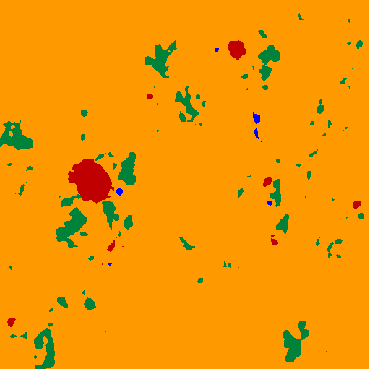} &
    \includegraphics[width=1.05\linewidth]{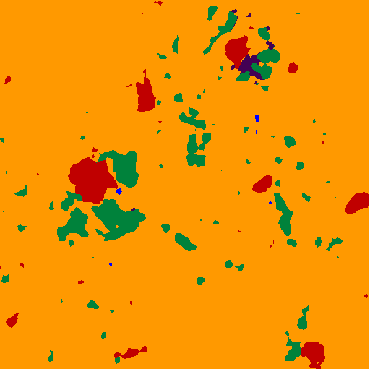} &
    \includegraphics[width=1.05\linewidth]{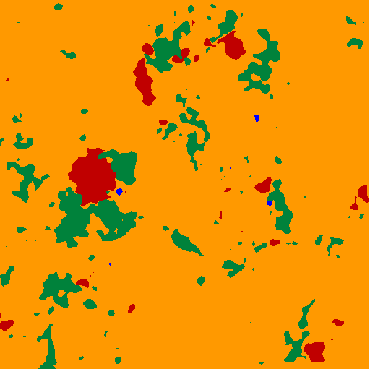} &
    \includegraphics[width=1.05\linewidth]{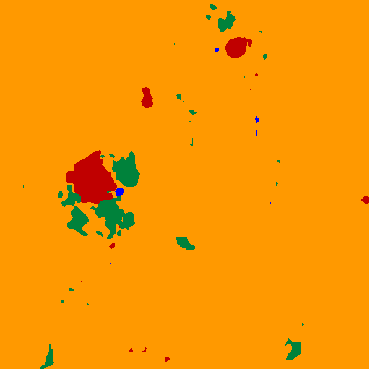} &
    \includegraphics[width=1.05\linewidth]{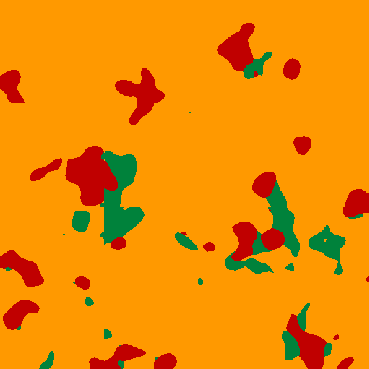} &
    \includegraphics[width=1.05\linewidth]{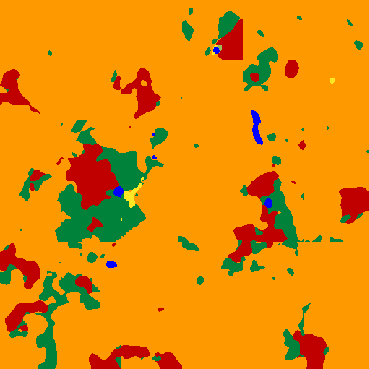} &
    \includegraphics[width=1.05\linewidth]{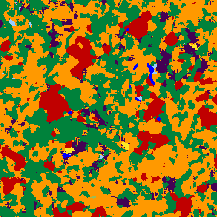} \\
    Satlas & DINO & MoCo & {DeCUR} & {scaleMAE} & {satMAE} & DW noisy labels  \\
    \includegraphics[width=1.05\linewidth]{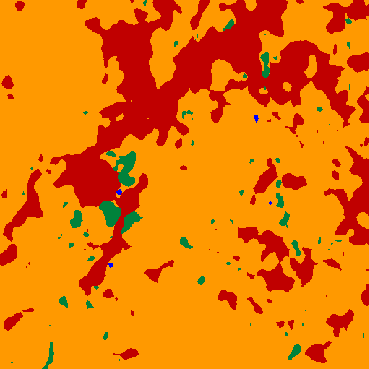} &
    \includegraphics[width=1.05\linewidth]{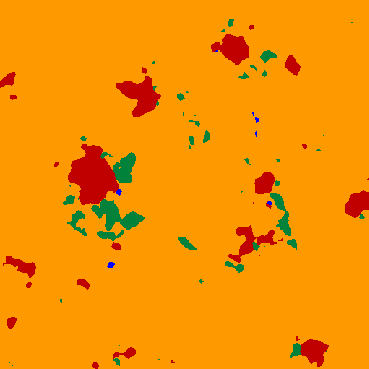} &
    \includegraphics[width=1.05\linewidth]{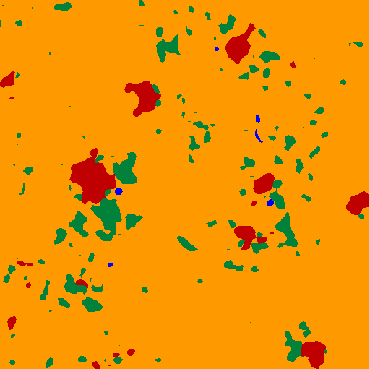} &
    \includegraphics[width=1.05\linewidth]{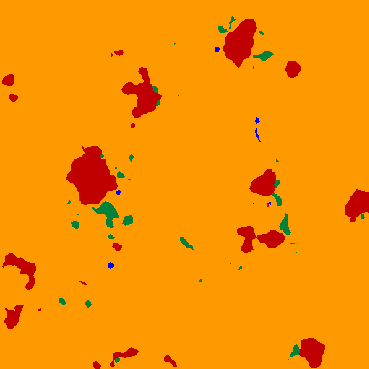} &
    \includegraphics[width=1.05\linewidth]{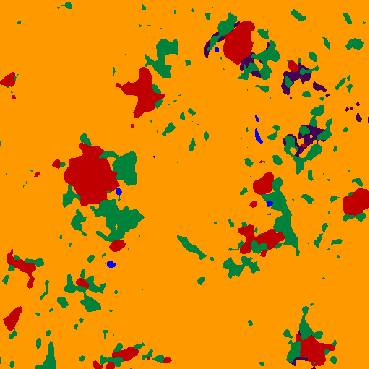} &
    \includegraphics[width=1.05\linewidth]{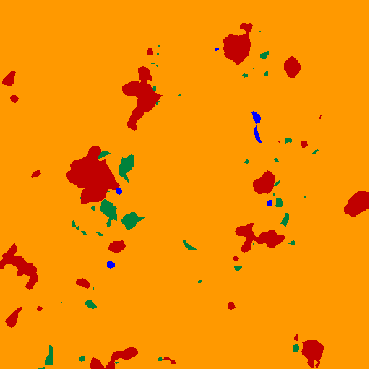} & \\
      random & single-modal & midF & lateF & CromSS-midF & CromSS-lateF & \\
    \end{tabular}
    \caption{{The segmentation maps obtained on the SSL4EO-S12@DW downstream dataset with S2 as inputs from UperNet with ViT-large pretrained by scaleMAE and satMAE, and the rest from PSPNets with ResNet-50 pretrained by different methods (upper: with frozen encoders, lower: with fine-tuned encoders).}}
    \label{fig:exp:dwsegmap}
\end{figure*}

We list the transfer learning results acquired from the SSL4EO-S12@DW downstream dataset in \cref{tab:exp:ftdw}. Notably, CromSS utilizing noisy labels for pretraining performs better than the other considered SSL methods for S2 data, showcasing a significant margin. Also, we found that ``more'' does not necessarily imply ``better''. For example, the models taking fewer bands as inputs tend to perform better, such as CromSS (9B) versus CromSS (13B). This is probably because the low-resolution bands introduce more ambiguity when classifying the 9 basic classes defined for this dataset. 
Regarding S1, the encoders pretrained with noisy labels showcase similar shortcomings, where DeCUR achieves the best results for S1, benefitting from the decoupled learning of two modalities in the pretraining stage. Nevertheless, after fine-tuning, the models pretrained with CromSS can obtain comparable results to DeCUR, demonstrating the effectiveness and potential of CromSS for S1. 
Furthermore, we present some segmentation maps generated with different pretrained encoders in \cref{fig:exp:dwsegmap} for visual comparison. Whether fixed or fine-tuned, encoders pretrained with CromSS can capture more intricate details than those obtained through other pretraining methods listed on the upper lines---particularly when the encoder weights are fixed.
\cy{However, we also observe that segmentation maps generated by the models pretrained on noisy labels tend to exhibit more pepper-and-salt noise. This is likely a side effect of the model's ability to capture finer details, which may also amplify minor inconsistencies or artifacts. While this noise can be seen as a drawback, it underscores the model's sensitivity to subtle features, which could be advantageous in applications where detail preservation is critical.}

\begin{table*}[htp]
\caption{{Transfer learning results (\%) on the SSL4EO-S12@OSM dataset from FPN with ResNet-50 and UperNet with ViT-large.}}
\label{tab:exp:ftosm}
\centering
\scriptsize
\begin{tabular}{c|c|c|cccc|cccc}
\hline\hline
\multirow{2}{*}{Inputs} & \multicolumn{2}{c|}{Encoder}                        & \multicolumn{4}{c|}{Frozen}                                                                            & \multicolumn{4}{c}{Fine-tuned}                                                                        \\
\cline{2-11}
& Backbone &   Weights (\#S2 bands)        & OA                      & AA                      & mIoU                    & mF1                     & OA                      & AA                      & mIoU                    & mF1                     \\     
\hline\hline
\multirow{6}{*}{S1}     & ViT-Large                  & DOFA               & 72.18$\pm$0.45          & 49.22$\pm$0.31          & 28.94$\pm$0.44          & 38.48$\pm$0.06          & 75.30$\pm$0.68          & 52.63$\pm$0.40          & 33.71$\pm$0.43          & 44.60$\pm$0.32          \\
                        \cline{2-11}
                        & \multirow{5}{*}{ResNet50}  & random             & 63.94$\pm$1.01          & \textbf{57.75$\pm$5.09} & 18.21$\pm$1.14          & 24.76$\pm$1.66          & 66.44$\pm$3.76          & 48.70$\pm$3.79          & 24.01$\pm$2.01          & 32.58$\pm$2.35          \\
                        &                            & MoCo               & 78.37$\pm$0.54          & 51.16$\pm$2.00          & 36.37$\pm$0.69          & 46.24$\pm$0.71          & 71.94$\pm$3.21          & 49.08$\pm$3.31          & 29.55$\pm$1.68          & 38.99$\pm$1.67          \\
                        &                            & \textbf{DeCUR}          & \textbf{80.10$\pm$0.28} & 51.13$\pm$0.67          & \textbf{38.16$\pm$0.57} & \textbf{48.25$\pm$0.59} & \textbf{79.83$\pm$0.53} & \textbf{58.91$\pm$0.69} & \textbf{36.55$\pm$0.29} & \textbf{46.92$\pm$0.23} \\
                        \cline{3-11}
                        &                            & CromSS-midF   & 76.69$\pm$0.58          & 48.11$\pm$0.78          & 33.61$\pm$0.57          & 43.83$\pm$0.66          & 77.72$\pm$1.18          & 51.46$\pm$2.12          & 36.15$\pm$0.97          & 46.65$\pm$0.85          \\
                        &                            & CromSS-lateF & 76.57$\pm$0.54          & 47.89$\pm$1.18          & 32.92$\pm$0.80          & 42.80$\pm$0.89          & 77.89$\pm$0.78          & 48.53$\pm$1.36          & 35.13$\pm$0.73          & 45.04$\pm$0.84          \\
\hline\hline
\multirow{14}{*}{S2}    & \multirow{4}{*}{ViT-Large} & scaleMAE (3B)      & 82.52$\pm$0.12          & 66.32$\pm$1.45          & 44.96$\pm$0.55          & 55.76$\pm$0.81          & 80.80$\pm$0.80          & 62.20$\pm$1.72          & 43.18$\pm$0.88          & 54.38$\pm$1.01          \\
                        &                            & DOFA (9B)         & 82.18$\pm$0.27         & 60.82$\pm$0.36          & 43.41$\pm$0.09          & 53.86$\pm$0.23          & 82.25$\pm$0.16          & 61.08$\pm$0.19          & 45.72$\pm$0.92          & 57.09$\pm$0.98          \\
                        &                            & satMAE (10B)       & 84.59$\pm$0.35          & \textbf{67.32$\pm$1.98}          & 48.07$\pm$1.55          & 59.32$\pm$2.19          & 84.26$\pm$0.06          & 69.03$\pm$0.27          & 46.20$\pm$0.47          & 56.74$\pm$0.55          \\
                        &                            & satMAE++ (10B)     & 83.65$\pm$0.13          & 65.48$\pm$0.36          & 45.60$\pm$0.07          & 56.71$\pm$0.10          & 84.24$\pm$0.27          & 69.51$\pm$0.79          & 46.19$\pm$1.04          & 56.92$\pm$1.25          \\
                        \cline{2-11}
                        & \multirow{10}{*}{ResNet50}  & random (9B)        & 70.89$\pm$0.40          & 60.04$\pm$3.57          & 28.35$\pm$0.53          & 36.71$\pm$0.57          & 67.38$\pm$7.02          & 50.80$\pm$3.17          & 27.98$\pm$3.90          & 36.39$\pm$4.29          \\
                        &                            & random (13B)       & 71.26$\pm$0.71          & 60.68$\pm$3.61          & 28.42$\pm$0.55          & 36.72$\pm$0.63          & 68.77$\pm$4.29          & 54.23$\pm$4.69          & 27.64$\pm$2.82          & 36.03$\pm$3.37          \\
                        &                            & Satlas (9B)        & 81.89$\pm$0.36          & 60.25$\pm$0.49          & 43.21$\pm$0.30          & 54.34$\pm$0.31          & 83.16$\pm$1.01          & 62.70$\pm$1.39          & 44.51$\pm$0.47          & 55.60$\pm$0.43         \\
                        &                            & DINO (13B)         & 83.46$\pm$0.30          & 60.29$\pm$1.77          & 42.77$\pm$0.73          & 53.68$\pm$0.93          & 78.53$\pm$2.30          & 57.87$\pm$2.85          & 37.54$\pm$2.58          & 47.60$\pm$2.63          \\
                        &                            & MoCo (13B)         & 83.87$\pm$0.43          & 61.82$\pm$1.28          & 46.79$\pm$0.75          & 56.93$\pm$0.80          & 80.67$\pm$1.25          & 62.75$\pm$2.84          & 41.99$\pm$1.93          & 52.74$\pm$2.07          \\
                        &                            & DeCUR (13B)        & 84.58$\pm$0.27          & 60.07$\pm$1.17          & 44.96$\pm$0.71          & 56.11$\pm$0.81          & 83.64$\pm$0.39          & 61.47$\pm$0.15          & 45.21$\pm$0.32          & 56.36$\pm$0.49          \\
                        \cline{3-11}
                        &   & CromSS-midF (9B)   & 83.98$\pm$0.14          & 62.27$\pm$2.20          & 46.58$\pm$0.49          & 57.31$\pm$0.56          & 84.57$\pm$0.63          & 64.01$\pm$3.74          & 47.69$\pm$0.97          & 58.76$\pm$1.04          \\
                        &                            & CromSS-lateF (9B)  & 83.64$\pm$0.30          & 60.55$\pm$0.25          & 45.72$\pm$0.36          & 56.61$\pm$0.42          & 85.25$\pm$0.12          & 64.45$\pm$0.27          & 49.38$\pm$0.16          & 60.62$\pm$0.36          \\
                        &                            & \textbf{CromSS-midF (13B)}  & \textbf{84.84$\pm$0.35} & 65.99$\pm$1.06          & \textbf{49.06$\pm$0.60} & \textbf{60.28$\pm$0.81} & \textbf{86.04$\pm$0.30} & \textbf{70.05$\pm$2.71} & \textbf{50.11$\pm$0.64} & \textbf{62.59$\pm$0.64} \\
                        &                            & CromSS-lateF (13B) & 84.75$\pm$0.43          & 66.63$\pm$2.48 & 47.38$\pm$0.79          & 58.49$\pm$0.85          & 85.36$\pm$0.37          & 65.84$\pm$1.87          & 49.38$\pm$0.66          & 60.67$\pm$0.56 \\
\hline\hline
\end{tabular}
\end{table*}

\begin{figure*}[htp]
    \centering
    \scriptsize
    \begin{tabular}{p{1.95cm}<{\centering}p{1.95cm}<{\centering}p{1.95cm}<{\centering}p{1.95cm}<{\centering}p{1.95cm}<{\centering}p{1.95cm}<{\centering}p{1.95cm}<{\centering}}
    \includegraphics[width=1.05\linewidth]{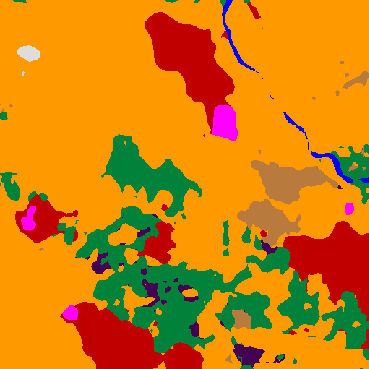} &
    \includegraphics[width=1.05\linewidth]{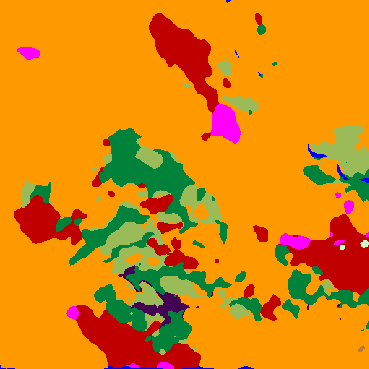} &
    \includegraphics[width=1.05\linewidth]{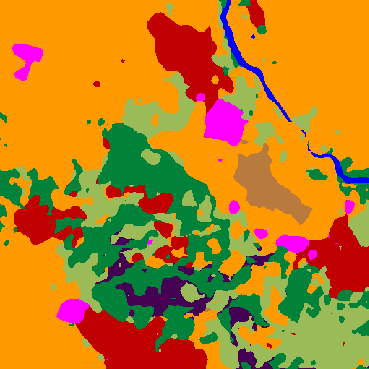} &
    \includegraphics[width=1.05\linewidth]{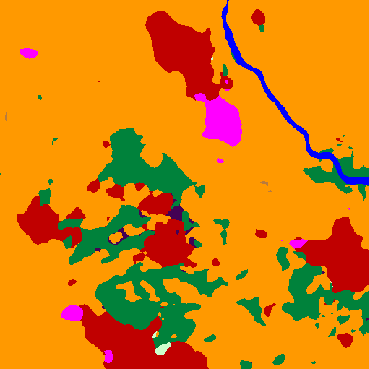} &
    \includegraphics[width=1.05\linewidth]{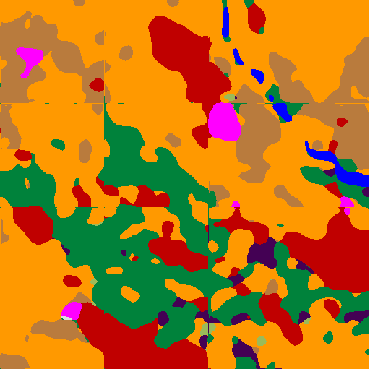} &
    \includegraphics[width=1.05\linewidth]{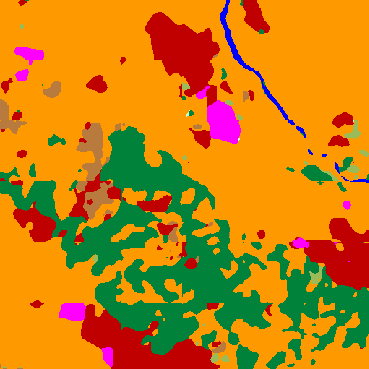} & 
    \includegraphics[width=1.05\linewidth]{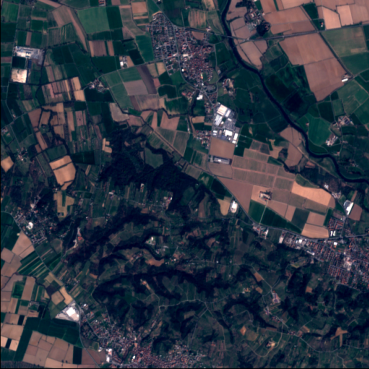} \\
    Satlas & DINO & MoCo & {DeCUR} & {scaleMAE} & {satMAE} & Optical  \\
    \includegraphics[width=1.05\linewidth]{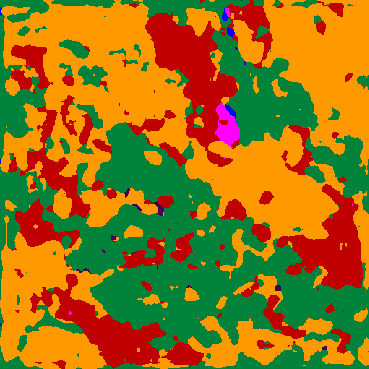} &
    \includegraphics[width=1.05\linewidth]{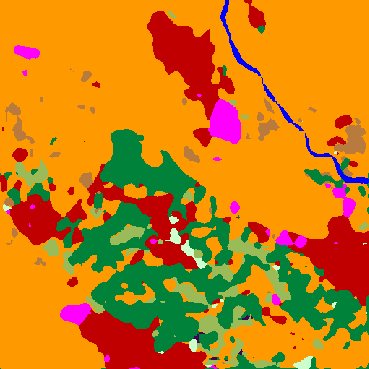} &
    \includegraphics[width=1.05\linewidth]{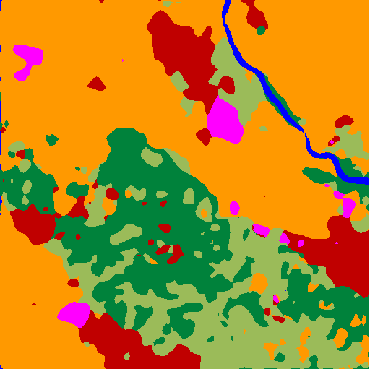} &
    \includegraphics[width=1.05\linewidth]{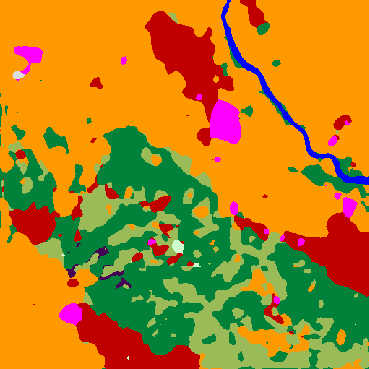} &
    \includegraphics[width=1.05\linewidth]{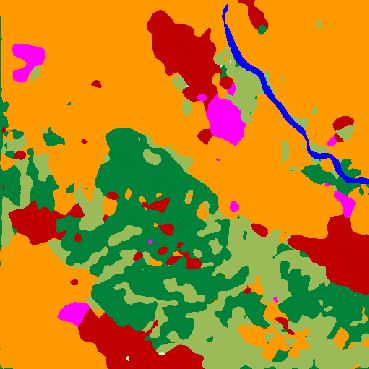} &
    \includegraphics[width=1.05\linewidth]{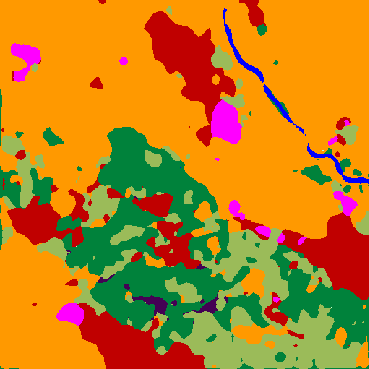} &
    \includegraphics[width=1.05\linewidth]{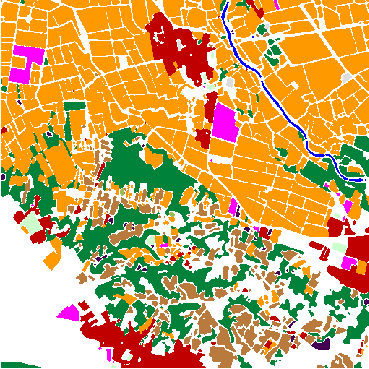} \\
    random & single-modal & midF & lateF & CromSS-midF & CromSS-lateF & GT (OSM) \\
    \cline{1-6}
    \\
    \includegraphics[width=1.05\linewidth]{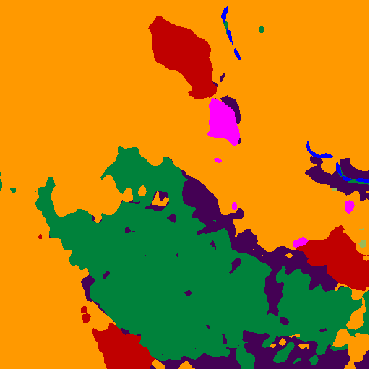} &
    \includegraphics[width=1.05\linewidth]{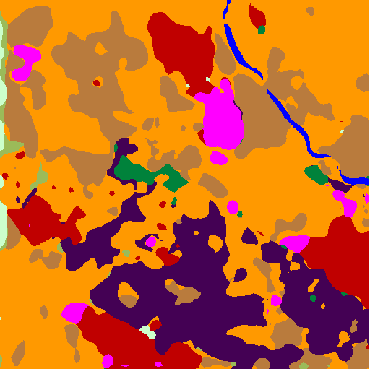} &
    \includegraphics[width=1.05\linewidth]{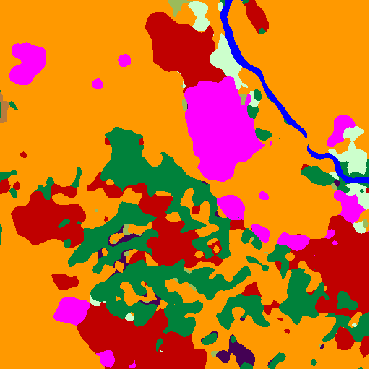} &
    \includegraphics[width=1.05\linewidth]{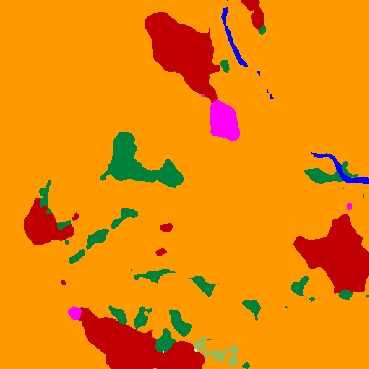} &
    \includegraphics[width=1.05\linewidth]{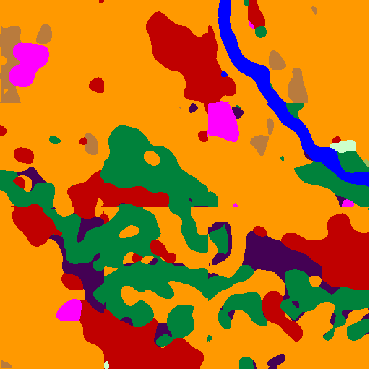} &
    \includegraphics[width=1.05\linewidth]{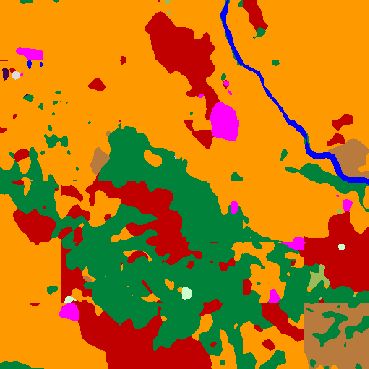} &
    \includegraphics[width=1.05\linewidth]{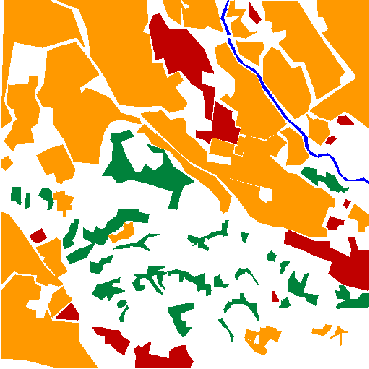} \\
    Satlas & DINO & MoCo & {DeCUR} & {scaleMAE} & {satMAE} &  GT (DW) \\
    \includegraphics[width=1.05\linewidth]{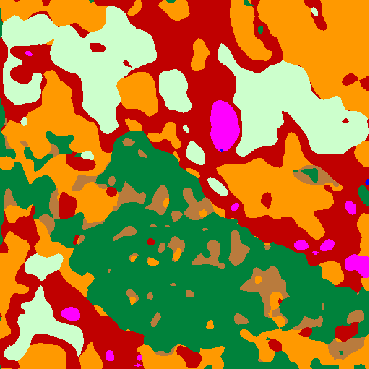} &
    \includegraphics[width=1.05\linewidth]{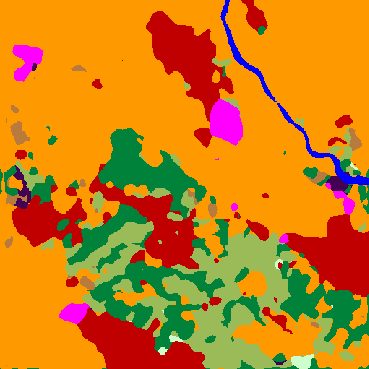} &
    \includegraphics[width=1.05\linewidth]{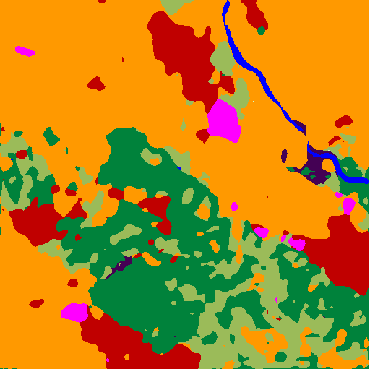} &
    \includegraphics[width=1.05\linewidth]{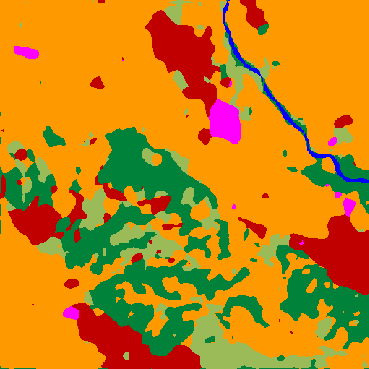} &
    \includegraphics[width=1.05\linewidth]{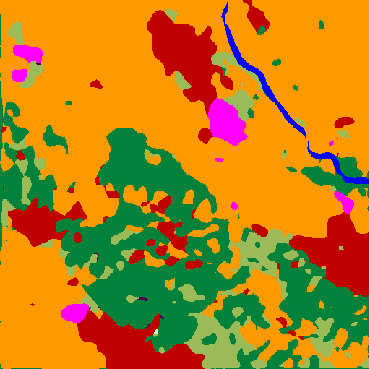} &
    \includegraphics[width=1.05\linewidth]{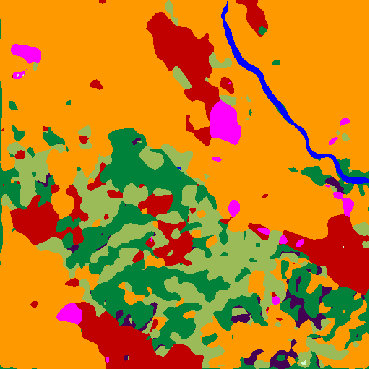} & \\
    random & single-modal & midF & lateF & CromSS-midF & CromSS-lateF & \\
    \end{tabular}
    \caption{{The segmentation maps obtained on the SSL4EO-S12@OSM downstream dataset with S2 as inputs from UperNet with ViT-large pretrained by scaleMAE and satMAE, and the rest from FPN with ResNet-50 pretrained by different methods (upper: with frozen encoders, lower: with fine-tuned encoders).}}
    \label{fig:exp:osmsegmap}
\end{figure*}

{Lastly, we present the results generated on the SSL4EO-S12@OSM downstream dataset in \cref{tab:exp:ftosm}. Consistent with the findings on the other two datasets, CromSS demonstrates superior performance compared to randomly initialized and SSL-pretrained models for S2 data. This dataset is challenging since it contains more fine-grained classes with fewer training samples than the other two downstream tasks. In this case, all the pretraining methods exhibit a significant advantage over random initialization, particularly our proposed CromSS. Besides, the models taking more bands as inputs perform better on this dataset since more complicated scenes require more information to parse. Moreover, when encoders are fine-tuned alongside decoders, the proposed methods exhibit less severe overfitting when dealing with very limited training samples. These observations align with the insights drawn from segmentation maps illustrated in \cref{fig:exp:osmsegmap}, where we also find that patch-based ViT models easily lead to oversmoothed segmentation maps, especially using a larger patch size such as scaleMAE with a patch size of 16 compared to satMAE with a patch size of 8. 
In contrast, the effectiveness of noisy label pretraining for S1 data requires further improvements compared to that for S2.}

\subsection{Ablation experiments} \label{sec:exp:ablation}

{In this section, we explore the roles of the components of CromSS, including multi-modal pretraining architectures, cross-modal sample selection, and spatial-temporal label smoothing, through a series of ablation experiments.
}

\begin{table*}[htp]
\caption{{Transfer learning results (\%) on the three downstream datasets obtained by ResNet-50 with different noisy label pretraining strategies: single-modal, multi-modal (midF and lateF), and multi-modal with sample selection (CromSS).}}
\centering
\scriptsize
\label{tab:exp:ablat-fuse}
\begin{tabular}{p{0.5cm}<{\centering}|p{1.5cm}<{\centering}||p{0.4cm}<{\centering}p{0.4cm}<{\centering}p{0.4cm}<{\centering}|p{0.4cm}<{\centering}p{0.4cm}<{\centering}p{0.4cm}<{\centering}||p{0.4cm}<{\centering}p{0.4cm}<{\centering}p{0.4cm}<{\centering}|p{0.4cm}<{\centering}p{0.4cm}<{\centering}p{0.4cm}<{\centering}||p{0.4cm}<{\centering}p{0.4cm}<{\centering}p{0.4cm}<{\centering}|p{0.4cm}<{\centering}p{0.4cm}<{\centering}p{0.4cm}<{\centering}}
\hline\hline
\multirow{3}{*}{Inputs} & \multirow{3}{*}{Weights} & \multicolumn{6}{c||}{DFC2020} & \multicolumn{6}{c||}{SSL4EO-S12@DW} & \multicolumn{6}{c}{SSL4EO-S12@OSM} \\
\cline{3-20}
 &   & \multicolumn{3}{c|}{Frozen}  & \multicolumn{3}{c||}{Fine-tuned} & \multicolumn{3}{c|}{Frozen}  & \multicolumn{3}{c||}{Fine-tuned} & \multicolumn{3}{c|}{Frozen}  & \multicolumn{3}{c}{Fine-tuned} \\
\cline{3-20}
&                            & OA                      & AA                      & mIoU                     & OA                      & AA                      & mIoU & OA                      & AA                      & mIoU                     & OA                      & AA                      & mIoU & OA                      & AA                      & mIoU                     & OA                      & AA                      & mIoU   \\
\hline\hline
\multirow{6}{*}{S1}     & random                           & 53.89          & 40.65          & 28.96          & 52.03         & 42.02         & 28.02   & 56.84          & 44.67     & 29.85    & 65.77          & 54.98         & 39.80   & 63.94          & 57.75          & 18.21          & 66.44          & 48.70          & 24.01      \\
                        & single-modal                 & 61.15          & 46.14 & 34.17 & 59.91          & 45.68          & 33.46        & 65.80          & 56.15          & 39.98          & 68.09          & 59.88          & 43.26 & 77.12        & 49.04        & 33.96      & 78.30 & 49.85 & 36.2   \\
                        \cdashline{2-20}
                        & midF                              & 60.00          & 44.16         & 33.29          & 59.99         & 45.69          & 33.13   & 67.04 & 54.73          & 40.30          & 67.89          & 57.68          & 42.39    & 77.21         & 49.53 & 34.10          & 78.44 & 51.94 & 36.88   \\
                        & lateF                             & 61.52 & 45.98          & 34.12            & 61.06 & 46.83 & 33.66   & 67.04 & 56.51 & 40.16          & 68.45         & 61.10 & 43.49  & 77.33 & 48.70          & 34.54 & 78.18          & 49.64          & 36.01   \\
                        \cdashline{2-20}
                        & CromSS-midF           & 60.75          & 45.66          & 34.12         & 60.07          & 46.44          & 33.48  & 66.25          & 54.29          & 39.63          & 68.04          & 59.75          & 43.71 & 76.69          & 48.11          & 33.61         & 77.72          & 51.46         & 36.15 \\
                        & CromSS-lateF          & 61.65 & 44.80          & 34.20          & 60.69          & 45.75          & 33.63 & 66.03          & 53.05          & 39.63         & 69.00 & 60.47          & 44.00 & 76.57          & 47.89          & 32.92          & 77.89          & 48.53          & 35.13 \\
\hline\hline
\multirow{6}{*}{S2}  & random                            & 56.11          & 44.63          & 31.26          & 58.83         & 45.89         & 33.63 & 64.80          & 60.30          & 46.21          & 73.36         & 68.58         & 54.72     & 71.26          & 60.68         & 28.42         & 68.77          & 54.23          & 27.64     \\
                     & single-modal                 & 66.36          & 52.52         & 40.27          & 67.47          & 53.22          & 41.15    & 78.46          & 72.54          & 59.30          & 78.36          & 72.37          & 59.56  & 83.94          & 64.87          & 45.29          & 83.72         & 64.72         & 46.68   \\
                     \cdashline{2-20}
                     & midF                              & 68.57          & 52.49          & 41.31          & 68.70          & 52.94          & 41.46   & 78.82 & 73.28 & 60.91 & 78.60          & 73.12          & 60.32    & 84.05          & 65.26         & 47.37          & 83.50          & 63.98          & 47.96   \\
                     & lateF                             & 68.88          & 53.58          & 41.30          & 68.42          & 53.24          & 41.40    & 78.69 & 72.93          & 60.29         & 78.71 & 73.80 & 61.39  & 84.73         & 64.42          & 47.62 & 84.43         & 66.50 & 48.89   \\
                     \cdashline{2-20}
                     & CromSS-midF  & 69.56 & 54.61          & 42.52 & 70.41 & 55.11          & 43.21 & 78.60          & 73.04          & 60.56          & 78.64          & 73.16          & 60.79  & 84.84 & 65.99 & 49.06 & 86.04 & 70.05 & 51.11   \\
                      & CromSS-lateF    & 68.93          & 52.62          & 41.35          & 68.37         & 53.19          & 41.83    & 78.48          & 72.24          & 60.50          & 79.33 & 73.46 & 61.10  & 84.75 & 66.63 & 47.38          & 85.36 & 65.84          & 49.38    \\
\hline\hline
\end{tabular}
\end{table*}

{We first test the effectiveness of the first two components. \cref{tab:exp:ablat-fuse} lists the transfer learning results of single-modal noisy label pretraining (S1/S2) and multi-modal pretraining without cross-modal sample selection (\texttt{midF} and \texttt{lateF} denote middle and late fusion), in comparison to random initialization and CromSS (multi-modal pretraining with cross-modal sample selection), along with corresponding segmentation maps illustrated in Figs. \ref{fig:exp:dwsegmap} and \ref{fig:exp:osmsegmap}. 
Notably, single-modal pretraining with noisy labels brings a significant performance gain compared to random initialization for both S1 and S2 data on three datasets. This margin is further amplified after introducing multi-modal learning into the noisy label pretraining framework (midF and lateF). These results suggest that multi-modal training can inherently aid models in alleviating the adverse impacts of label noise. However, the margin is smaller for S1 than that for S2. As mentioned above, S1 is a ``weak'' modality in the pretraining setup, where the learning is more easily affected by label noise, thus benefiting less from noisy labels. This further impacts the effectiveness of the sample selection module for S1 data.
On the contrary, we observe some further improvements for S2 data after integrating cross-modal sample selection into the multi-modal pretraining framework. However, CromSS fails to surpass midF and lateF on the SSL4EO-S12@DW dataset. We attribute this shortcoming to the following: Although the sample selection masks are designed to suppress label noise, they may block relevant information, such as boundaries between areas of two classes or fine details where many different classes co-exist in a small neighborhood of pixels. If sample selection masks block this information, the model performance degrades when training is continued with a limited number of exact labels under the same definitions (see \cref{fig:exp:selectionmask}, and we will analyze this issue in detail in \cref{sec:exp:pre-training}). Nevertheless, the performance of CromSS is still comparable to midF and lateF.
Remarkably, CromSS exhibits a significant advantage on the other two datasets. Interestingly, the partial loss of specific details during the pretraining stage contributes to enhanced generality when transferred to other datasets. The advantage is more significant in the middle fusion cases, implying that explicit constraint via decoder weight sharing can potentially boost sample selection. Moreover, when encoders are fine-tuned alongside decoders, CromSS shows less severe overfitting when dealing with limited training samples.
}

\begin{table}[htp]
\caption{{Transfer learning results (mIoU, \%) obtained using ResNet-50 pretrained by CromSS-midF under different label smoothing settings, where we report the scores with ``frozen/fine-tuned" encoders on the three downstream datasets.}}
\label{tab:exp:ablation-ls}
\centering
\scriptsize
\begin{tabular}{c|c|c|c|c}
\hline\hline
Spatial-temporal & Uniform & \multirow{2}{*}{DFC2020} & SSL4EO-S12 & SSL4EO-S12 \\
($\alpha$) & ($\mu$)   &     & @DW &  @OSM       \\
\hline\hline
0     & 0    & \textbf{43.26}/42.63 & \textbf{61.07}/61.26  & 47.24/49.80 \\
0     & 0.2  & 41.64/42.32          & 58.65/59.36           & 47.67/50.57 \\
\hdashline
0.2   & 0    & 42.3/\textbf{43.52}  & 60.98/\textbf{61.61}  & 46.98/49.49 \\
0.15  & 0.05 & 42.52/43.21          & 60.64/60.66           & \textbf{49.00}/\textbf{51.28} \\
\hline\hline
\end{tabular}
\end{table}


We then evaluate the effectiveness of the spatial-temporal label smoothing technique within the CromSS-midF framework. As shown in \cref{eq:meth:ls}, in addition to the commonly used class uniform term (weighed by $\mu$), we introduce a spatial-temporal prior term (weighed by $\alpha$), derived from 4-seasonal label masks at each location for label smoothing. 
{\Cref{tab:exp:ablation-ls} presents the transfer learning results obtained with various combinations of these two terms. The results indicate that the combined approach ($\alpha=0.15,\mu=0.05$) used in our method yields more stable performance across all datasets. While it may not outperform other settings on each dataset, it provides comparable results without significant degradation. In contrast, the plain uniform label smoothing strategy leads to a notable performance drop on nearly all datasets, and the pure spatial-temporal label smoothing strategy does not perform satisfactorily on the SSL4EO-S12@OSM dataset. This highlights the robustness and adaptability of our proposed spatial-temporal label smoothing technique across different data sources.} 

\subsection{Model performance in the pretraining stage} \label{sec:exp:pre-training}

\begin{figure*}[htp]
    \centering
    \scriptsize
    \begin{tabular}{cccc}
    \includegraphics[height=.17\linewidth]{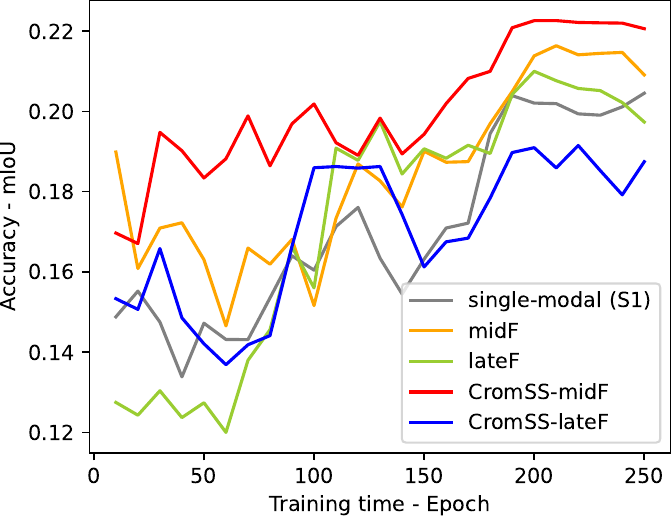} &
    \includegraphics[height=.17\linewidth]{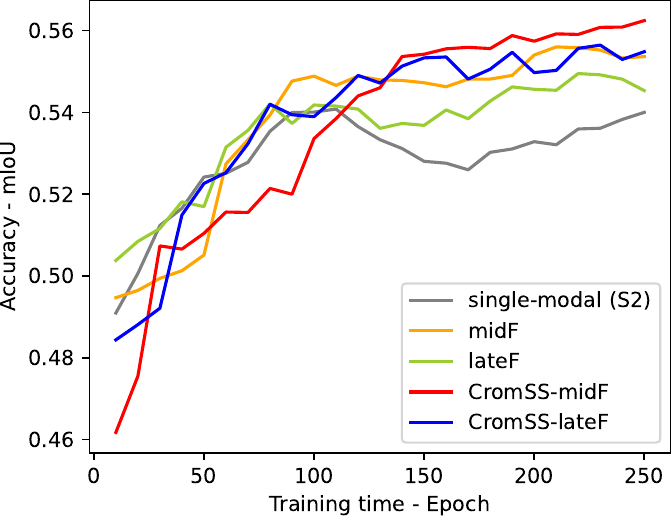} &
    \includegraphics[height=.17\linewidth]{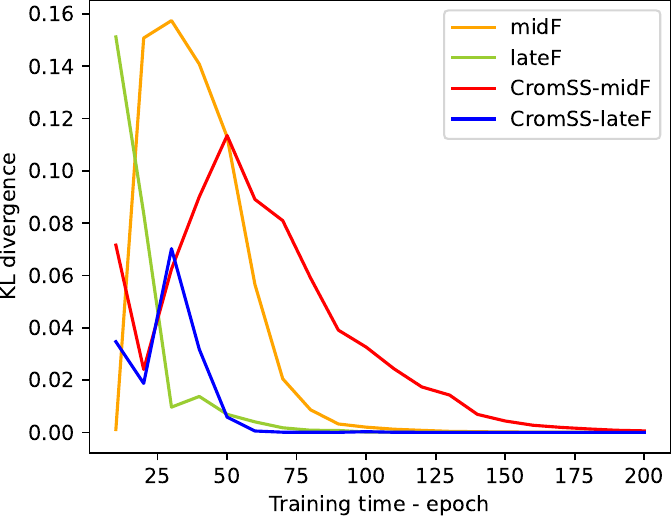} &
    \includegraphics[height=.17\linewidth]{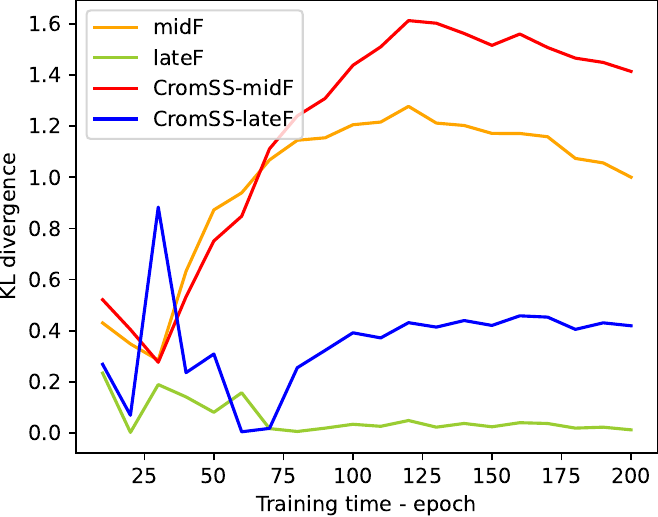} \\
    (a) pretrained model performance (S1) & (b) pretrained model performance (S2) & (c) feature statistics KL (encoder)  & (d) feature statistics KL (decoder)
    \end{tabular}
    \caption{The accuracy (mIoU) of the pretrained models versus training time for (a) S1 and (b) S2 data, and the similarity (KL divergence) of the feature means and standard deviations in the batch normalization layer between the single-modal and multi-modal pretraining for S2 data versus training time for (c) encoders and (d) decoders. The curves were obtained with an interval of 10 epochs, and smoothed by average moving with a window size of 5.}
    \label{fig:exp:pre-trainplots}
\end{figure*}

\begin{figure*}[htp]
    \centering
    \scriptsize
    \begin{tabular}{p{1.95cm}<{\centering}p{1.95cm}<{\centering}p{1.95cm}<{\centering}p{1.95cm}<{\centering}p{1.95cm}<{\centering}p{1.95cm}<{\centering}p{1.95cm}<{\centering}}
         \includegraphics[width=1.05\linewidth]{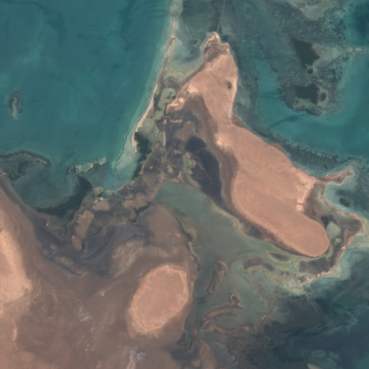} &
         \includegraphics[width=1.05\linewidth]{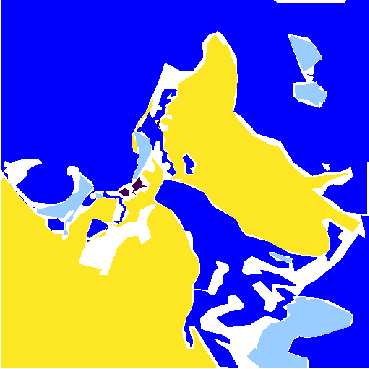} &
         \includegraphics[width=1.05\linewidth]{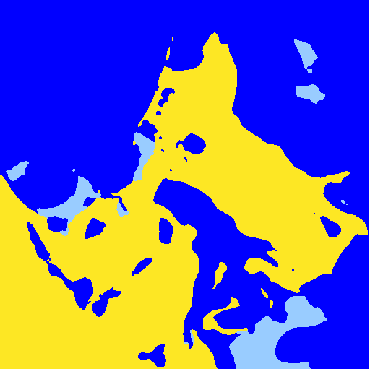} &  
         \includegraphics[width=1.05\linewidth]{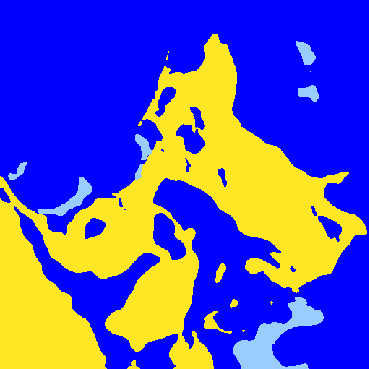} & 
         \includegraphics[width=1.05\linewidth]{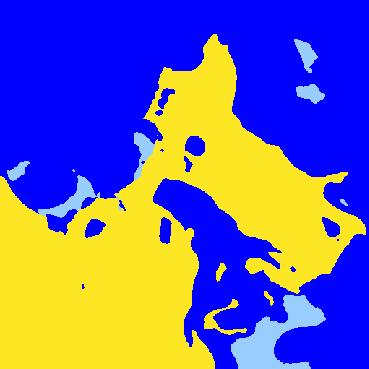} & 
         \includegraphics[width=1.05\linewidth]{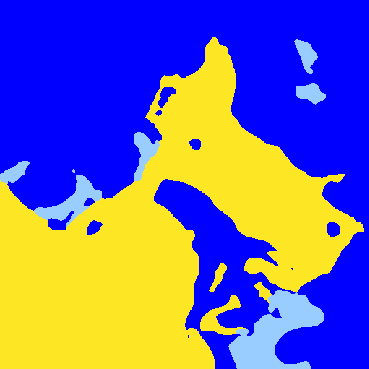} & 
         \includegraphics[width=1.05\linewidth]{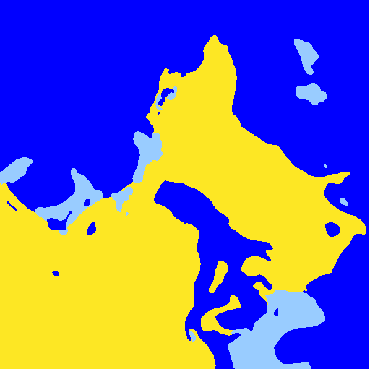} \\
         \includegraphics[width=1.05\linewidth]{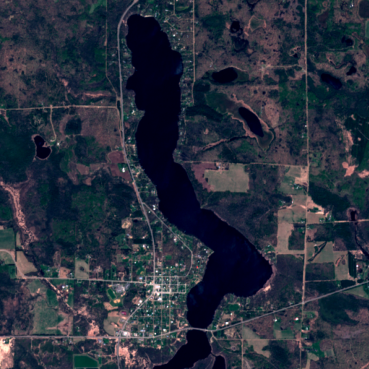} &
         \includegraphics[width=1.05\linewidth]{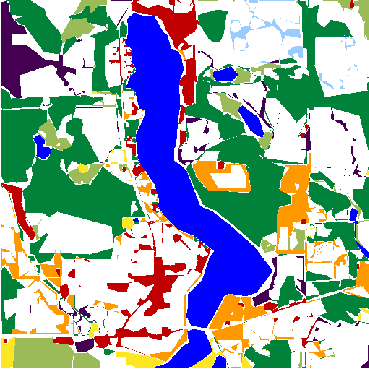} &
         \includegraphics[width=1.05\linewidth]{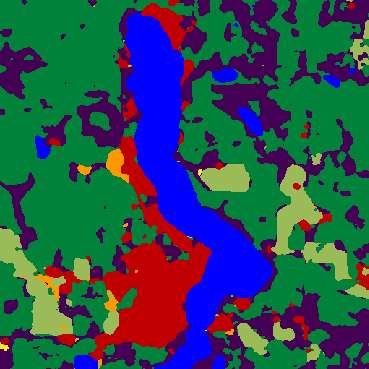} &  
         \includegraphics[width=1.05\linewidth]{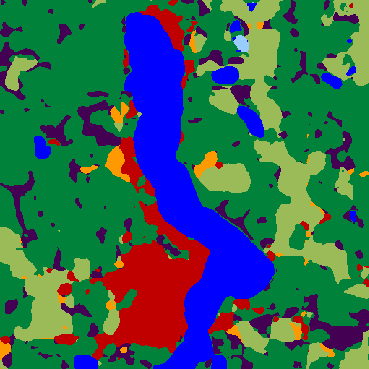} & 
         \includegraphics[width=1.05\linewidth]{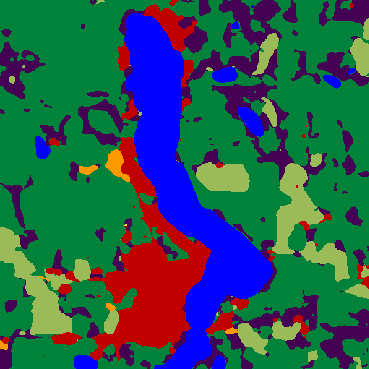} & 
         \includegraphics[width=1.05\linewidth]{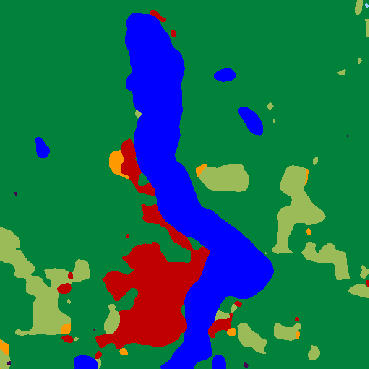} & 
         \includegraphics[width=1.05\linewidth]{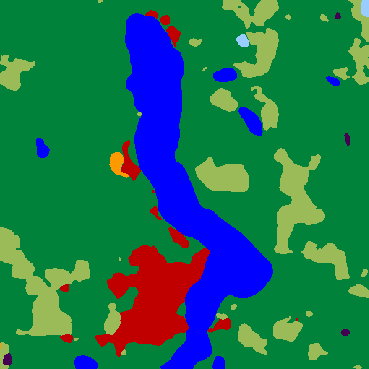} \\
         S2-Optical & GT (DW) & single-modal (S2) & midF & lateF & CromSS-midF & CromSS-lateF
    \end{tabular}
    \caption{Segmentation maps generated by the pretrained models (Encoder-Decoder for Modality 2 in \cref{fig:meth:cromss_details}) with noisy labels taking S2 as inputs. The color coding is the same as that in \cref{fig:data:pre-train} for the DW dataset.}
    \label{fig:exp:pre-trainsegmaps}
\end{figure*}

\begin{figure*}[htp]
    \centering
    \scriptsize
    \begin{tabular}{p{1.95cm}<{\centering}p{1.95cm}<{\centering}p{1.95cm}<{\centering}p{1.95cm}<{\centering}p{1.95cm}<{\centering}p{1.95cm}<{\centering}p{1.95cm}<{\centering}}
         \includegraphics[width=1.05\linewidth]{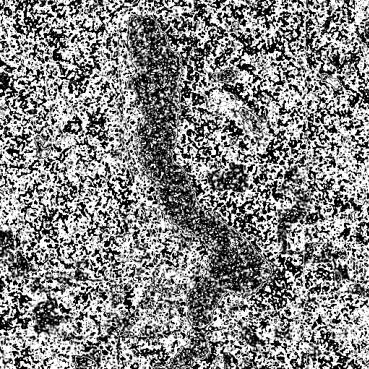} &
         \includegraphics[width=1.05\linewidth]{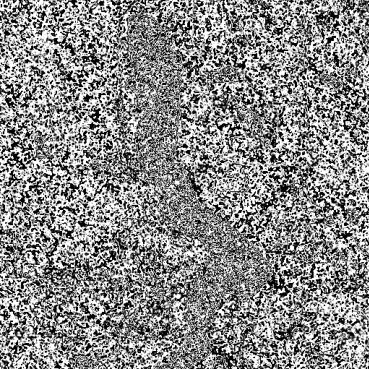} &
         \includegraphics[width=1.05\linewidth]{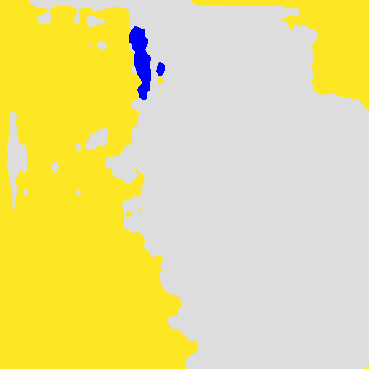} &  
         \includegraphics[width=1.05\linewidth]{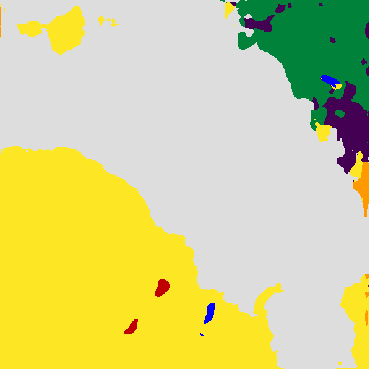} & 
         \includegraphics[width=1.05\linewidth]{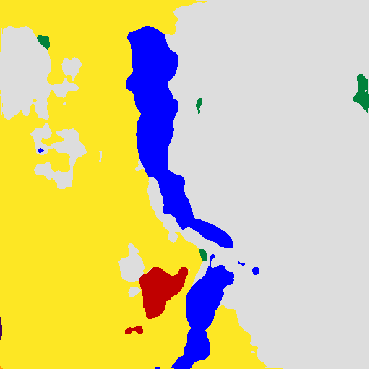} & 
         \includegraphics[width=1.05\linewidth]{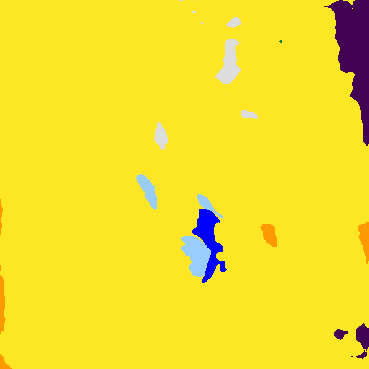} & 
         \includegraphics[width=1.05\linewidth]{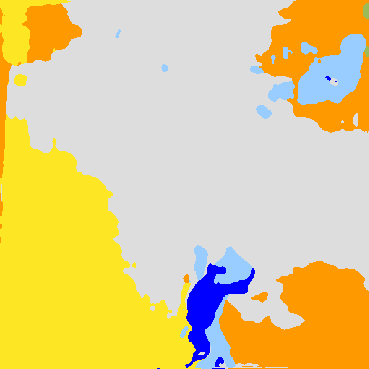} \\
         S1-VV & S1-VH & single-modal (S1) & midF & lateF & CromSS-midF & CromSS-lateF
    \end{tabular}
    \caption{Segmentation maps inferred by the pretrained model (Encoder-Decoder for Modality 1 in \cref{fig:meth:cromss_details}) with noisy labels taking S1 as inputs. The color coding is the same as that in \cref{fig:data:pre-train} for the DW dataset, and the scene illuminated by the radar sensor is the same as in \cref{fig:exp:pre-trainsegmaps} (bottom).}
    \label{fig:exp:pre-trainsegmaps1}
\end{figure*}

To further demonstrate the effectiveness of CromSS, we evaluated the accuracy of the pretrained U-Net models with noisy labels using the SSL4EO-S12@DW test set. As depicted in \cref{fig:exp:pre-trainplots} (a) and (b), multi-modal training, either through middle or late fusion, consistently improves model performance for both S1 and S2 data compared to single-modal training. Remarkably, incorporating cross-modal sample selection enhances the ability of middle fusion to mitigate label noise for both S1 and S2 data. In contrast, the performance gain from late fusion is most prominent for S2 data. This observation again underlines the potential of sharing decoders as a robust constraint for training models with noisy labels.

To analyze the disparity between encoders and decoders, we quantify their similarity in feature statistics (the mean and variance values stored in the batch normalization layers) across models pretrained in single-modal and multi-modal setups for S2 data, as illustrated in Figs. \ref{fig:exp:pre-trainplots} (c) and (d). An intriguing observation we notice is that encoders trained in distinct setups tend to converge to similar representations as training progresses, while, at the same time, decoders exhibit more noticeable discrepancies. This observation becomes more prominent in the later stages of model training. A potential mechanism behind may read: whereas encoders prioritize extracting general features, decoders focus on more specific ones. The discrepancy is more prominent in middle fusion settings than in late fusion. The KL divergence values of single-modal models and the late fusion counterparts approach zero, indicating similar behavior and possibly less need for adjustments.

\begin{figure*}[htp]
    \small
    \centering
    \begin{tabular}{m{0.5cm}<{\centering}m{1.95cm}<{\centering}m{1.95cm}<{\centering}m{1.95cm}<{\centering}m{1.95cm}<{\centering}m{1.95cm}<{\centering}m{1.95cm}}
        & Input & $W_e$ from $F_e$ & $W_e$ from $F'_e$ & $W_l$ from $F_l$ & $W_l$ from $F'_l$ & \hspace{0.2cm}noisy labels \\
        S1 & \includegraphics[width=1.05\linewidth]{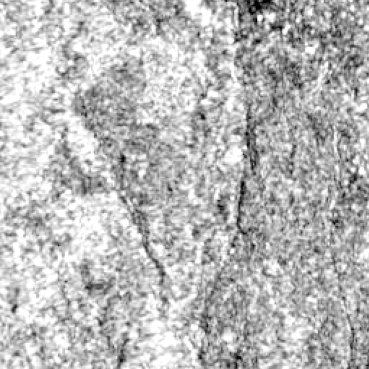} &  
        \includegraphics[width=1.05\linewidth]{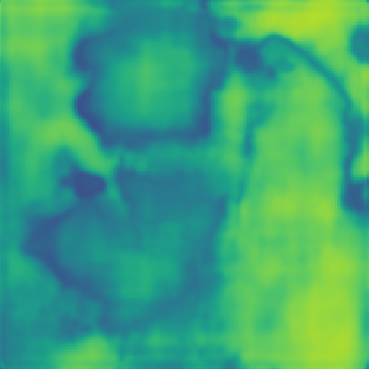} &  
        \includegraphics[width=1.05\linewidth]{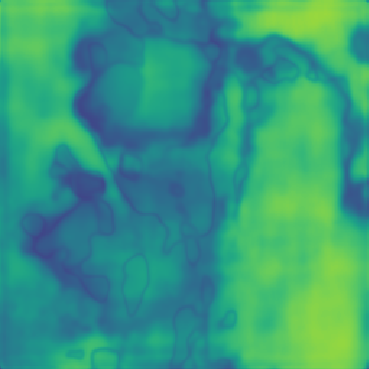} &  
        \includegraphics[width=1.05\linewidth]{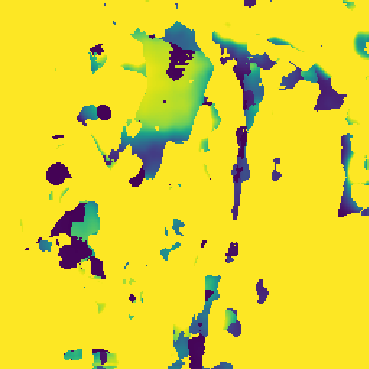} &  
        \includegraphics[width=1.05\linewidth]{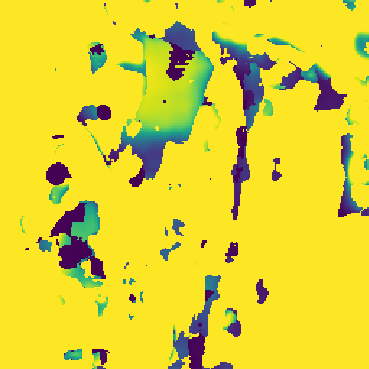} &  
        \includegraphics[width=1.05\linewidth]{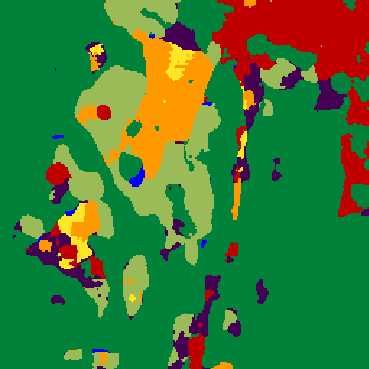} \\
        S2 & \includegraphics[width=1.05\linewidth]{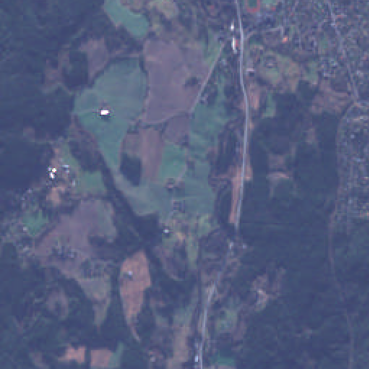} &  
        \includegraphics[width=1.05\linewidth]{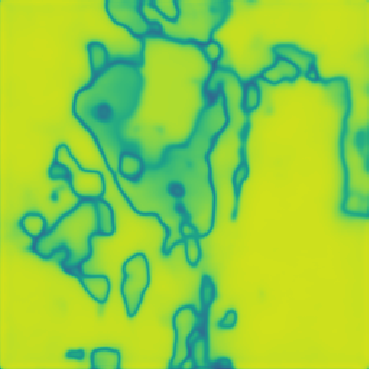} &  
        \includegraphics[width=1.05\linewidth]{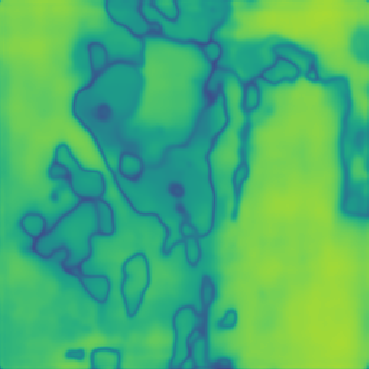} &  
        \includegraphics[width=1.05\linewidth]{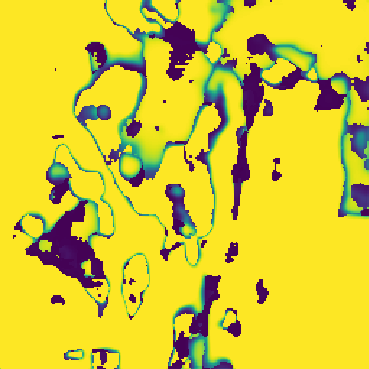} &  
        \includegraphics[width=1.05\linewidth]{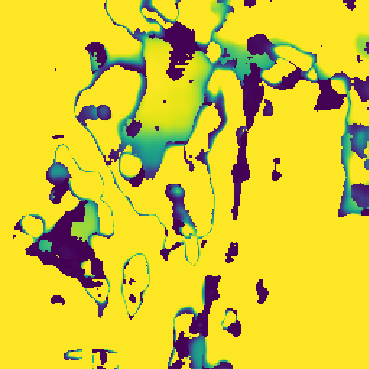} &  
        \includegraphics[height=1.05\linewidth]{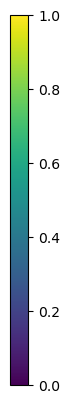} \\
        \hline 
        S1 & 
        \vspace{0.1cm} \includegraphics[width=1.05\linewidth]{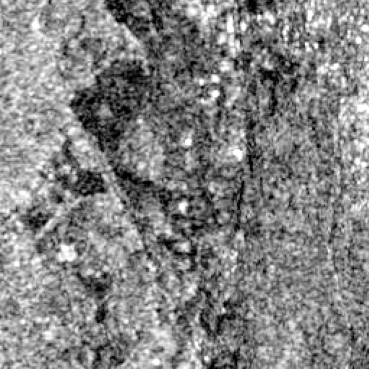} &  
        \vspace{0.1cm} \includegraphics[width=1.05\linewidth]{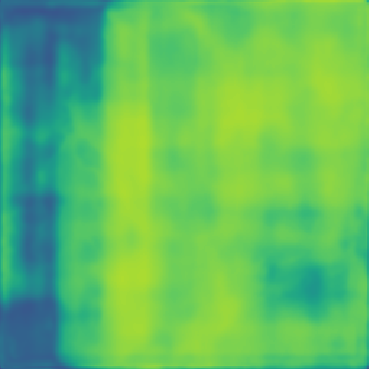} &  
        \vspace{0.1cm} \includegraphics[width=1.05\linewidth]{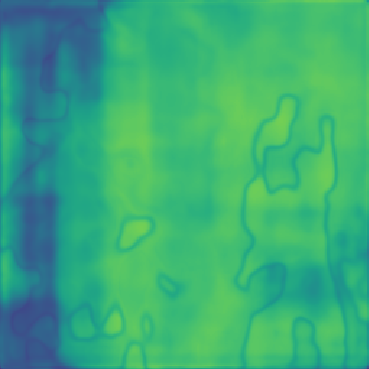} &  
        \vspace{0.1cm} \includegraphics[width=1.05\linewidth]{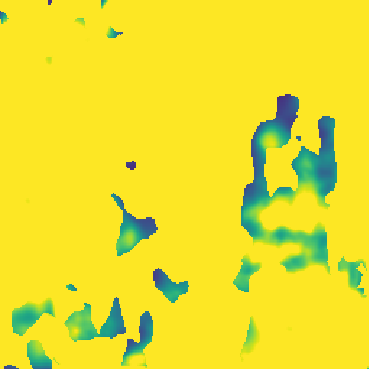} &  
        \vspace{0.1cm} \includegraphics[width=1.05\linewidth]{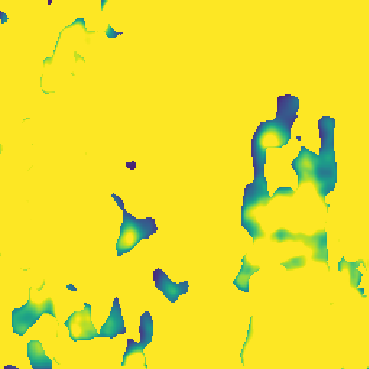} &  
        \vspace{0.1cm} \includegraphics[width=1.05\linewidth]{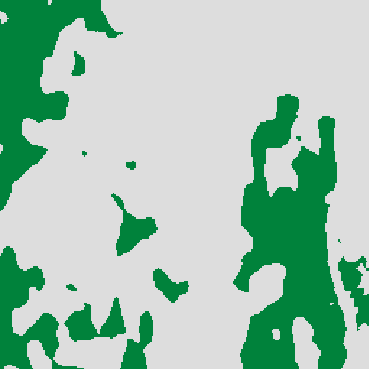} \\
        S2 & \includegraphics[width=1.05\linewidth]{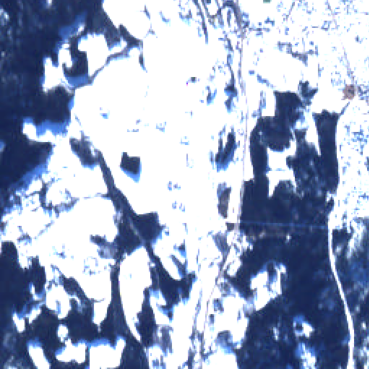} &  
        \includegraphics[width=1.05\linewidth]{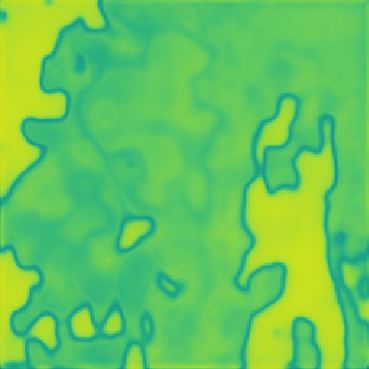} &  
        \includegraphics[width=1.05\linewidth]{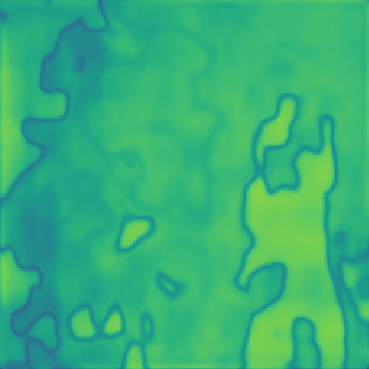} &  
        \includegraphics[width=1.05\linewidth]{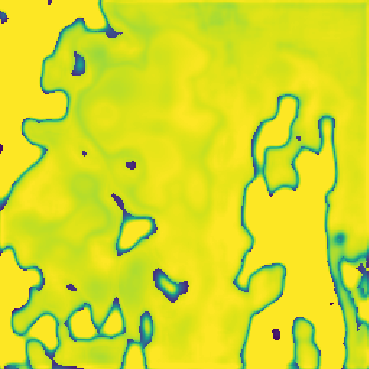} &  
        \includegraphics[width=1.05\linewidth]{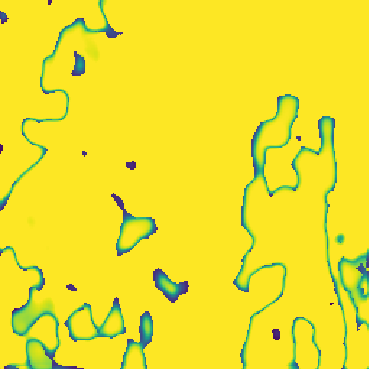} &  
        \includegraphics[height=1.05\linewidth]{Figures/sample_select/probability_cbar.png}\\
    \end{tabular}
    \caption{The entity-based and label-based sample selection masks ($W_e$ and $W_l$) generated by CromSS-midF (13B) from the single-modal and cross-modal confidence masks ($F$ and $F'$) for a given scene at two different seasons. The colorbar for the sample selection maps is displayed on the right-hand side, while the legend for the noisy labels can be found in \cref{fig:data:downstream}.}
    \label{fig:exp:selectionmask}
\end{figure*}

Furthermore, we present two sets of segmentation maps generated from S2 optical images by different U-Net models trained with noisy labels in \cref{fig:exp:pre-trainsegmaps}. In the first set of maps (top row), employing sample selection leads to a more accurate delineation of water bodies. Conversely, in the second set (bottom row), maps generated by CromSS appear somewhat over-smoothed, with spatial details missing. This observation aligns with the degraded performance of encoders pretrained with CromSS on the SSL4EO-S12@DW dataset compared to other multi-modal pretrained schemes, as discussed in \cref{sec:exp:ablation}. 
The sample selection bias is one potential root cause of our observation of over-smoothed semantic segmentation maps that discard spatial details. As illustrated in \cref{fig:exp:selectionmask}, the small objects (cf.\ dark spots in the selection masks $\cy{\boldsymbol{W}_l}$) or class boundaries may get treated as uncertain and thus will get assigned smaller weight values during the selection process. The absence of spatial details may lead to degraded accuracy for a downstream task (e.g., exact DW dataset labels) identical to the noisy labels in the pretraining phase. However, masking geospatially small objects can partly avoid model overfitting to the pretraining dataset for other downstream tasks (e.g., exact OSM dataset labels). Consequently, our sample selection strategy can enhance the effectiveness of noisy label pretrained encoders when transferred to other datasets. Besides, as observed from the last row of \cref{fig:exp:selectionmask}, informing the confidence mask of S2 ($\cy{\boldsymbol{W}_l}$ from $\cy{\boldsymbol{F}_l}$) with that of S1 can reduce uncertainty in the ice region within the final selection mask of S2 ($\cy{\boldsymbol{W}_l}$ from $\cy{\boldsymbol{F}'_l}$). This example indicates the effectiveness of the cross-modal confidence integration in calibrating the single-modal selection masks. 

Finally, we display segmentation maps generated from S1 data by U-Net models in \cref{fig:exp:pre-trainsegmaps1}. On the contrary to \cref{fig:exp:pre-trainsegmaps}, these segmentation maps are much less accurate. As discussed, S1 is a ``weak'' modality for land cover land use classification. We can hardly distinguish different types of landscapes from the VV and VH images by eye, given pronounced noise patterns. Correspondingly, the training on S1 data is less effective than training on S2 data, leading to sub-optimal transfer learning results in the downstream tasks.

\subsection{Comparison in pretrained encoder characteristics} \label{sec:exp:feat}

\begin{figure*}[ht]
    \centering
    \scriptsize
    \begin{tabular}{ccc}
    \includegraphics[height=3.5cm]{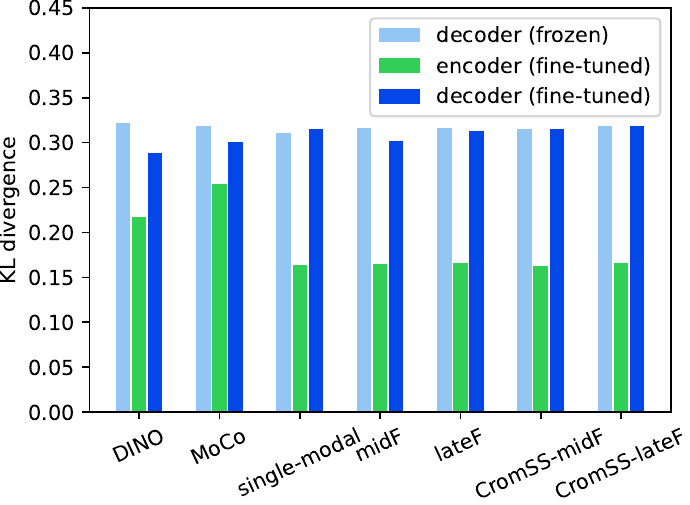}     &  
    \includegraphics[height=3.5cm]{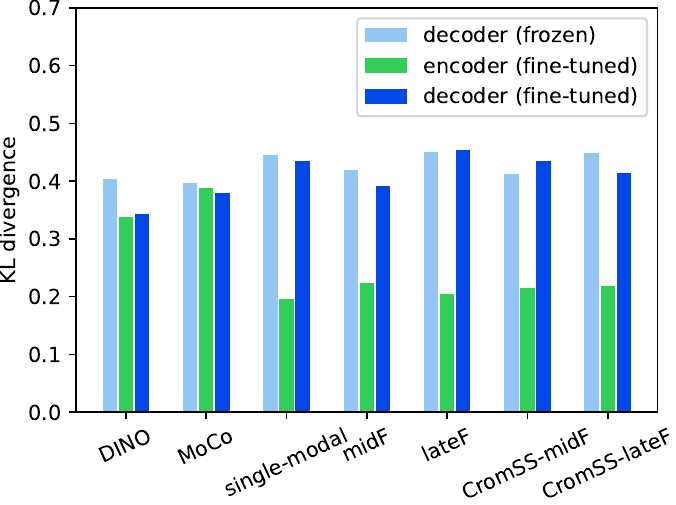}      &
    \includegraphics[height=3.5cm]{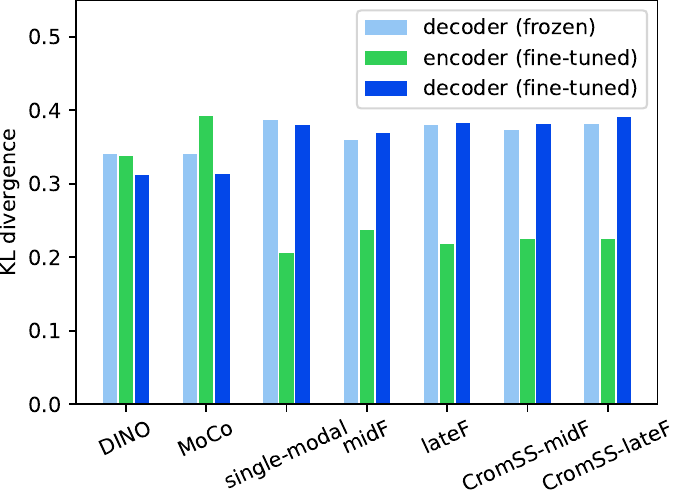}     \\ 
    (a) DFC2020 & (b) DW & (c) OSM \\
    \end{tabular}
    \caption{{KL divergence of the encoder and decoder weights before and after fine-tuning, where the brackets indicate the type of encoder weights. We display -1/log(KL) instead of the raw KL divergence values for improved visualization.}}
    \label{fig:exp:featkl}
\end{figure*}

{To directly compare encoders pretrained with different methods, we present the KL divergence values of encoder (ResNet-50) and decoder weights before and after fine-tuning across various downstream tasks. We choose two classical image-wise SSL methods, DINO and MoCo, as representatives to compare with the noisy label pretraining ones to support the discussion. We use the discrete form of KL divergence and obtain the discrete distributions by binning data with $n=$100 even bins. 
As shown in \cref{fig:exp:featkl}, the differences in the KL divergence of various pretraining methods tend to be smaller for decoders when compared to corresponding encoders. Specifically, encoders pretrained with noisy labels exhibit less optimization change than those pretrained by DINO and MoCo, where MoCo exhibits the largest variation. This suggests that encoders pretrained with noisy labels focusing on pixel-wise information are closer to the optimal solution than those mainly utilizing image-level information.}
Furthermore, we observe that variation in the weights of decoders in the fine-tuning stage is more substantial when using the encoders pretrained on noisy labels. A performant encoder enables decoders to generate better solutions, albeit such an encoder potentially renders the decoder prone to overfitting. These adverse effects need to be taken care of in stages of fine-tuning.

\begin{figure*}[htp]
    \centering
    \scriptsize
    \begin{tabular}{m{3cm}<{\centering}m{2.85cm}<{\centering}m{2.85cm}<{\centering}m{2.85cm}<{\centering}m{2.85cm}<{\centering}}
    \includegraphics[height=2.6cm]{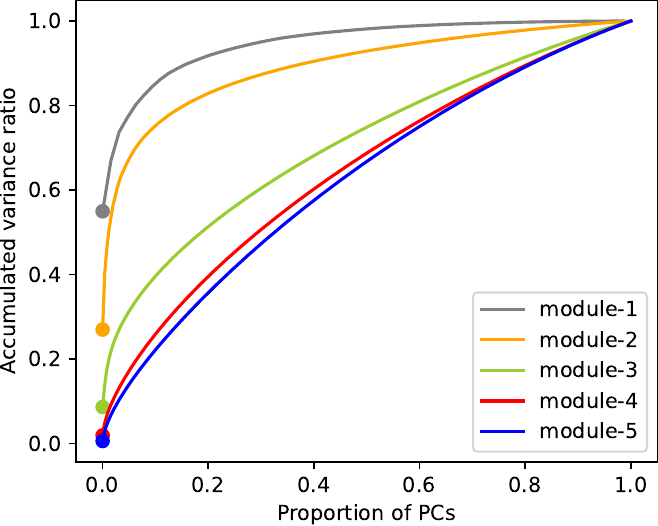} &
    \includegraphics[height=2.6cm]{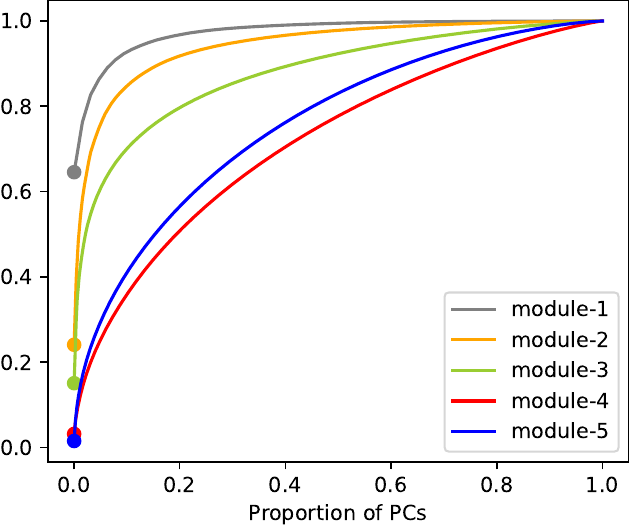} &
    \includegraphics[height=2.6cm]{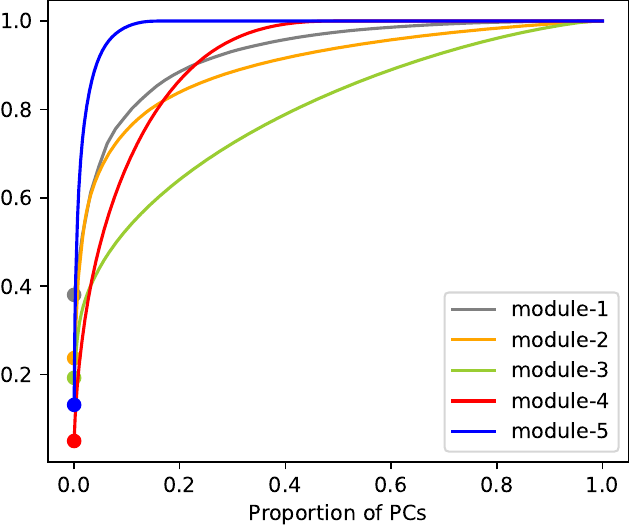} &
    \includegraphics[height=2.6cm]{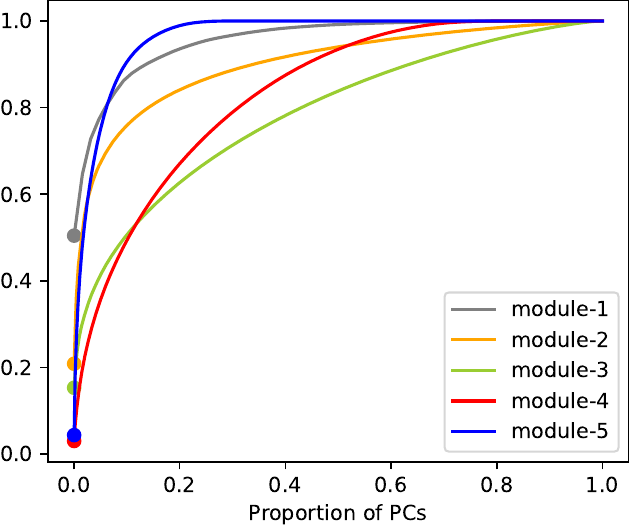} &
    \includegraphics[height=2.6cm]{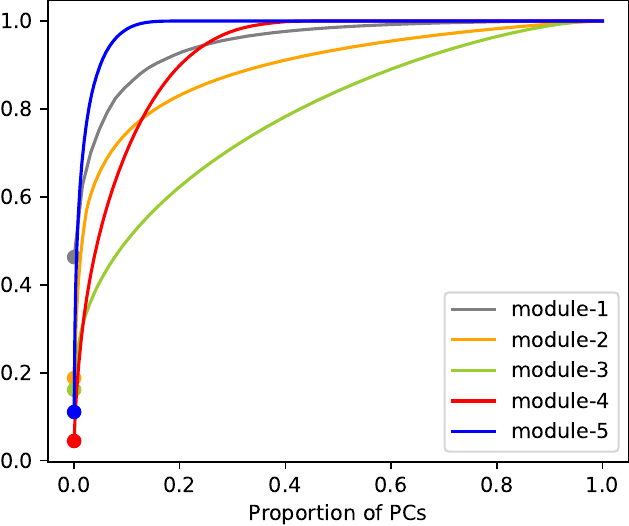} \\
    \includegraphics[height=2.6cm]{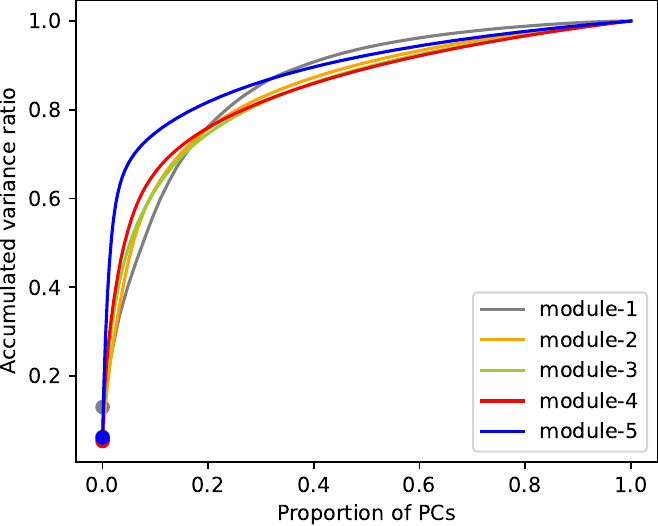} &
    \includegraphics[height=2.6cm]{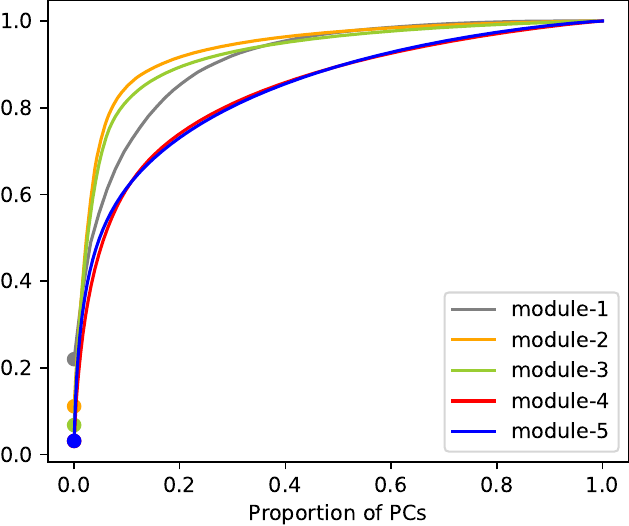} &
    \includegraphics[height=2.6cm]{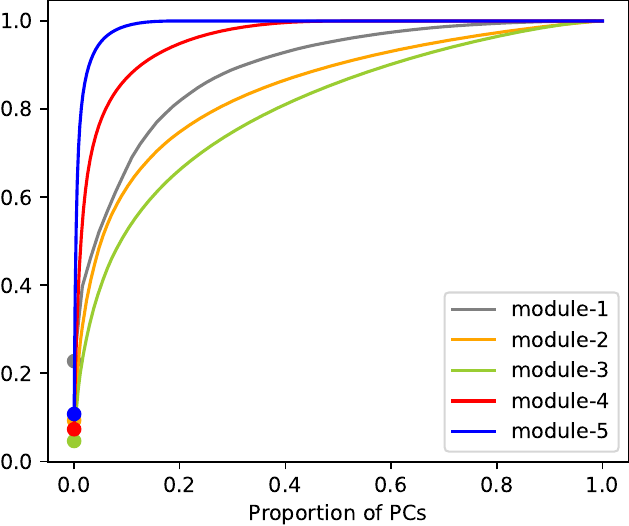} &
    \includegraphics[height=2.6cm]{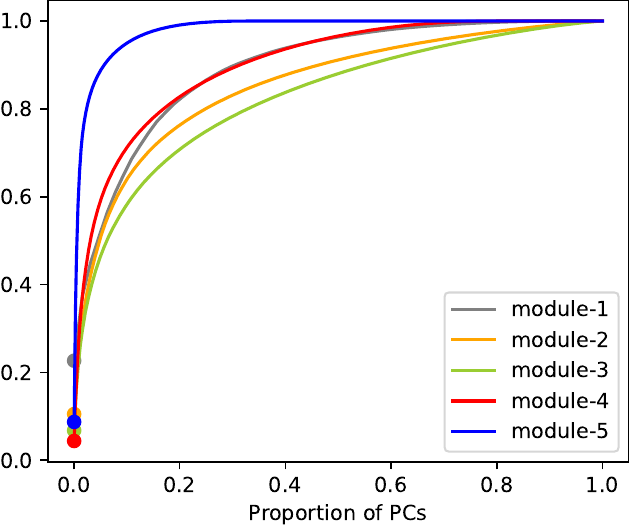} &
    \includegraphics[height=2.6cm]{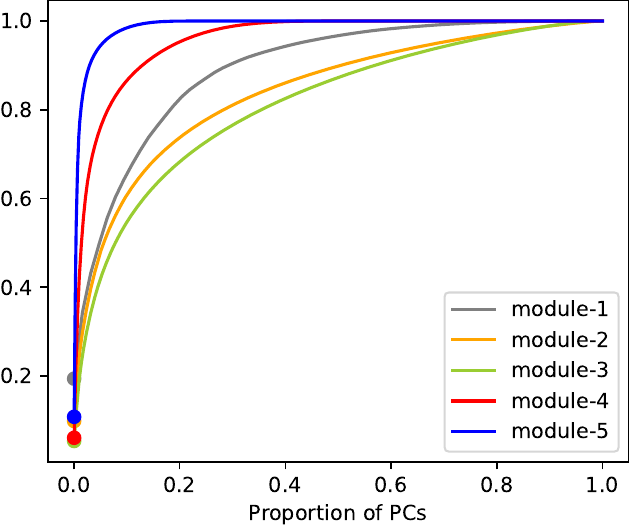} \\
    DINO & MoCo & single-modal & CromSS-midF & CromSS-lateF   \\
    \end{tabular}
    \caption{Accumulated variance ratio plotted against the number of principal components (PCs, in proportion) of output features from various modules/blocks in the ResNet-50 encoders using S2 as inputs. Modules 1-5 represent the modules from the input to the bottleneck. Upper panels: Image-level features generated from the EuroSAT dataset \cite{helber_eurosat_2019}. Lower panels: Pixel-level features generated from the SSL4EO-S12@DW test set.}
    \label{fig:exp:featpca}
\end{figure*}

{Finally, we present a comparison in feature extraction patterns of the ResNet-50 encoders trained by image-wise SSL methods, DINO and MoCo, and our proposed pixel-wise pretraining with noisy labels. We extract the features from each ResNet-50 block/module and then apply Principal Component Analysis (PCA) to get the accumulated variance ratios\footnote{Estimated using the function of sklearn.decomposition.PCA (\url{https://scikit-learn.org/stable/modules/generated/sklearn.decomposition.PCA.html}).} for each feature set. We plot these accumulated variance ratios as curves per module in \cref{fig:exp:featpca}, where the upper and lower rows correspond to the results obtained at the image and pixel levels for EuroSAT and SSL4EO-S12@DW, respectively. 
Empirically, a closer alignment of the accumulated variance ratio curves $y=y(x)$  with the diagonal $y=x$ indicates a more equal importance of all features. A steep slope $dy/dx$ for $x\ll1$ underlines the dominance of a small subset of the features, suggesting higher feature redundancy. It’s important to note that these variance ratios are relative within each feature set, and absolute values are not directly comparable across different sets. Therefore, accumulated variance ratios primarily aim to illustrate the differences in feature distributions across modules within each encoder rather than for direct comparison between models.  
As depicted in \cref{fig:exp:featpca}, the encoders pretrained by image-level SSL methods, DINO and MoCo, typically produce the image-level feature sets with the variance ratio curves of the bottleneck module (module 5) more closely aligning with the diagonal. This tendency gets weak on the pixel-level feature sets. In contrast, the encoders pretrained with pixel-wise noisy labels exhibit the least feature redundancy in the middle layers (module 3), with redundancy increasing in both the shallower layers (modules 1 and 2) and the deeper layers (modules 4 and 5), for either image-level or pixel-level features. This increase is particularly pronounced in the deeper layers (modules 4 and 5). 
These findings highlight the orthogonal nature of encoders pretrained with image-wise (discriminative-like DINO and MoCo) versus pixel-wise (generative-like noisy label pretraining) methods. While pixel-wise pretraining methods emphasize learning detailed, high-resolution features in the shallower layers, they appear to be less sensitive to global semantic information. Nevertheless, the image-wise pretraining is the opposite. This suggests that the choice of pretraining methods significantly influences the model's focus on specific aspects of the images.
}


\section{Conclusions \& Perspectives} \label{sec:conc}

This work introduces a pretraining strategy called CromSS, guided by noisy labels for large-scale remote sensing image segmentation. CromSS exploits a cross-modal sample selection strategy based on the pixel-wise estimated class distributions over classes modeled by multiple sensors/modalities of a given geospatial scene. Two kinds of confidence masks are generated for sample selection purposes: label-based confidence for noisy label sampling with segmentation losses and entity-level confidence for cross-modal consistency modeling with consistency losses. On top of this, a confidence correction strategy is introduced to enhance the confidence masks from one modality through those from another. Furthermore, we test middle and late fusion for pretraining architecture design with noisy labels. To evaluate the effectiveness of the proposed CromSS, we collected a NoLDO-S12 dataset with S1 and S2 as two modalities, alongside the noisy labels sourced from DW for pretraining and GT labels retrieved from DW and OSM serving as downstream tasks. Transfer learning results on three differently defined datasets demonstrate the effectiveness of the CromSS-pretrained ResNet-50 encoders. Specifically, we find that multi-modal training has a strong ability to combat label noise. Sample selection risks losing detailed information when the pretrained models are transferred to the downstream tasks identical to the pretraining one. At the same time, it can boost the generability of pretrained models to some extent when applied to a different setting. 

{Our work represents an initial exploration of using noisy labels for pretraining in segmentation tasks. Though containing noise, these labels are easier to acquire in greater quantities on a larger scale than exact labels. However, the relationship between label quantity and quality remains an open question. In future work, we will investigate this aspect to gain deeper insights into optimizing this trade-off and refining the usage of noisy labels for feature learning.} 
Additionally, we will explore the potential of CromSS for ViT pretraining, comparing it to MIM for remote sensing image segmentation. Another promising research direction is to study the discrepancies in CromSS performance across modalities, particularly focusing on improving its performance on weaker modalities like S1, as observed in our experiments.





{\small
\bibliographystyle{IEEEtran}
\bibliography{refs}

\begin{thebibliography}{10}
\providecommand{\url}[1]{#1}
\csname url@samestyle\endcsname
\providecommand{\newblock}{\relax}
\providecommand{\bibinfo}[2]{#2}
\providecommand{\BIBentrySTDinterwordspacing}{\spaceskip=0pt\relax}
\providecommand{\BIBentryALTinterwordstretchfactor}{4}
\providecommand{\BIBentryALTinterwordspacing}{\spaceskip=\fontdimen2\font plus
\BIBentryALTinterwordstretchfactor\fontdimen3\font minus \fontdimen4\font\relax}
\providecommand{\BIBforeignlanguage}[2]{{%
\expandafter\ifx\csname l@#1\endcsname\relax
\typeout{** WARNING: IEEEtran.bst: No hyphenation pattern has been}%
\typeout{** loaded for the language `#1'. Using the pattern for}%
\typeout{** the default language instead.}%
\else
\language=\csname l@#1\endcsname
\fi
#2}}
\providecommand{\BIBdecl}{\relax}
\BIBdecl

\bibitem{zhao_artificial_2024}
\BIBentryALTinterwordspacing
T.~Zhao, S.~Wang, C.~Ouyang, M.~Chen, C.~Liu, J.~Zhang, L.~Yu, F.~Wang, Y.~Xie, J.~Li, F.~Wang, S.~Grunwald, B.~M. Wong, F.~Zhang, Z.~Qian, Y.~Xu, C.~Yu, W.~Han, T.~Sun, Z.~Shao, T.~Qian, Z.~Chen, J.~Zeng, H.~Zhang, H.~Letu, B.~Zhang, L.~Wang, L.~Luo, C.~Shi, H.~Su, H.~Zhang, S.~Yin, N.~Huang, W.~Zhao, N.~Li, C.~Zheng, Y.~Zhou, C.~Huang, D.~Feng, Q.~Xu, Y.~Wu, D.~Hong, Z.~Wang, Y.~Lin, T.~Zhang, P.~Kumar, A.~Plaza, J.~Chanussot, J.~Zhang, J.~Shi, and L.~Wang, ``Artificial intelligence for geoscience: Progress, challenges, and perspectives,'' \emph{The Innovation}, vol.~5, no.~5, p. 100691, 2024. [Online]. Available: \url{https://www.sciencedirect.com/science/article/pii/S2666675824001292}
\BIBentrySTDinterwordspacing

\bibitem{li_learning_2021}
\BIBentryALTinterwordspacing
Y.~Li, T.~Shi, Y.~Zhang, W.~Chen, Z.~Wang, and H.~Li, ``Learning deep semantic segmentation network under multiple weakly-supervised constraints for cross-domain remote sensing image semantic segmentation,'' \emph{ISPRS Journal of Photogrammetry and Remote Sensing}, vol. 175, pp. 20--33, 2021. [Online]. Available: \url{https://www.sciencedirect.com/science/article/pii/S0924271621000423}
\BIBentrySTDinterwordspacing

\bibitem{wang_self-supervised_2022}
\BIBentryALTinterwordspacing
Y.~Wang, C.~M. Albrecht, N.~A.~A. Braham, L.~Mou, and X.~X. Zhu, ``Self-{Supervised} {Learning} in {Remote} {Sensing}: {A} review,'' \emph{IEEE Geoscience and Remote Sensing Magazine}, vol.~10, no.~4, pp. 213--247, Dec. 2022, number: 4 Conference Name: IEEE Geoscience and Remote Sensing Magazine. [Online]. Available: \url{https://ieeexplore.ieee.org/document/9875399}
\BIBentrySTDinterwordspacing

\bibitem{zhu_foundations_2024}
\BIBentryALTinterwordspacing
X.~X. Zhu, Z.~Xiong, Y.~Wang, A.~J. Stewart, K.~Heidler, Y.~Wang, Z.~Yuan, T.~Dujardin, Q.~Xu, and Y.~Shi, ``On the foundations of earth and climate foundation models,'' May 2024, arXiv:2405.04285 [cs]. [Online]. Available: \url{https://arxiv.org/abs/2405.04285}
\BIBentrySTDinterwordspacing

\bibitem{he_masked_2022}
\BIBentryALTinterwordspacing
K.~He, X.~Chen, S.~Xie, Y.~Li, P.~Dollár, and R.~Girshick, ``\BIBforeignlanguage{en}{Masked {Autoencoders} {Are} {Scalable} {Vision} {Learners}},'' in \emph{\BIBforeignlanguage{en}{Proceedings of the {IEEE}/{CVF} {Conference} on {Computer} {Vision} and {Pattern} {Recognition}}}, 2022, pp. 16\,000--16\,009. [Online]. Available: \url{https://openaccess.thecvf.com/content/CVPR2022/html/He_Masked_Autoencoders_Are_Scalable_Vision_Learners_CVPR_2022_paper}
\BIBentrySTDinterwordspacing

\bibitem{caron_emerging_2021}
\BIBentryALTinterwordspacing
M.~Caron, H.~Touvron, I.~Misra, H.~Jégou, J.~Mairal, P.~Bojanowski, and A.~Joulin, ``\BIBforeignlanguage{en}{Emerging {Properties} in {Self}-{Supervised} {Vision} {Transformers}},'' 2021, pp. 9650--9660. [Online]. Available: \url{https://openaccess.thecvf.com/content/ICCV2021/html/Caron_Emerging_Properties_in_Self-Supervised_Vision_Transformers_ICCV_2021_paper.html}
\BIBentrySTDinterwordspacing

\bibitem{chen_improved_2020}
\BIBentryALTinterwordspacing
X.~Chen, H.~Fan, R.~Girshick, and K.~He, ``Improved {Baselines} with {Momentum} {Contrastive} {Learning},'' Mar. 2020, issue: arXiv:2003.04297 arXiv:2003.04297 [cs]. [Online]. Available: \url{http://arxiv.org/abs/2003.04297}
\BIBentrySTDinterwordspacing

\bibitem{dosovitskiy_image_2021}
\BIBentryALTinterwordspacing
A.~Dosovitskiy, L.~Beyer, A.~Kolesnikov, D.~Weissenborn, X.~Zhai, T.~Unterthiner, M.~Dehghani, M.~Minderer, G.~Heigold, S.~Gelly, J.~Uszkoreit, and N.~Houlsby, ``An {Image} is {Worth} 16x16 {Words}: {Transformers} for {Image} {Recognition} at {Scale},'' Jun. 2021, issue: arXiv:2010.11929 arXiv:2010.11929 [cs]. [Online]. Available: \url{http://arxiv.org/abs/2010.11929}
\BIBentrySTDinterwordspacing

\bibitem{sun_ringmo_2023}
\BIBentryALTinterwordspacing
X.~Sun, P.~Wang, W.~Lu, Z.~Zhu, X.~Lu, Q.~He, J.~Li, X.~Rong, Z.~Yang, H.~Chang, Q.~He, G.~Yang, R.~Wang, J.~Lu, and K.~Fu, ``{RingMo}: {A} {Remote} {Sensing} {Foundation} {Model} {With} {Masked} {Image} {Modeling},'' \emph{IEEE Transactions on Geoscience and Remote Sensing}, vol.~61, pp. 1--22, 2023, conference Name: IEEE Transactions on Geoscience and Remote Sensing. [Online]. Available: \url{https://ieeexplore.ieee.org/abstract/document/9844015}
\BIBentrySTDinterwordspacing

\bibitem{cong_satmae_2022}
\BIBentryALTinterwordspacing
Y.~Cong, S.~Khanna, C.~Meng, P.~Liu, E.~Rozi, Y.~He, M.~Burke, D.~Lobell, and S.~Ermon, ``\BIBforeignlanguage{en}{{SatMAE}: {Pre}-training {Transformers} for {Temporal} and {Multi}-{Spectral} {Satellite} {Imagery}},'' in \emph{\BIBforeignlanguage{en}{Advances in {Neural} {Information} {Processing} {Systems}}}, vol.~35, Dec. 2022, pp. 197--211. [Online]. Available: \url{https://proceedings.neurips.cc/paper_files/paper/2022/hash/01c561df365429f33fcd7a7faa44c985-Abstract-Conference.html}
\BIBentrySTDinterwordspacing

\bibitem{noman_rethinking_2024}
M.~Noman, M.~Naseer, H.~Cholakkal, R.~M. Anwer, S.~Khan, and F.~S. Khan, ``Rethinking transformers pre-training for multi-spectral satellite imagery,'' in \emph{Proceedings of the IEEE/CVF Conference on Computer Vision and Pattern Recognition (CVPR)}, June 2024, pp. 27\,811--27\,819.

\bibitem{reed_scale-mae_2023}
C.~J. Reed, R.~Gupta, S.~Li, S.~Brockman, C.~Funk, B.~Clipp, K.~Keutzer, S.~Candido, M.~Uyttendaele, and T.~Darrell, ``Scale-mae: A scale-aware masked autoencoder for multiscale geospatial representation learning,'' in \emph{Proceedings of the IEEE/CVF International Conference on Computer Vision (ICCV)}, October 2023, pp. 4088--4099.

\bibitem{he_deep_2016}
\BIBentryALTinterwordspacing
K.~He, X.~Zhang, S.~Ren, and J.~Sun, ``Deep {Residual} {Learning} for {Image} {Recognition},'' in \emph{Proceedings of the {IEEE} {Conference} on {Computer} {Vision} and {Pattern} {Recognition}}, 2016, pp. 770--778. [Online]. Available: \url{https://openaccess.thecvf.com/content_cvpr_2016/html/He_Deep_Residual_Learning_CVPR_2016_paper.html}
\BIBentrySTDinterwordspacing

\bibitem{ayush_geography-aware_2021}
\BIBentryALTinterwordspacing
K.~Ayush, B.~Uzkent, C.~Meng, K.~Tanmay, M.~Burke, D.~Lobell, and S.~Ermon, ``\BIBforeignlanguage{en}{Geography-{Aware} {Self}-{Supervised} {Learning}},'' in \emph{\BIBforeignlanguage{en}{Proceedings of the {IEEE}/{CVF} {International} {Conference} on {Computer} {Vision}}}, 2021, pp. 10\,181--10\,190. [Online]. Available: \url{https://openaccess.thecvf.com/content/ICCV2021/html/Ayush_Geography-Aware_Self-Supervised_Learning_ICCV_2021_paper.html}
\BIBentrySTDinterwordspacing

\bibitem{manas_seasonal_2021}
\BIBentryALTinterwordspacing
O.~Mañas, A.~Lacoste, X.~Giro-i Nieto, D.~Vazquez, and P.~Rodriguez, ``\BIBforeignlanguage{en}{Seasonal {Contrast}: {Unsupervised} {Pre}-{Training} from {Uncurated} {Remote} {Sensing} {Data}},'' May 2021, pp. 9414--9423, issue: arXiv:2103.16607 arXiv:2103.16607 [cs] type: article. [Online]. Available: \url{https://openaccess.thecvf.com/content/ICCV2021/html/Manas_Seasonal_Contrast_Unsupervised_Pre-Training_From_Uncurated_Remote_Sensing_Data_ICCV_2021_paper.html}
\BIBentrySTDinterwordspacing

\bibitem{wanyan_dino-mc_2023}
\BIBentryALTinterwordspacing
X.~Wanyan, S.~Seneviratne, S.~Shen, and M.~Kirley, ``{DINO}-{MC}: {Self}-supervised {Contrastive} {Learning} for {Remote} {Sensing} {Imagery} with {Multi}-sized {Local} {Crops},'' Mar. 2023, issue: arXiv:2303.06670 arXiv:2303.06670 [cs]. [Online]. Available: \url{http://arxiv.org/abs/2303.06670}
\BIBentrySTDinterwordspacing

\bibitem{mall_change-aware_2023}
\BIBentryALTinterwordspacing
U.~Mall, B.~Hariharan, and K.~Bala, ``\BIBforeignlanguage{en}{Change-{Aware} {Sampling} and {Contrastive} {Learning} for {Satellite} {Images}},'' 2023, pp. 5261--5270. [Online]. Available: \url{https://openaccess.thecvf.com/content/CVPR2023/html/Mall_Change-Aware_Sampling_and_Contrastive_Learning_for_Satellite_Images_CVPR_2023_paper.html}
\BIBentrySTDinterwordspacing

\bibitem{wang_ssl4eo-s12_2023}
\BIBentryALTinterwordspacing
Y.~Wang, N.~A.~A. Braham, Z.~Xiong, C.~Liu, C.~M. Albrecht, and X.~X. Zhu, ``{SSL4EO}-{S12}: {A} large-scale multimodal, multitemporal dataset for self-supervised learning in {Earth} observation [{Software} and {Data} {Sets}],'' \emph{IEEE Geoscience and Remote Sensing Magazine}, vol.~11, no.~3, pp. 98--106, Sep. 2023, number: 3 Conference Name: IEEE Geoscience and Remote Sensing Magazine. [Online]. Available: \url{https://ieeexplore.ieee.org/abstract/document/10261879}
\BIBentrySTDinterwordspacing

\bibitem{deng_imagenet_2009}
\BIBentryALTinterwordspacing
J.~Deng, W.~Dong, R.~Socher, L.-J. Li, K.~Li, and L.~Fei-Fei, ``{ImageNet}: {A} large-scale hierarchical image database,'' in \emph{2009 {IEEE} {Conference} on {Computer} {Vision} and {Pattern} {Recognition}}, Jun. 2009, pp. 248--255, iSSN: 1063-6919. [Online]. Available: \url{https://ieeexplore.ieee.org/abstract/document/5206848}
\BIBentrySTDinterwordspacing

\bibitem{kirillov_segment_2023}
\BIBentryALTinterwordspacing
A.~Kirillov, E.~Mintun, N.~Ravi, H.~Mao, C.~Rolland, L.~Gustafson, T.~Xiao, S.~Whitehead, A.~C. Berg, W.-Y. Lo, P.~Dollár, and R.~Girshick, ``\BIBforeignlanguage{en}{Segment {Anything}},'' Apr. 2023, arXiv:2304.02643 [cs]. [Online]. Available: \url{http://arxiv.org/abs/2304.02643}
\BIBentrySTDinterwordspacing

\bibitem{ravi_sam_2024}
\BIBentryALTinterwordspacing
N.~Ravi, V.~Gabeur, Y.-T. Hu, R.~Hu, C.~Ryali, T.~Ma, H.~Khedr, R.~Rädle, C.~Rolland, L.~Gustafson, E.~Mintun, J.~Pan, K.~V. Alwala, N.~Carion, C.-Y. Wu, R.~Girshick, P.~Dollar, and C.~Feichtenhofer, ``\BIBforeignlanguage{en}{{SAM} 2: {Segment} {Anything} in {Images} and {Videos}},'' in \emph{\BIBforeignlanguage{en}{Proceedings of the International Conference on Learning Representations (ICLR)}}, Oct. 2024. [Online]. Available: \url{https://openreview.net/forum?id=Ha6RTeWMd0}
\BIBentrySTDinterwordspacing

\bibitem{wang_large_2025}
\BIBentryALTinterwordspacing
H.~Wang, K.~K\"{o}ser, and P.~Ren, ``Large {Foundation} {Model} {Empowered} {Discriminative} {Underwater} {Image} {Enhancement},'' \emph{IEEE Transactions on Geoscience and Remote Sensing}, vol.~63, pp. 1--17, 2025. [Online]. Available: \url{https://ieeexplore.ieee.org/abstract/document/10824846}
\BIBentrySTDinterwordspacing

\bibitem{zhang_understanding_2021}
\BIBentryALTinterwordspacing
C.~Zhang, S.~Bengio, M.~Hardt, B.~Recht, and O.~Vinyals, ``Understanding deep learning (still) requires rethinking generalization,'' \emph{Communications of the ACM}, vol.~64, no.~3, pp. 107--115, Feb. 2021, number: 3. [Online]. Available: \url{https://doi.org/10.1145/3446776}
\BIBentrySTDinterwordspacing

\bibitem{liu_ai02_2024}
C.~Liu, C.~M. Albrecht, Y.~Wang, Q.~Li, and X.~X. Zhu, ``A{I}{O}2: Online correction of object labels for deep learning with incomplete annotation in remote sensing image segmentation,'' \emph{IEEE Transactions on Geoscience and Remote Sensing}, pp. 1--1, 2024.

\bibitem{mahajan_exploring_2018}
\BIBentryALTinterwordspacing
D.~Mahajan, R.~Girshick, V.~Ramanathan, K.~He, M.~Paluri, Y.~Li, A.~Bharambe, and L.~van~der Maaten, ``Exploring the {Limits} of {Weakly} {Supervised} {Pretraining},'' 2018, pp. 181--196. [Online]. Available: \url{https://openaccess.thecvf.com/content_ECCV_2018/html/Dhruv_Mahajan_Exploring_the_Limits_ECCV_2018_paper.html}
\BIBentrySTDinterwordspacing

\bibitem{ghadiyaram_large-scale_2019}
\BIBentryALTinterwordspacing
D.~Ghadiyaram, D.~Tran, and D.~Mahajan, ``Large-{Scale} {Weakly}-{Supervised} {Pre}-{Training} for {Video} {Action} {Recognition},'' 2019, pp. 12\,046--12\,055. [Online]. Available: \url{https://openaccess.thecvf.com/content_CVPR_2019/html/Ghadiyaram_Large-Scale_Weakly-Supervised_Pre-Training_for_Video_Action_Recognition_CVPR_2019_paper.html}
\BIBentrySTDinterwordspacing

\bibitem{kaiser_learning_2017}
P.~Kaiser, J.~D. Wegner, A.~Lucchi, M.~Jaggi, T.~Hofmann, and K.~Schindler, ``Learning {Aerial} {Image} {Segmentation} {From} {Online} {Maps},'' \emph{IEEE Transactions on Geoscience and Remote Sensing}, vol.~55, no.~11, pp. 6054--6068, Nov. 2017, number: 11.

\bibitem{maggiori_convolutional_2017}
E.~Maggiori, Y.~Tarabalka, G.~Charpiat, and P.~Alliez, ``Convolutional {Neural} {Networks} for {Large}-{Scale} {Remote}-{Sensing} {Image} {Classification},'' \emph{IEEE Transactions on Geoscience and Remote Sensing}, vol.~55, no.~2, pp. 645--657, Feb. 2017, number: 2.

\bibitem{liu_task_2024}
C.~Liu, C.~M. Albrecht, Y.~Wang, and X.~X. Zhu, ``Task specific pretraining with noisy labels for remote sensing image segmentation,'' in \emph{IGARSS 2024 - 2024 IEEE International Geoscience and Remote Sensing Symposium}, 2024, pp. 7040--7044.

\bibitem{li_underwater_2025}
\BIBentryALTinterwordspacing
H.~Li, H.~Wang, Y.~Zhang, L.~Li, and P.~Ren, ``Underwater image captioning: {Challenges}, models, and datasets,'' \emph{ISPRS Journal of Photogrammetry and Remote Sensing}, vol. 220, pp. 440--453, Feb. 2025. [Online]. Available: \url{https://www.sciencedirect.com/science/article/pii/S0924271624004726}
\BIBentrySTDinterwordspacing

\bibitem{yuan_easy_2022}
\BIBentryALTinterwordspacing
Z.~Yuan, L.~Mou, Q.~Wang, and X.~X. Zhu, ``From {Easy} to {Hard}: {Learning} {Language}-{Guided} {Curriculum} for {Visual} {Question} {Answering} on {Remote} {Sensing} {Data},'' \emph{IEEE Transactions on Geoscience and Remote Sensing}, vol.~60, pp. 1--11, 2022. [Online]. Available: \url{https://ieeexplore.ieee.org/abstract/document/9771224}
\BIBentrySTDinterwordspacing

\bibitem{schmitt2015}
M.~Schmitt and X.~X. Zhu, ``Data fusion and remote sensing: An ever-growing relationship,'' \emph{IEEE Geoscience and Remote Sensing Magazine}, vol.~4, no.~4, pp. 6--23, 2016.

\bibitem{hong_multimodal_2021}
\BIBentryALTinterwordspacing
D.~Hong, J.~Hu, J.~Yao, J.~Chanussot, and X.~X. Zhu, ``Multimodal remote sensing benchmark datasets for land cover classification with a shared and specific feature learning model,'' \emph{ISPRS Journal of Photogrammetry and Remote Sensing}, vol. 178, pp. 68--80, Aug. 2021. [Online]. Available: \url{https://www.sciencedirect.com/science/article/pii/S0924271621001362}
\BIBentrySTDinterwordspacing

\bibitem{sainte_fare_garnot_multi-modal_2022}
\BIBentryALTinterwordspacing
V.~Sainte Fare~Garnot, L.~Landrieu, and N.~Chehata, ``Multi-modal temporal attention models for crop mapping from satellite time series,'' \emph{ISPRS Journal of Photogrammetry and Remote Sensing}, vol. 187, pp. 294--305, May 2022. [Online]. Available: \url{https://www.sciencedirect.com/science/article/pii/S0924271622000855}
\BIBentrySTDinterwordspacing

\bibitem{cai_improving_2023}
\BIBentryALTinterwordspacing
Z.~Cai, Q.~Hu, X.~Zhang, J.~Yang, H.~Wei, J.~Wang, Y.~Zeng, G.~Yin, W.~Li, L.~You, B.~Xu, and Z.~Shi, ``Improving agricultural field parcel delineation with a dual branch spatiotemporal fusion network by integrating multimodal satellite data,'' \emph{ISPRS Journal of Photogrammetry and Remote Sensing}, vol. 205, pp. 34--49, Nov. 2023. [Online]. Available: \url{https://www.sciencedirect.com/science/article/pii/S0924271623002630}
\BIBentrySTDinterwordspacing

\bibitem{sun_similarity_2024}
\BIBentryALTinterwordspacing
Y.~Sun, L.~Lei, Z.~Li, and G.~Kuang, ``Similarity and dissimilarity relationships based graphs for multimodal change detection,'' \emph{ISPRS Journal of Photogrammetry and Remote Sensing}, vol. 208, pp. 70--88, Feb. 2024. [Online]. Available: \url{https://www.sciencedirect.com/science/article/pii/S0924271624000029}
\BIBentrySTDinterwordspacing

\bibitem{fuller_croma_2023}
\BIBentryALTinterwordspacing
A.~Fuller, K.~Millard, and J.~Green, ``\BIBforeignlanguage{en}{{CROMA}: {Remote} {Sensing} {Representations} with {Contrastive} {Radar}-{Optical} {Masked} {Autoencoders}},'' in \emph{\BIBforeignlanguage{en}{Advances in {Neural} {Information} {Processing} {Systems}}}, vol.~36, Dec. 2023. [Online]. Available: \url{https://proceedings.neurips.cc/paper_files/paper/2023/hash/11822e84689e631615199db3b75cd0e4-Abstract-Conference.html}
\BIBentrySTDinterwordspacing

\bibitem{guo_skysense_2023}
\BIBentryALTinterwordspacing
X.~Guo, J.~Lao, B.~Dang, Y.~Zhang, L.~Yu, L.~Ru, L.~Zhong, Z.~Huang, K.~Wu, D.~Hu, H.~He, J.~Wang, J.~Chen, M.~Yang, Y.~Zhang, and Y.~Li, ``{SkySense}: {A} {Multi}-{Modal} {Remote} {Sensing} {Foundation} {Model} {Towards} {Universal} {Interpretation} for {Earth} {Observation} {Imagery},'' Dec. 2023, issue: arXiv:2312.10115 arXiv:2312.10115 [cs]. [Online]. Available: \url{http://arxiv.org/abs/2312.10115}
\BIBentrySTDinterwordspacing

\bibitem{wang_decur_2023}
Y.~Wang, C.~M. Albrecht, N.~A.~A. Braham, C.~Liu, Z.~Xiong, and X.~X. Zhu, ``Decoupling common and unique representations for multimodal self-supervised learning,'' \emph{arXiv preprint arXiv:2309.05300}, 2024.

\bibitem{xiong_neural_2024}
\BIBentryALTinterwordspacing
Z.~Xiong, Y.~Wang, F.~Zhang, A.~J. Stewart, J.~Hanna, D.~Borth, I.~Papoutsis, B.~L. Saux, G.~Camps-Valls, and X.~X. Zhu, ``Neural {Plasticity}-{Inspired} {Multimodal} {Foundation} {Model} for {Earth} {Observation},'' Jun. 2024, arXiv:2403.15356 [cs]. [Online]. Available: \url{http://arxiv.org/abs/2403.15356}
\BIBentrySTDinterwordspacing

\bibitem{chen_self-supervised_2022}
\BIBentryALTinterwordspacing
Y.~Chen and L.~Bruzzone, ``Self-{Supervised} {SAR}-{Optical} {Data} {Fusion} of {Sentinel}-1/-2 {Images},'' \emph{IEEE Transactions on Geoscience and Remote Sensing}, vol.~60, pp. 1--11, 2022, conference Name: IEEE Transactions on Geoscience and Remote Sensing. [Online]. Available: \url{https://ieeexplore.ieee.org/document/9614157}
\BIBentrySTDinterwordspacing

\bibitem{xie_co-learning_2023}
\BIBentryALTinterwordspacing
Y.~Xie, J.~Tian, and X.~X. Zhu, ``A co-learning method to utilize optical images and photogrammetric point clouds for building extraction,'' \emph{International Journal of Applied Earth Observation and Geoinformation}, vol. 116, p. 103165, Feb. 2023. [Online]. Available: \url{https://www.sciencedirect.com/science/article/pii/S1569843222003533}
\BIBentrySTDinterwordspacing

\bibitem{huang_co-seg_2021}
Z.~Huang, H.~Zhang, A.~Laine, E.~Angelini, C.~Hendon, and Y.~Gan, ``Co-{Seg}: {An} {Image} {Segmentation} {Framework} {Against} {Label} {Corruption},'' in \emph{2021 {IEEE} 18th {International} {Symposium} on {Biomedical} {Imaging} ({ISBI})}, Apr. 2021, pp. 550--553, iSSN: 1945-8452.

\bibitem{malach_decoupling_2017}
\BIBentryALTinterwordspacing
E.~Malach and S.~Shalev-Shwartz, ``Decoupling "when to update" from "how to update",'' in \emph{Advances in {Neural} {Information} {Processing} {Systems}}, vol.~30.\hskip 1em plus 0.5em minus 0.4em\relax Curran Associates, Inc., 2017. [Online]. Available: \url{https://proceedings.neurips.cc/paper_files/paper/2017/hash/58d4d1e7b1e97b258c9ed0b37e02d087-Abstract.html}
\BIBentrySTDinterwordspacing

\bibitem{han_co-teaching_2018}
B.~Han, Q.~Yao, X.~Yu, G.~Niu, M.~Xu, W.~Hu, I.~W. Tsang, and M.~Sugiyama, ``Co-teaching: robust training of deep neural networks with extremely noisy labels,'' in \emph{Proceedings of the 32nd {International} {Conference} on {Neural} {Information} {Processing} {Systems}}, ser. {NIPS}'18.\hskip 1em plus 0.5em minus 0.4em\relax Red Hook, NY, USA: Curran Associates Inc., Dec. 2018, pp. 8536--8546.

\bibitem{ronneberger_u-net_2015}
O.~Ronneberger, P.~Fischer, and T.~Brox, ``\BIBforeignlanguage{en}{U-{Net}: {Convolutional} {Networks} for {Biomedical} {Image} {Segmentation}},'' in \emph{\BIBforeignlanguage{en}{Medical {Image} {Computing} and {Computer}-{Assisted} {Intervention} – {MICCAI} 2015}}, ser. Lecture {Notes} in {Computer} {Science}, N.~Navab, J.~Hornegger, W.~M. Wells, and A.~F. Frangi, Eds.\hskip 1em plus 0.5em minus 0.4em\relax Cham: Springer International Publishing, 2015, pp. 234--241.

\bibitem{brown_dynamic_2022}
\BIBentryALTinterwordspacing
C.~F. Brown, S.~P. Brumby, B.~Guzder-Williams, T.~Birch, S.~B. Hyde, J.~Mazzariello, W.~Czerwinski, V.~J. Pasquarella, R.~Haertel, S.~Ilyushchenko, K.~Schwehr, M.~Weisse, F.~Stolle, C.~Hanson, O.~Guinan, R.~Moore, and A.~M. Tait, ``\BIBforeignlanguage{en}{Dynamic {World}, {Near} real-time global 10 m land use land cover mapping},'' \emph{\BIBforeignlanguage{en}{Scientific Data}}, vol.~9, no.~1, p. 251, Jun. 2022, number: 1. [Online]. Available: \url{https://www.nature.com/articles/s41597-022-01307-4}
\BIBentrySTDinterwordspacing

\bibitem{yokoya_2020_2019}
\BIBentryALTinterwordspacing
N.~Yokoya, ``\BIBforeignlanguage{en}{2020 {IEEE} {GRSS} {Data} {Fusion} {Contest}},'' Dec. 2019. [Online]. Available: \url{https://ieee-dataport.org/competitions/2020-ieee-grss-data-fusion-contest}
\BIBentrySTDinterwordspacing

\bibitem{joint_research_centre_european_commission_lucas_2020}
\BIBentryALTinterwordspacing
{Joint Research Centre (European Commission)}, A.~Jones, O.~Fernández-Ugalde, and S.~Scarpa, \emph{\BIBforeignlanguage{eng}{{LUCAS} 2015 topsoil survey: presentation of dataset and results}}.\hskip 1em plus 0.5em minus 0.4em\relax LU: Publications Office of the European Union, 2020. [Online]. Available: \url{https://data.europa.eu/doi/10.2760/616084}
\BIBentrySTDinterwordspacing

\bibitem{schultz_open_2017}
\BIBentryALTinterwordspacing
M.~Schultz, J.~Voss, M.~Auer, S.~Carter, and A.~Zipf, ``Open land cover from {OpenStreetMap} and remote sensing,'' \emph{International Journal of Applied Earth Observation and Geoinformation}, vol.~63, pp. 206--213, Dec. 2017. [Online]. Available: \url{https://www.sciencedirect.com/science/article/pii/S0303243417301605}
\BIBentrySTDinterwordspacing

\bibitem{noauthor_corine_nodate}
\BIBentryALTinterwordspacing
``\BIBforeignlanguage{en}{{CORINE} {Land} {Cover}}.'' [Online]. Available: \url{https://land.copernicus.eu/en/products/corine-land-cover}
\BIBentrySTDinterwordspacing

\bibitem{jadon_survey_2020}
\BIBentryALTinterwordspacing
S.~Jadon, ``A survey of loss functions for semantic segmentation,'' in \emph{2020 {IEEE} {Conference} on {Computational} {Intelligence} in {Bioinformatics} and {Computational} {Biology} ({CIBCB})}, Oct. 2020, pp. 1--7. [Online]. Available: \url{https://ieeexplore.ieee.org/document/9277638}
\BIBentrySTDinterwordspacing

\bibitem{szegedy_rethinking_2016}
\BIBentryALTinterwordspacing
C.~Szegedy, V.~Vanhoucke, S.~Ioffe, J.~Shlens, and Z.~Wojna, ``Rethinking the {Inception} {Architecture} for {Computer} {Vision},'' in \emph{Proceedings of the {IEEE} {Conference} on {Computer} {Vision} and {Pattern} {Recognition}}, 2016, pp. 2818--2826. [Online]. Available: \url{https://www.cv-foundation.org/openaccess/content_cvpr_2016/html/Szegedy_Rethinking_the_Inception_CVPR_2016_paper.html}
\BIBentrySTDinterwordspacing

\bibitem{kingma_adam_2017}
\BIBentryALTinterwordspacing
D.~P. Kingma and J.~Ba, ``Adam: {A} {Method} for {Stochastic} {Optimization},'' Jan. 2017, arXiv:1412.6980 [cs]. [Online]. Available: \url{http://arxiv.org/abs/1412.6980}
\BIBentrySTDinterwordspacing

\bibitem{zhao_pyramid_2017}
H.~Zhao, J.~Shi, X.~Qi, X.~Wang, and J.~Jia, ``Pyramid {Scene} {Parsing} {Network},'' in \emph{Proceedings of the {IEEE} {Conference} on {Computer} {Vision} and {Pattern} {Recognition}}, 2017, pp. 2881--2890.

\bibitem{chen_encoder-decoder_2018}
L.-C. Chen, Y.~Zhu, G.~Papandreou, F.~Schroff, and H.~Adam, ``Encoder-decoder with atrous separable convolution for semantic image segmentation,'' in \emph{Proceedings of the European Conference on Computer Vision (ECCV)}, September 2018.

\bibitem{lin_feature_2017}
T.-Y. Lin, P.~Dollar, R.~Girshick, K.~He, B.~Hariharan, and S.~Belongie, ``Feature pyramid networks for object detection,'' in \emph{Proceedings of the IEEE Conference on Computer Vision and Pattern Recognition (CVPR)}, July 2017.

\bibitem{Xiao_2018_ECCV}
T.~Xiao, Y.~Liu, B.~Zhou, Y.~Jiang, and J.~Sun, ``Unified perceptual parsing for scene understanding,'' in \emph{Proceedings of the European Conference on Computer Vision (ECCV)}, September 2018.

\bibitem{bastani_satlaspretrain_2023}
F.~Bastani, P.~Wolters, R.~Gupta, J.~Ferdinando, and A.~Kembhavi, ``Satlaspretrain: A large-scale dataset for remote sensing image understanding,'' in \emph{Proceedings of the IEEE/CVF International Conference on Computer Vision (ICCV)}, October 2023, pp. 16\,772--16\,782.

\bibitem{helber_eurosat_2019}
P.~Helber, B.~Bischke, A.~Dengel, and D.~Borth, ``Eurosat: A novel dataset and deep learning benchmark for land use and land cover classification,'' \emph{IEEE Journal of Selected Topics in Applied Earth Observations and Remote Sensing}, 2019.

\end{thebibliography}
}







\end{document}